\documentclass{article} % For LaTeX2e
\usepackage[final]{colm2025_conference}

\usepackage{microtype}
\usepackage{hyperref}
\usepackage{url}
\usepackage{booktabs}

\usepackage{lineno}

\definecolor{darkblue}{rgb}{0, 0, 0.5}
\hypersetup{colorlinks=true, citecolor=darkblue, linkcolor=darkblue, urlcolor=darkblue}

\usepackage{booktabs}
\usepackage{multirow}
\usepackage{mdframed}
\usepackage{lipsum}
\usepackage{xcolor}
\usepackage{color, colortbl}
\usepackage{tcolorbox}
\usepackage{amsfonts}
\tcbuselibrary{theorems}
\usepackage{ulem} % for communication, delete it when official
\usepackage{makecell}
\usepackage{rotating}
\usepackage{graphicx}
\usepackage{enumitem}

\usepackage{subfigure}
\usepackage{subcaption}

\definecolor{tablecolor}{RGB}{251, 198, 194}

\newcommand{\smalldiamond}{\scalebox{0.7}{$\Diamond$}}
\newcommand{\smallspade}{\scalebox{0.6}{$\spadesuit$}}

\definecolor{color_understanding}{HTML}{add8e6}
\definecolor{color_process}{HTML}{E59693}
\definecolor{color_check}{HTML}{6CC3B5}
\definecolor{color_answer}{HTML}{dfbf7f}

\newcommand{\m}[1]{\mathbf{#1}}
\newcommand{\ms}[1]{\boldsymbol{#1}}

\title{Quantization Hurts Reasoning? An Empirical Study on Quantized Reasoning Models}

\author{Ruikang Liu$^{\smalldiamond*}$,
Yuxuan Sun$^{\smallspade*}$,
Manyi Zhang$^{\smallspade}$\thanks{Equal contribution; $\#$: Corresponding authors.},
\ \ Haoli Bai$^{\smallspade\#}$,
Xianzhi Yu$^{\smallspade}$,
\\\textbf{Tiezheng Yu$^{\smallspade}$,}
\textbf{Chun Yuan$^{\smalldiamond}$,}
\textbf{Lu Hou$^{\smallspade\#}$}
\\
$^{\smalldiamond}$Shenzhen International Graduate School, Tsinghua University\\
$^{\smallspade}$Huawei Noah's Ark Lab \\
\texttt{\{baihaoli,houlu3\}@huawei.com} \\
}

\begin{document}

\ifcolmsubmission
\linenumbers
\fi

\maketitle

\begin{abstract}
% Recent advances in large language models (LLMs) \hou{, especially reasoning models} have shown remarkable performance in complex reasoning tasks.
% Despite the increased accuracy, they 
% but they suffer from increased inference overhead due to the long chain-of-thought reasoning steps. 
% While quantization has been extensively studied to reduce the inference cost, 
% there are few studies of the impact on reasoning tasks. In this work, we provide the first overview of the reasoning capabilities of the quantized LLMs, analyzing the effects of different quantization configurations and algorithms, task difficulty, and model scaling. 
% We find that 8-bit weight-activation quantization is lossless across tasks and model sizes, while 4-bit weight-only quantization achieves near lossless accuracies. Quantization tends to degrade performance more on harder tasks but does not generally increase output length, and quantized models are often more accurate than BF16 LLMs of the same size. All quantized models and evaluation scripts will be open-sourced.
Recent advancements in reasoning language models have demonstrated remarkable performance in complex tasks, 
but their extended chain-of-thought reasoning process increases inference overhead. 
While quantization has been widely adopted to reduce the inference cost of large language models, 
its impact on reasoning models remains understudied. 
In this paper, we conduct the first systematic study on quantized reasoning models, 
evaluating the open-sourced DeepSeek-R1-Distilled Qwen and LLaMA families ranging from 1.5B to 70B parameters, QwQ-32B, and Qwen3-8B. 
Our investigation covers weight, KV cache, and activation quantization using state-of-the-art algorithms at varying bit-widths, 
with extensive evaluation across mathematical (AIME, MATH-500), scientific (GPQA), and programming (LiveCodeBench) reasoning benchmarks. 
Our findings reveal that while lossless quantization can be achieved with W8A8 or W4A16 quantization, 
lower bit-widths introduce significant accuracy risks. 
We further identify model size, model origin, and task difficulty as critical determinants of performance. 
Contrary to expectations, quantized models do not exhibit increased output lengths. In addition, strategically scaling the model sizes or reasoning steps can effectively enhance the performance.
% All quantized models and codes will be open-sourced for the community.
All quantized models and codes are open-sourced in \url{https://github.com/ruikangliu/Quantized-Reasoning-Models}.

% The model sizes, origins, and task difficulties are also critical factors that affect performance. 
% Meanwhile, quantized reasoning models do not necessarily generate longer outputs, yet scaling the reasoning steps still improves the performance. 
% \hou{
% Recent reasoning language models, though powerful in many complex reasoning tasks, are slower during inference due to the longer chain-of-thought reasoning process. 
% While various quantization methods have been proposed to reduce the inference cost of large language models, 
% few of them report results on reasoning language models. 
% In this work, we provide the first overview of how quantization affects the performance of reasoning models in aspects of (i) different quantization configurations and algorithms, (ii) task difficulty, and (iii) model size and length scaling. 
% Empirical results show } that 8-bit weight-activation quantization is lossless across tasks and model sizes, 
% while 4-bit weight-only quantization achieves near lossless accuracies. 
% Quantization tends to degrade performance more on harder tasks but does not generally increase output length, and quantized models are often more accurate than BF16 LLMs of the same size. All quantized models and evaluation scripts will be open-sourced\footnote{these findings are not interesting enough, rewrite. A simple way is to include one finding for each of the aspects studied.}.

\end{abstract}

\begin{figure}[h] 
    \centering % 图片居中
    \includegraphics[width=0.98\textwidth]{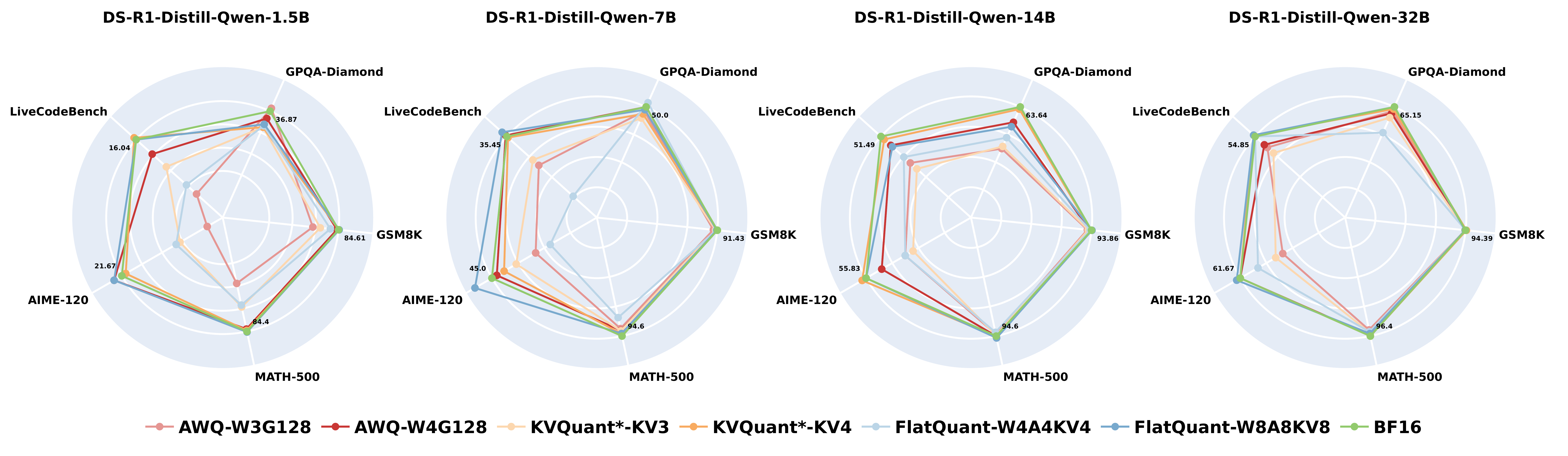} 
    \caption{Performance of the quantized DeepSeek-R1-Distill-Qwen models on five benchmarks: AIME-120, MATH-500, GSM8K, GPQA-Diamond, and LiveCodeBench.
    Each chart presents the performance of different quantization strategies compared to the BF16 baseline.}
    % across the five evaluation benchmarks: AIME-120, MATH-500, GSM8K, GPQA-Diamond, and LiveCodeBench.}
    % The models span four sizes from 1.5B to 32B parameters. Quantized variants maintain high performance across most tasks, with minimal degradation, especially in larger models.} 
    \label{fig:overview} 
\end{figure}

\section{Introduction}
Recent large language models~(LLMs) ~\citep{jaech2024openai,team2025kimi,guo2025deepseek,qwq32b} trained with enhanced reasoning abilities demonstrate strong performance on complex reasoning tasks, such as multi-disciplinary question answering~\citep{rein2024gpqa} and mathematical competitions~\citep{aime2024}.
% and coding~\citep{jain2024livecodebench}. 
However, their improved performance increases inference overhead from prolonged reasoning processes, often causing them to ``overthink'' simple tasks and generate outputs hundreds of times longer than non-reasoning models~\citep{chen2024not}.

To promote efficient LLM inference, various compression and acceleration techniques have been proposed, including quantization~\citep{bai2021binarybert,frantar2022optq,xiao2023smoothquant,liu2024intactkv}, pruning~\citep{frantar2023sparsegpt,sunsimple,zhang2024plug,chen2025simple}, and knowledge distillation~\citep{gu2023minillm,liu2023llm,muralidharan2024compact}. 
Among these methods, quantization, which reduces the numerical precision of weights or activations, is particularly favored due to the good balance between model size and performance preservation. 
Nonetheless, most existing quantization methods are proposed for non-reasoning LLMs~\citep{li2025quantization}, and their performance on reasoning models remains under-explored. 
Intuitively, reasoning models are at higher risks of performance degradation by quantization, since the quantization error may accumulate along the long chain-of-thought (CoT) reasoning steps~\citep{wei2022chain}. 

In this study, we provide a comprehensive empirical study on the quantization of reasoning models, as outlined in Figure~\ref{fig:overview}. 
We evaluate state-of-the-art quantization methods across weight-only, weight-activation, and KV cache quantization. 
The reasoning models for evaluation span form 1.5B to 70B parameters, such as the series of the open-source DeepSeek-R1-Qwen-Distill~\citep{qwen2.5,guo2025deepseek}, DeepSeek-R1-LLaMA-Distill~\citep{llama3modelcard,guo2025deepseek}, QwQ~\citep{qwq32b}, and Qwen3~\citep{yang2025qwen3}, with prevalent reasoning benchmarks such as AIME-120~\citep{aime2024}, MATH-500~\citep{lightman2023let}, GSM8K~\citep{cobbe2021gsm8k}. GPQA-Diamond~\citep{rein2024gpqa} and LiveCodeBench~\citep{jain2024livecodebench}. 
The key findings of this study are summarized below:

\begin{description}[leftmargin=20pt,labelindent=10pt]
    \item[1. Lossless Quantization (\S\ref{sec:rq1})]: 8-bit weight-activation quantization preserves accuracy across tasks and model sizes, while 4-bit weight-only quantization or KV cache quantization also achieves near-lossless results ($\leq$1\% drop).
    \item[2. Quantization Algorithm (\S\ref{sec:rq2})]: Among the evaluated quantization algorithms, we suggest AWQ for weight-only quantization and QuaRot for KV cache quantization. While SmoothQuant suffices for 8-bit weight-activation quantization, FlatQuant takes the lead in the 4-bit scenario.
    \item[3. Impact of Task Difficulty (\S\ref{sec:rq3})]: Harder tasks (e.g., AIME-120) suffer up to 4$\times$ greater degradation than simpler ones (e.g., GSM8K). 
    \item[4. Model Origins Matter (\S\ref{sec:rq4})]: Distillation-based and RL-based reasoning models display varying levels of tolerance to quantization, even when derived from the same base model. Additionally, different model families (e.g., Qwen and LLaMA) exhibit distinct sensitivities to quantization.
    % \item[5. Choice of Calibration Data (\S\ref{sec:discussion_calib})]: Constructing the calibration set from reasoning data is crucial for maintaining performance, especially for methods like GPTQ, while other methods like KVQuant* and FlatQuant are more robust to domain shifts.
    \item[5. Analysis of Output Length (\S\ref{sec:output_length})]: Quantized models with minor performance drops do not exhibit longer reasoning steps, but aggressive low-bit quantization can lead to increased output lengths, particularly in smaller models.
    \item[6. Scaling Effects  (\S\ref{sec:scaling_size}  and \S\ref{sec:scaling_steps})]: Larger quantized models achieve superior accuracy-latency trade-offs compared to smaller BF16 models, and while longer reasoning steps improve performance, the gains diminish at lower rates than in BF16 models.

\end{description}  
% To the best of our knowledge, 
This work is among the first attempts to comprehensively evaluate the quantization of reasoning models, concurrent to  \citet{kurtic2025}. We hope our research provides valuable guidance for the community toward better quantization methods for reasoning models.

\section{Preliminary and Related Work}

% \bai{[to review the basic notations and methods we will be using for evaluation (section 3)]}

% We begin by reviewing the basics of LLM quantization, and introduce the prevalent algorithms to quantize weights, activations and KV cache, which will be used for the evaluation throughout this work. Finally, we also provide a quick recap of the reasoning ability of recent LLMs, and introduce the potential challenges with LLM quantization.

\subsection{Quantization}
\label{sec:prelim_quant}

Quantization converts high-precision values in $\m{X} \in \mathbb{R}^{m\times n}$
% \footnote{notation not right, this is notation for a matrix} 
to lower-precision representations. 
In this study, we focus mainly on hardware-efficient uniform quantization, where the $b$-bit quantization function $\mathcal{Q}(\ \cdot\ ; b)$ can be represented by
\begin{align}
\label{eq:weight_quant}
\hat{\m X} &= \mathcal{Q}(\m X; b) =  s\cdot \Pi_{\Omega(b)}(\m X / s),
% s &= \frac{\textrm{max}(\m W) - \textrm{min}(\m W)}{2^b-1}, 
\end{align}
where $s=\frac{\textrm{max}(\m X) - \textrm{min}(\m X)}{2^b-1} \in \mathbb{R}^{+}$ is the quantization step size, $\Pi(\cdot)$ is the projection function that maps the value to the closest element in the set $\Omega(b)=\{0, 1, ..., 2^{b}-1\}$ of $b$-bit integer points. The scope of this study includes the following quantization configurations, where more implementation details of these quantization algorithms can be found in Appendix~\ref{sec:apdx_algo}.

\paragraph{Weight-only Quantization.}
Following the conventional practice~\citep{frantar2022optq}, we quantize the weight matrices of all linear layers in the Transformer. and leave the rest parameters in BF16. 
By quantizing the values of weight matrix $\m W$ into low-bit integers, the model size and memory access can be effectively reduced during runtime. We focus mainly on GPTQ~\citep{frantar2022optq} and AWQ~\citep{lin2023awq}, two widely applied approaches in the research community. 
We adopt asymmetric quantization with a group size of 128, i.e., the step size is calculated per 128 consecutive elements per channel in $\m W$. 

\paragraph{KV Cache Quantization.}
The KV cache temporarily stores intermediate results during autoregressive generation to improve computational efficiency. The quantization of the KV cache is particularly helpful in reducing the memory access in long sequences, i.e., reasoning with long chain-of-thought steps.  
We study the recently popular methods, including QuaRot~\citep{ashkboos2024quarot} and KVQuant~\citep{hooper2024kvquant}. For QuaRot, we adopt asymmetric per-token KV cache quantization with a group size of 128. We implement KVQuant$^*$ as a simplified baseline, with details in Appendix~\ref{sec:apdx_kv_algo}. KVQuant$^*$ adopts static per-channel asymmetric quantization for K cache and per-token asymmetric quantization for V cache with a group size of 128.
% Besides, we also implement pre-bias K cache quantization for KVQuant$^*$ on DeepSeek-R1-Distill-Qwen-1.5B and 7B models, which can notably improve the accuracy, as detailed in Appendix~\ref{sec:apdx_1.5b_7b}.

% The KV cache is a critical component in transformer-based models like DeepSeek R1/O1, as it stores the key and value vectors used for attention calculations. Quantizing the KV cache can significantly reduce the memory requirements for these vectors, enabling the model to handle longer context windows and more complex tasks. Recent research has shown that the key cache is generally more sensitive to quantization than the value cache, due to its higher norm values and greater impact on error propagation. A mixed-precision quantization strategy, such as allocating more bits to the key cache and fewer bits to the value cache, can effectively mitigate error propagation while achieving significant memory savings. For instance, using 4-bit quantization for the key cache and 2-bit quantization for the value cache has been demonstrated to maintain high model accuracy while reducing memory usage.

\paragraph{Weight-activation Quantization.}
Weight-activation quantization employs low-precision arithmetic for matrix multiplications to further save computation. The evaluated algorithms include SmoothQuant~\citep{xiao2023smoothquant}, QuaRot~\citep{ashkboos2024quarot}, and FlatQuant~\citep{sun2024flatquant}. Following the convention of these methods, we apply per-channel symmetric quantization to weights and per-token asymmetric quantization to activations. We quantize all linear layers in the Transformer, including the KV cache. The KV cache is quantized using group-wise asymmetric quantization with a group size of 128. We leave the queries in full precision, following~\citet{ashkboos2024quarot}. Additionally, inspired by the recent progress on FP4 inference~\citep{nvidia_blackwell_triton}, we evaluate MXFP4~\citep{rouhani2023microscaling} quantization, where every contiguous 32 elements share the same scale.

\subsection{Related Work}
\label{sec:prelim_related}
% recap of reasoning-models
% In this study, we provide a comprehensive evaluation of the reasoning abilities of quantized LLMs. 

\paragraph{Families of Reasoning Models.}
Recent advancements in reasoning models like OpenAI's o1~\citep{jaech2024openai} and DeepSeek’s R1~\citep{guo2025deepseek}  have demonstrated the potential to solve complex tasks that require logical deduction, multi-step reasoning, and contextual understanding. These models can be broadly categorized based on how they achieve their reasoning abilities.
% : either through distillation from a larger, pre-trained model or via self post-training techniques.} 
% The recent advances of reasoning models can be divided into the following categories, depending on their training paradigm. 
The first category of reasoning models achieves the reasoning ability by distilling from the outputs of existing stronger models. For instance, DeepSeek-R1-Distill-Qwen and DeepSeek-R1-Distill-LLaMA are distilled from 800K curated samples from DeepSeek-R1~\citep{guo2025deepseek}. 
The other type of reasoning models obtain the reasoning ability via self post-training techniques like supervised finetuning~\citep{ye2025limo,muennighoff2025s1}, reinforcement learning (RL)~\citep{guo2025deepseek,qwq32b}, or their combinations~\citep{qwq32b,team2025kimi,guo2025deepseek}.

\paragraph{Efficient Reasoning.}
Despite the enhanced reasoning abilities, the long chain-of-thought reasoning process also incurs a greater challenge of inference overhead due to the increased output length. To reduce the inference cost, several approaches have been proposed.
For reasoning models trained with reinforcement learning,  a common way is to add a length penalty term on the reward in addition to the accuracy~\citep{luo2025o1,aggarwal2025l1,team2025kimi}. 
 % to balance the reasoning length and accuracy
% effectively shortening the reasoning steps while maintaining task performance
Supervised fine-tuning with variable-length Chain-of-Thought (CoT) data is another effective approach. 
For instance, Token-Budget~\citep{han2024token} guides LLMs to complete reasoning within a specified token limit, while S1~\citep{muennighoff2025s1}  
controls the output length through forcefully terminating the model’s thinking process or lengthening it by appending “Wait” multiple times.
% appends special tokens to control the reasoning steps. 
One can also reduce the reasoning lengths via direct performance optimization~(DPO), where short and long CoT data can be curated as positive and negative data pairs~\citep{chen2024not,team2025kimi}, or merging the long CoT model with a short CoT model~\citep{team2025kimi}. Apart from training, efficient reasoning can also be realized by better test-time scaling with parallel inference~\citep{rodionov2025hogwild,wang2025faster,pan2025learning}.
% Dynamic reasoning methods like speculative rejection~\citep{shi2024keep} is another direction to  optimize inference by dynamically rejecting unpromising sequences. 

% target and connection with past research
Among the above efforts, quantization is an orthogonal method to achieve efficient reasoning. 
We believe a comprehensive empirical study on quantized reasoning models is essential to fully understand the trade-offs between reasoning efficiency and accuracy.

\section{Evaluation of Quantized Reasoning Models}

\subsection{Setup}
\label{sec:exp_setup}
\paragraph{Evaluation Benchmarks.} We evaluate the quantized models with the algorithms in \S~\ref{sec:prelim_quant} on the following reasoning benchmarks. 
1) Three mathematical reasoning benchmarks sorted by their difficulty: AIME-120 which consists of 120 problems from the American Invitational Mathematics Examination (AIME) from 2022 to 2025 to minimize evaluation variations; MATH-500~\citep{lightman2023let}, a benchmark that contains a mix of easy and hard mathematical problems designed to test comprehensive reasoning abilities; and GSM8K~\citep{cobbe2021gsm8k}, a dataset composed of primary school level questions focused on basic arithmetic and algebra.
2) LiveCodeBench~\citep{jain2024livecodebench}, a benchmark for evaluating large language models on code generation tasks, designed to assess their ability to produce high-quality and functional code.
3) GPQA-Diamond~\citep{rein2024gpqa}, a graduate-level proof question and answer benchmark that tests the ability of models to generate accurate mathematical proofs.
We use Lighteval~\citep{lighteval} with the vLLM~\citep{kwon2023efficient} backend for evaluation, with a sampling temperature of $0.6$ and top-$p$ of $0.95$. The maximum number of generation tokens is limited to 32,768. For all results, we repeat the experiments with three different seeds to reduce the evaluation variations.

\paragraph{The Evaluated Reasoning Models.}
\label{sec:rq}
We evaluate the two categories of reasoning models as described in \S~\ref{sec:prelim_related}. For distillation-based reasoning models, we adopt the series of DeepSeek-R1-Distill-Qwen~\citep{qwen2.5} with varying sizes from 1.5B, 7B, 14B and 32B, and DeepSeek-R1-Distill-LLaMA~\citep{llama3modelcard}, with both 8B and 70B models. 
For reasoning models obtained via reinforcement learning, we choose QwQ-32B~\citep{qwq32b} and Qwen-3-8B~\citep{yang2025qwen3}, the recent performant open-source models.

We study the following research questions in the next sections.

\begin{tcolorbox}[colback=green!5,colframe=green!35!black, left=1ex]
\begin{description}    
    \item[RQ1] (\S\ref{sec:rq1}): What are the \textbf{lossless quantization configurations} for reasoning models?
    \item[RQ2] (\S\ref{sec:rq2}): What are the preferred \textbf{quantization algorithms} on reasoning tasks?
    % \item[RQ3] (\S\ref{sec:rq4}): \textit{What is the effect of calibration data on reasoning?}
    \item[RQ3] (\S\ref{sec:rq3}): 
    How does the \textbf{task difficulty} influence the quantized LLMs?
    \item[RQ4] (\S\ref{sec:rq4}): 
    What is the impact of \textbf{LLM origins} on quantization?
\end{description}
\end{tcolorbox}

\begin{table*}[!t]
\begin{center}
\resizebox{1.0\linewidth}{!}
{
    \begin{tabular}{c|l|l|l|rrr|r|r|r|r}
    \hline\hline
    \textbf{Model} & \textbf{Quantization} & \textbf{\makecell{W-A-KV\\\# Bits}} & \textbf{Methods} & \textbf{\makecell{AIME-\\120}}  & \textbf{\makecell{MATH-\\500}} & \textbf{GSM8K} & \textbf{\makecell{GPQA-\\Diamond}} & \textbf{\makecell{LiveCode-\\Bench}} & \textbf{Avg.} & \textbf{Drop}$\downarrow$ \\ 
    \hline
    %%%%%%%%%%%%%%%%%%%%%%%%%%%%%%%%%%%%%%%%%%%%%%%%%%%%%%%%%%%%%%%%%%%%%%%%%%%%%%%%%%%%%%
    % DeepSeek-R1-Distill-Qwen-1.5B
    \multirow{15}{*}{\rotatebox[origin=c]{90}{\textbf{DeepSeek-R1-Distill-Qwen-1.5B}}}
    & \textbf{BF16} & - & - & \textbf{23.3$_{\pm 2.2}$} & \textbf{84.7$_{\pm 1.5}$} & \textbf{84.5$_{\pm 1.3}$} & \textbf{36.2$_{\pm 0.8}$} & \textbf{16.4$_{\pm 1.1}$} & \textbf{49.0$_{\pm 0.3}$} & - \\
    \cline{2-11}
    & \multirow{4}{*}{\textbf{\textit{Weight-only}}} &  & AWQ & 21.1$_{\pm 2.1}$ & 81.4$_{\pm 1.6}$ & 82.7$_{\pm 0.7}$ & 35.4$_{\pm 2.7}$ & 13.9$_{\pm 1.5}$ & 46.9$_{\pm 1.2}$ & \cellcolor{orange!10}{\footnotesize -2.1} \\  
    &  & \multirow{-2}{*}{4-16-16} & GPTQ & 21.4$_{\pm 3.9}$ & 83.0$_{\pm 1.0}$ & 83.3$_{\pm 0.6}$ & 32.0$_{\pm 5.2}$ & 13.4$_{\pm 0.8}$ & 46.6$_{\pm 1.9}$ & \cellcolor{orange!10}{\footnotesize -2.4} \\  
    \cline{3-11}
    &  & \multirow{2}{*}{3-16-16} & AWQ & 4.4$_{\pm 1.3}$ & 48.1$_{\pm 0.3}$ & 64.5$_{\pm 1.3}$ & 31.8$_{\pm 6.1}$ & 3.2$_{\pm 1.1}$ & 30.4$_{\pm 1.3}$ & \cellcolor{red!10}{\footnotesize -18.6} \\  
    &  &  & GPTQ & 15.6$_{\pm 1.3}$ & 75.3$_{\pm 0.4}$ & 77.8$_{\pm 0.7}$ & 29.8$_{\pm 4.0}$ & 8.7$_{\pm 0.8}$ & 41.4$_{\pm 0.7}$ & \cellcolor{red!10}{\footnotesize -7.6} \\
    \cline{2-11}
    & \multirow{4}{*}{\textbf{\textit{KV Cache}}} & \multirow{2}{*}{16-16-4} & KVQuant$^*$ & 20.3$_{\pm 1.0}$ & 83.9$_{\pm 0.3}$ & 84.2$_{\pm 0.6}$ & 34.0$_{\pm 2.8}$ & 15.9$_{\pm 1.4}$ & 47.7$_{\pm 0.4}$ & \cellcolor{orange!10}{\footnotesize -1.4} \\  
    &  &  & QuaRot & 0.3$_{\pm 0.5}$ & 1.5$_{\pm 0.1}$ & 0.9$_{\pm 0.2}$ & 9.6$_{\pm 2.6}$ & 0.0$_{\pm 0.0}$ & 2.5$_{\pm 0.5}$ & \cellcolor{red!10}{\footnotesize -46.6} \\  
    \cline{3-11}
     &  & \multirow{2}{*}{16-16-3} & KVQuant$^*$ & 6.7$_{\pm 2.5}$ & 65.7$_{\pm 1.2}$ & 70.2$_{\pm 0.7}$ & 30.5$_{\pm 1.1}$ & 10.8$_{\pm 0.4}$ & 36.8$_{\pm 0.9}$ & \cellcolor{red!10}{\footnotesize -12.3} \\  
    &  &  & QuaRot & 0.0$_{\pm 0.0}$ & 1.3$_{\pm 0.4}$ & 0.9$_{\pm 0.3}$ & 21.0$_{\pm 2.8}$ & 0.0$_{\pm 0.0}$ & 4.7$_{\pm 0.6}$ & \cellcolor{red!10}{\footnotesize -44.4} \\  
    \cline{2-11}
    & \multirow{6}{*}{\textbf{\textit{Weight-Act.}}} &  & SmoothQuant  & 16.9$_{\pm 2.4}$ & 78.9$_{\pm 1.0}$ & 83.0$_{\pm 0.8}$ & 31.3$_{\pm 2.2}$ & 15.7$_{\pm 1.0}$ & 45.2$_{\pm 0.4}$ & \cellcolor{red!10}{\footnotesize -3.9} \\   
    &  & 8-8-8 & QuaRot & 22.8$_{\pm 1.3}$ & 84.2$_{\pm 0.9}$ & 83.8$_{\pm 0.8}$ & 38.4$_{\pm 2.2}$ & 15.7$_{\pm 2.3}$ & 49.0$_{\pm 1.0}$ & \cellcolor{green!10}{\footnotesize -0.1} \\  
    &  &  & FlatQuant & 22.5$_{\pm 0.8}$ & 85.0$_{\pm 0.5}$ & 84.7$_{\pm 0.4}$ & 37.0$_{\pm 4.1}$ & 16.8$_{\pm 0.4}$ & 49.2$_{\pm 0.6}$ & \cellcolor{green!10}{\footnotesize 0.2} \\  
    \cline{3-11}
    &  & \multirow{3}{*}{4-4-4} & MXFP4 & 0.0$_{\pm 0.0}$ & 1.1$_{\pm 0.5}$ & 0.6$_{\pm 0.2}$ & 18.4$_{\pm 2.5}$ & 0.0$_{\pm 0.0}$ & 4.0$_{\pm 0.4}$ & \cellcolor{red!10}{\footnotesize -45.0} \\  
    &  &  & QuaRot & 0.0$_{\pm 0.0}$ & 1.4$_{\pm 1.1}$ & 0.9$_{\pm 0.3}$ & 12.0$_{\pm 1.1}$ & 0.0$_{\pm 0.0}$ & 2.9$_{\pm 0.2}$ & \cellcolor{red!10}{\footnotesize -46.2} \\ 
    &  &  & FlatQuant & 8.6$_{\pm 1.3}$ & 66.1$_{\pm 1.7}$ & 76.8$_{\pm 2.0}$ & 32.0$_{\pm 0.8}$ & 7.5$_{\pm 0.8}$ & 38.2$_{\pm 0.3}$ & \cellcolor{red!10}{\footnotesize -10.9} \\
    \cline{3-11}
    \hline\hline
    %%%%%%%%%%%%%%%%%%%%%%%%%%%%%%%%%%%%%%%%%%%%%%%%%%%%%%%%%%%%%%%%%%%%%%%%%%%%%%%%%%%%%%
    % DeepSeek-R1-Distill-Qwen-7B
    \multirow{13}{*}{\rotatebox[origin=c]{90}{\textbf{DeepSeek-R1-Distill-Qwen-7B}}}
    & \textbf{BF16} & - & - & \textbf{46.1$_{\pm 1.0}$} & \textbf{93.9$_{\pm 0.7}$} & \textbf{91.2$_{\pm 0.6}$} & \textbf{51.0$_{\pm 1.0}$} & \textbf{36.7$_{\pm 2.5}$} & \textbf{63.8$_{\pm 0.6}$} & - \\
    \cline{2-11}
    & \multirow{4}{*}{\textbf{\textit{Weight-only}}} &  &  AWQ & 42.2$_{\pm 1.0}$ & 92.5$_{\pm 1.4}$ & 90.7$_{\pm 0.4}$ & 48.1$_{\pm 2.8}$ & 35.1$_{\pm 2.9}$ & 61.7$_{\pm 1.1}$ & \cellcolor{orange!10}{\footnotesize -2.1} \\  
    &  & \multirow{-2}{*}{4-16-16} & GPTQ & 44.4$_{\pm 3.4}$ & 93.3$_{\pm 0.1}$ & 91.0$_{\pm 0.7}$ & 49.0$_{\pm 1.0}$ & 33.3$_{\pm 1.1}$ & 62.2$_{\pm 0.8}$ & \cellcolor{orange!10}{\footnotesize -1.6} \\
    \cline{3-11}
    &  & \multirow{2}{*}{3-16-16} & AWQ & 32.2$_{\pm 3.2}$ & 91.0$_{\pm 1.4}$ & 89.5$_{\pm 0.4}$ & 47.8$_{\pm 1.2}$ & 27.2$_{\pm 1.0}$ & 57.6$_{\pm 0.8}$ & \cellcolor{red!10}{\footnotesize -6.3}  \\  
    &  &  & GPTQ & 37.2$_{\pm 2.9}$ & 91.2$_{\pm 0.2}$ & 89.7$_{\pm 0.6}$ & 47.6$_{\pm 2.5}$ & 26.4$_{\pm 1.1}$ & 58.4$_{\pm 0.8}$ & \cellcolor{red!10}{\footnotesize -5.4} \\
    \cline{2-11}
    & \multirow{4}{*}{\textbf{\textit{KV Cache}}} & \multirow{2}{*}{16-16-4} & KVQuant$^*$ & 44.4$_{\pm 3.8}$ & 93.4$_{\pm 0.2}$ & 90.9$_{\pm 0.7}$ & 49.0$_{\pm 1.5}$ & 36.3$_{\pm 0.8}$ & 62.8$_{\pm 1.0}$ & \cellcolor{green!10}{\footnotesize -1.0} \\ 
    &  &  &  QuaRot & 0.0$_{\pm 0.0}$ & 1.0$_{\pm 0.6}$ & 0.7$_{\pm 0.1}$ & 22.9$_{\pm 5.3}$ & 0.0$_{\pm 0.0}$ & 4.9$_{\pm 1.2}$ & \cellcolor{red!10}{\footnotesize -58.9} \\  
    \cline{3-11}
    &  & \multirow{2}{*}{16-16-3} & KVQuant$^*$ & 33.6$_{\pm 3.9}$ & 91.5$_{\pm 0.8}$ & 90.5$_{\pm 0.6}$ & 46.1$_{\pm 0.8}$ & 28.1$_{\pm 3.1}$ & 58.0$_{\pm 1.4}$ & \cellcolor{red!10}{\footnotesize -5.9} \\  
    &  &  &  QuaRot & 0.3$_{\pm 0.5}$ & 0.8$_{\pm 0.2}$ & 0.4$_{\pm 0.3}$ & 24.6$_{\pm 1.8}$ & 0.0$_{\pm 0.0}$ & 5.2$_{\pm 0.4}$ & \cellcolor{red!10}{\footnotesize -58.6} \\  
    \cline{2-11}
    & \multirow{4}{*}{\textbf{\textit{Weight-Act.}}} &  & SmoothQuant & 45.0$_{\pm 3.8}$ & 94.0$_{\pm 0.7}$ & 90.8$_{\pm 0.3}$ & 50.7$_{\pm 1.3}$ & 35.7$_{\pm 1.1}$ & 63.2$_{\pm 0.6}$ & \cellcolor{green!10}{\footnotesize -0.6} \\   
    &  & \multirow{-2}{*}{8-8-8} & FlatQuant & 47.8$_{\pm 2.7}$ & 93.8$_{\pm 0.8}$ & 91.0$_{\pm 0.6}$ & 49.3$_{\pm 0.6}$ & 36.2$_{\pm 1.0}$ & 63.6$_{\pm 0.5}$ & \cellcolor{green!10}{\footnotesize -0.2} \\  
    \cline{3-11}
    &  & \multirow{2}{*}{4-4-4} & MXFP4 & 0.0$_{\pm 0.0}$ & 0.5$_{\pm 0.1}$ & 0.2$_{\pm 0.1}$ & 20.0$_{\pm 2.3}$ & 0.0$_{\pm 0.0}$ & 4.1$_{\pm 0.5}$ & \cellcolor{red!10}{\footnotesize -59.7} \\  
    &  &  & FlatQuant & 26.1$_{\pm 1.3}$ & 84.1$_{\pm 1.3}$ & 90.9$_{\pm 0.4}$ & 46.6$_{\pm 4.6}$ & 12.4$_{\pm 2.3}$ & 52.0$_{\pm 0.9}$ & \cellcolor{red!10}{\footnotesize -11.8} \\
    \cline{3-11}
    \hline\hline
    %%%%%%%%%%%%%%%%%%%%%%%%%%%%%%%%%%%%%%%%%%%%%%%%%%%%%%%%%%%%%%%%%%%%%%%%%%%%%%%%%%%%%%
    % DeepSeek-R1-Distill-Qwen-14B
    \multirow{15}{*}{\rotatebox[origin=c]{90}{\textbf{DeepSeek-R1-Distill-Qwen-14B}}}
    & \textbf{BF16} & - & - & \textbf{54.7$_{\pm 0.5}$} & \textbf{95.5$_{\pm 0.4}$} & \textbf{93.7$_{\pm 0.2}$} & \textbf{62.6$_{\pm 1.8}$} & \textbf{50.9$_{\pm 0.8}$} & \textbf{71.5$_{\pm 0.2}$} & - \\
    \cline{2-11}
    & \multirow{4}{*}{\textbf{\textit{Weight-only}}} &  &  AWQ & 53.9$_{\pm 1.7}$ & 94.7$_{\pm 0.8}$ & 93.3$_{\pm 0.6}$ & 60.6$_{\pm 1.0}$ & 49.6$_{\pm 0.8}$ & 70.4$_{\pm 0.2}$ & \cellcolor{orange!10}{\footnotesize -1.1} \\  
    &  & \multirow{-2}{*}{4-16-16} & GPTQ & 52.5$_{\pm 2.2}$ & 94.9$_{\pm 0.3}$ & 93.7$_{\pm 0.3}$ & 60.3$_{\pm 2.0}$ & 48.3$_{\pm 0.6}$ & 69.9$_{\pm 0.3}$ & \cellcolor{orange!10}{\footnotesize -1.6} \\  
    \cline{3-11}
    &  & \multirow{2}{*}{3-16-16} & AWQ & 42.2$_{\pm 1.3}$ & 94.1$_{\pm 0.6}$ & 93.0$_{\pm 0.4}$ & 53.7$_{\pm 2.3}$ & 43.3$_{\pm 1.0}$ & 65.3$_{\pm 0.7}$ & \cellcolor{red!10}{\footnotesize -6.2}  \\  
    &  &  & GPTQ & 47.2$_{\pm 4.7}$ & 94.1$_{\pm 0.2}$ & 93.5$_{\pm 0.4}$ & 55.2$_{\pm 1.5}$ & 44.8$_{\pm 2.0}$ & 67.0$_{\pm 1.5}$ & \cellcolor{red!10}{\footnotesize -4.5} \\
    \cline{2-11}
    & \multirow{4}{*}{\textbf{\textit{KV Cache}}} &  &  KVQuant$^*$ & 55.3$_{\pm 4.2}$ & 94.7$_{\pm 0.6}$ & 93.8$_{\pm 0.2}$ & 61.3$_{\pm 1.5}$ & 49.6$_{\pm 0.4}$ & 70.9$_{\pm 0.8}$ & \cellcolor{green!10}{\footnotesize -0.6} \\  
    &  & \multirow{-2}{*}{16-16-4} & QuaRot & 54.7$_{\pm 3.2}$ & 95.0$_{\pm 0.5}$ & 93.8$_{\pm 0.3}$ & 60.9$_{\pm 1.9}$ & 52.6$_{\pm 1.5}$ & 71.4$_{\pm 0.8}$ & \cellcolor{green!10}{\footnotesize -0.1} \\  
    \cline{3-11}
    &  &  &  KVQuant$^*$ & 42.8$_{\pm 3.2}$ & 93.2$_{\pm 0.7}$ & 93.1$_{\pm 0.3}$ & 56.1$_{\pm 1.3}$ & 45.5$_{\pm 1.3}$ & 66.1$_{\pm 0.7}$ & \cellcolor{red!10}{\footnotesize -5.4} \\  
    &  & \multirow{-2}{*}{16-16-3} & QuaRot & 52.8$_{\pm 3.2}$ & 95.1$_{\pm 0.5}$ & 93.5$_{\pm 0.1}$ & 60.1$_{\pm 1.8}$ & 44.2$_{\pm 2.7}$ & 69.1$_{\pm 0.9}$ & \cellcolor{orange!10}{\footnotesize -2.4} \\  
    \cline{2-11}
    & \multirow{6}{*}{\textbf{\textit{Weight-Act.}}} &  & SmoothQuant & 59.4$_{\pm 1.3}$ & 95.5$_{\pm 0.5}$ & 94.0$_{\pm 0.4}$ & 60.9$_{\pm 2.4}$ & 50.6$_{\pm 1.8}$ & 72.1$_{\pm 0.5}$ & \cellcolor{green!10}{\footnotesize 0.6} \\   
    &  & 8-8-8 & QuaRot & 56.7$_{\pm 2.5}$ & 95.2$_{\pm 0.7}$ & 94.0$_{\pm 0.3}$ & 58.8$_{\pm 3.0}$ & 51.0$_{\pm 0.6}$ & 71.1$_{\pm 0.2}$ & \cellcolor{green!10}{\footnotesize -0.4} \\  
    &  &  & FlatQuant & 57.5$_{\pm 0.8}$ & 95.1$_{\pm 0.1}$ & 93.8$_{\pm 0.2}$ & 61.3$_{\pm 3.8}$ & 50.8$_{\pm 1.0}$ & 71.7$_{\pm 0.7}$ & \cellcolor{green!10}{\footnotesize 0.2} \\  
    \cline{3-11}
    &  &  & MXFP4 & 23.3$_{\pm 4.2}$ & 87.5$_{\pm 0.8}$ & 91.3$_{\pm 0.4}$ & 47.5$_{\pm 3.6}$ & 32.8$_{\pm 2.0}$ & 56.5$_{\pm 1.1}$ & \cellcolor{red!10}{\footnotesize -15.0} \\  
    &  & 4-4-4 & QuaRot & 42.5$_{\pm 4.4}$ & 92.7$_{\pm 0.9}$ & 93.0$_{\pm 0.3}$ & 55.7$_{\pm 3.0}$ & 45.7$_{\pm 1.3}$ & 65.9$_{\pm 0.5}$ & \cellcolor{red!10}{\footnotesize -5.6} \\  
    &  &  & FlatQuant & 50.3$_{\pm 2.4}$ & 94.7$_{\pm 0.3}$ & 93.3$_{\pm 0.1}$ & 55.7$_{\pm 2.0}$ & 48.4$_{\pm 1.2}$ & 68.5$_{\pm 0.4}$ & \cellcolor{red!10}{\footnotesize -3.0} \\
    \cline{3-11}
    \hline\hline
    %%%%%%%%%%%%%%%%%%%%%%%%%%%%%%%%%%%%%%%%%%%%%%%%%%%%%%%%%%%%%%%%%%%%%%%%%%%%%%%%%%%%%%
    % DeepSeek-R1-Distill-Qwen-32B
    \multirow{15}{*}{\rotatebox[origin=c]{90}{\textbf{DeepSeek-R1-Distill-Qwen-32B}}}
    & \textbf{BF16} & - & - & \textbf{61.7$_{\pm 1.7}$} & \textbf{96.3$_{\pm 0.5}$} & \textbf{94.2$_{\pm 0.2}$} & \textbf{65.7$_{\pm 1.8}$} & \textbf{56.0$_{\pm 2.3}$} & \textbf{74.8$_{\pm 0.4}$} & - \\
    \cline{2-11}
    & \multirow{4}{*}{\textbf{\textit{Weight-only}}} &  &  AWQ & 63.6$_{\pm 1.3}$ & 95.9$_{\pm 0.6}$ & 94.4$_{\pm 0.2}$ & 63.5$_{\pm 1.9}$ & 54.6$_{\pm 1.5}$ & 74.4$_{\pm 0.6}$ & \cellcolor{green!10}{\footnotesize -0.4} \\  
    &  & \multirow{-2}{*}{4-16-16} & GPTQ & 57.5$_{\pm 1.4}$ & 95.9$_{\pm 0.8}$ & 94.3$_{\pm 0.1}$ & 62.3$_{\pm 1.5}$ & 54.2$_{\pm 0.4}$ & 72.9$_{\pm 0.5}$ & \cellcolor{orange!10}{\footnotesize -1.9} \\  
    \cline{3-11}
    &  &  &  AWQ  & 53.1$_{\pm 1.7}$ & 94.1$_{\pm 1.1}$ & 94.1$_{\pm 0.1}$ & 62.6$_{\pm 1.5}$ & 50.4$_{\pm 1.6}$ & 70.9$_{\pm 1.0}$ & \cellcolor{red!10}{\footnotesize -3.9} \\  
    &  & \multirow{-2}{*}{3-16-16} & GPTQ & 51.7$_{\pm 4.4}$ & 94.7$_{\pm 0.3}$ & 94.1$_{\pm 0.2}$ & 57.6$_{\pm 2.7}$ & 50.1$_{\pm 1.5}$ & 69.6$_{\pm 1.0}$ & \cellcolor{red!10}{\footnotesize -5.1} \\
    \cline{2-11}
    & \multirow{4}{*}{\textbf{\textit{KV Cache}}} &  & KVQuant$^*$ & 57.5$_{\pm 1.4}$ & 96.0$_{\pm 0.4}$ & 94.7$_{\pm 0.4}$ & 64.3$_{\pm 1.9}$ & 55.9$_{\pm 0.8}$ & 73.7$_{\pm 0.5}$ & \cellcolor{orange!10}{\footnotesize -1.1} \\  
    &  & \multirow{-2}{*}{16-16-4} & QuaRot & 63.1$_{\pm 1.9}$ & 95.8$_{\pm 0.2}$ & 94.2$_{\pm 0.4}$ & 63.1$_{\pm 1.8}$ & 56.3$_{\pm 1.1}$ & 74.5$_{\pm 0.9}$ & \cellcolor{green!10}{\footnotesize -0.3} \\  
    \cline{3-11}
    &  &  & KVQuant$^*$ & 50.8$_{\pm 4.2}$ & 94.8$_{\pm 0.9}$ & 94.4$_{\pm 0.1}$ & 62.3$_{\pm 2.1}$ & 51.6$_{\pm 0.6}$ & 70.8$_{\pm 0.8}$ & \cellcolor{red!10}{\footnotesize -4.0} \\  
    &  &\multirow{-2}{*}{16-16-3}  & QuaRot & 57.2$_{\pm 2.7}$ & 94.9$_{\pm 0.5}$ & 94.1$_{\pm 0.5}$ & 63.8$_{\pm 1.1}$ & 48.6$_{\pm 2.3}$ & 71.7$_{\pm 0.8}$ & \cellcolor{red!10}{\footnotesize -3.0} \\  
    \cline{2-11}
    & \multirow{6}{*}{\textbf{\textit{Weight-Act.}}} &  & SmoothQuant & 59.2$_{\pm 2.2}$ & 95.4$_{\pm 0.4}$ & 94.2$_{\pm 0.3}$ & 64.0$_{\pm 1.2}$ & 56.7$_{\pm 1.7}$ & 73.9$_{\pm 0.5}$ & \cellcolor{green!10}{\footnotesize -0.9} \\   
    &  & 8-8-8 & QuaRot & 56.9$_{\pm 2.6}$ & 95.3$_{\pm 0.3}$ & 94.3$_{\pm 0.3}$ & 63.0$_{\pm 0.8}$ & 55.5$_{\pm 1.7}$ & 73.0$_{\pm 0.7}$ & \cellcolor{orange!10}{\footnotesize -1.7} \\  
    &  &  & FlatQuant & 62.2$_{\pm 3.4}$ & 95.9$_{\pm 0.6}$ & 94.4$_{\pm 0.3}$ & 65.2$_{\pm 0.5}$ & 56.1$_{\pm 0.6}$ & 74.7$_{\pm 0.8}$ & \cellcolor{green!10}{\footnotesize -0.0} \\  
    \cline{3-11}
    &  &  & MXFP4 & 35.3$_{\pm 3.5}$ & 92.4$_{\pm 0.4}$ & 93.7$_{\pm 0.6}$ & 57.1$_{\pm 3.6}$ & 36.2$_{\pm 0.7}$ & 62.9$_{\pm 0.7}$ & \cellcolor{red!10}{\footnotesize -11.8} \\  
    &  & 4-4-4 & QuaRot & 46.9$_{\pm 2.1}$ & 94.1$_{\pm 0.4}$ & 93.9$_{\pm 0.2}$ & 60.1$_{\pm 2.2}$ & 49.4$_{\pm 0.6}$ & 68.9$_{\pm 0.3}$ & \cellcolor{red!10}{\footnotesize -5.9} \\  
    &  &  & FlatQuant & 57.8$_{\pm 4.3}$ & 95.1$_{\pm 0.6}$ & 94.2$_{\pm 0.0}$ & 58.1$_{\pm 2.2}$ & 54.2$_{\pm 1.3}$ & 71.9$_{\pm 1.6}$ & \cellcolor{orange!10}{\footnotesize -2.9} \\
    \cline{3-11}
    \hline\hline
    %%%%%%%%%%%%%%%%%%%%%%%%%%%%%%%%%%%%%%%%%%%%%%%%%%%%%%%%%%%%%%%%%%%%%%%%%%%%%%%%%%%%%%
    \end{tabular}
    }
    \end{center}
    \caption{
    The overall assessment of quantized DeepSeek-R1-Distill-Qwen models on various reasoning benchmarks. QuaRot for the 7B model weight-activation quantization is skipped because of the incompatible hidden dimension for Hadamard transformation.
    The \textcolor[HTML]{009900}{green}, \textcolor{orange}{orange} and \textcolor{red}{red} cells stand for the lossless (\textcolor[HTML]{009900}{$\leq$1\%}), the fair (\textcolor{orange}{1\%-3\%}) and the risky (\textcolor{red}{$\geq$3\%}) respectively.
    Note that 1.5B and 7B models are exceptions with severe degradation for low-bit KV cache and weight-activation quantization, as will be discussed in Appendix~\ref{sec:apdx_1.5b_7b}.
    Results on more models are available in Appendix~\ref{sec:apdx-exp}.}
\vspace{-2ex}
\label{tab:r1-qwen}
\end{table*}

\subsection{Lossless Quantization Bit-width}
\label{sec:rq1}

The overall results for the DeepSeek-R1-Distill-Qwen models are summarized in Table~\ref{tab:r1-qwen}. 
Results on more models (including DeepSeek-R1-Distill-LLaMA, QwQ-32B, and Qwen3-8B) can be found in Appendix~\ref{sec:apdx_more_llms}.
For ease of interpretation, we categorize the performance degradation into three classes: lossless (\textcolor[HTML]{009900}{$\leq$1\%}), fair (\textcolor{orange}{$1\%-3\%$}), and risky (\textcolor{red}{$\geq$3\%}). 
We draw the following conclusions based on best-performing quantization algorithms, and leave the algorithm comparisons in \S~\ref{sec:rq2}.

\paragraph{For weight-only quantization, 4-bit is nearly lossless, but 3-bit becomes risky.}
From Table~\ref{tab:r1-qwen}, quantizing the weights of the distilled Qwen models to $4$ bits incurs a slight performance drop compared to their BF16 counterparts, i.e., 2.1\% and 0.4\% for the 1.5B and 32B models, respectively.
% we compare 4-bit and 3-bit weight-only quantization on the \sout{DeepSeek-R1-Distill-Qwen-32B models}. 
% However, the performance drop of the distilled LLaMA-8B model is much larger, even with the better-performed GPTQ algorithm. We speculate this is because
% \footnote{\hou{any explanation for the varied behavior of different models?}}
% As can be seen, AWQ is lossless in 4 bit weight-only quantization with groupsize = 128, while GPTQ loss $4.4\%$. 
However, when reducing to 3 bits, there is a sharp drop in performance for each model, e.g., over 7\% accuracy drop for the 1.5B model and 3\% for the 32B model. 
Meanwhile, smaller reasoning models tend to suffer more from quantization, which aligns past experiences on non-reasoning models~\citep{li2024evaluating}.
Similar observations can be found on other sizes and families of LLMs in Appendix~\ref{sec:apdx_more_llms}.

\paragraph{For KV cache, we suggest 4-bit quantization.}
% 4-bit is generally fair or lossless, while 3-bit can be risky for smaller LLMs.
We find that 4-bit quantized KV cache achieves lossless performance on large models (e.g., 14B or 32B). 
% performance drop across varying sizes of DeepSeek-R1-Distill Qwen models. 
In particular, the accuracy drop can remain smaller than 1\% even on AIME-120, where the model generates lengthy output of over 10K tokens, as will be discussed in \S\ref{sec:output_length}.
In addition,  the performance of smaller LLMs is also relatively fair, with an average drop of 1.4\% on 1.5B and 1.0\% on 7B, respectively. 
However, for more aggressive KV cache quantization, the performance incurs a sharp drop, especially for smaller LLMs, i.e., smaller models (1.5B and 7B) and larger models (14B and 32B) incur over 5\% and 2\% accuracy drop when the bitwidth of KV cache quantization reduces to 3 bits, respectively.

\paragraph{We recommend 8-bit quantization for weights, activation, and KV cache, which is lossless across various LLM sizes and reasoning tasks.}
It can be observed that for all reasoning models and tasks evaluated, the best-performing W8A8KV8 
quantization algorithm achieves a performance drop of less than 1 point.
% less than a performance drop of 1 point. 
This holds true even for the smallest DeepSeek-R1-Distill-Qwen-1.5B model. 
Nevertheless, when we apply more aggressive quantization with 4 bits, even the large 32B model incurs an accuracy drop of 2.9\%. 
The degradation becomes significantly more pronounced for smaller LLMs, with accuracy drops exceeding 10\% on the 1.5B and 7B models.
% The degradation can be much riskier on smaller LLMs, e.g., over 10\% accuracy drop on the 1.5B and 7B models. 
Thus, improving 4-bit weight-activation quantization for reasoning models remains an open challenge for the community.

\subsection{Comparisons of Quantization Algorithms}
\label{sec:rq2}

\paragraph{AWQ is preferred over GPTQ for weight quantization, considering its comparable performance but more efficient implementation.}
% with comparable accuracy.}
% \footnote{do we use --act-order for GPTQ?}}
According to Table~\ref{tab:r1-qwen}, AWQ performs on par with GPTQ for both 4-bit and 3-bit quantization. 
For instance, it surpassed GPTQ three times and lost once for 4-bit quantization over the evaluated models. In addition, AWQ is usually faster to implement compared with GPTQ, since it does not require the iterative update of LLM parameters. Besides, AWQ is also more robust to the choice of calibration data than GPTQ, as discussed in Section~\ref{sec:discussion_calib}. We thus recommend AWQ as the default algorithm for the weight-only quantization of reasoning models.

\textbf{QuaRot is generally preferred over KVQuant$^*$ for KV cache Quantization, except for Qwen 1.5B and 7B models which have unexpected huge outliers over bias.} 
According to Table~\ref{tab:r1-qwen}, QuaRot generally outperforms KVQuant$^*$ on the 14B and 32B models. 
For instance, on the DeepSeek-R1-Distill-Qwen-32B model, QuaRot with 3-bit quantization achieves an average performance of 71.7\%, while KVQuant$^*$ only reaches 70.8\%. 
% This trend is consistent across both QwQ and LLaMA models, as detailed in Appendix~\ref{sec:apdx_more_llms}. 
However, exceptions are observed in the 1.5B and 7B models, where KVQuant$^*$ significantly surpasses QuaRot. 
We find that this is due to the huge outliers over biases in the key and value layers of these two models. 
While KVQuant$^*$ can mitigate this by quantizing the output before bias, it is not applicable for QuaRot. 
More discussions can be found in Appendix~\ref{sec:apdx_1.5b_7b}.
 % with an average accuracy of 47.41\% and 61.59\%, 
% We find such performance gaps can be largely attributed to huge outliers over biases of key and value layers on the 1.5B and 7B LLMs. 

% In the DeepSeek-R1-Distill-Qwen-32B model with the same 16-16-4 configuration, KVQuant reaches an average performance of 74.51, with almost no performance drop (\cellcolor{green!10}{\footnotesize 0.02}). These results indicate that KVQuant can effectively maintain model performance for KV cache quantization, making it a suitable choice.

\textbf{For 8-bit weight-activation quantization, SmoothQuant proves sufficient, while FlatQuant emerges as the leading algorithm for 4-bit quantization.} In the 8-bit scenario, all algorithms achieve near-lossless results across most models, with no clear leading algorithm. We therefore recommend SmoothQuant for 8-bit quantization due to its zero-overhead advantage. However, in the more challenging 4-bit setting, FlatQuant demonstrates significant superiority over competing algorithms while introducing only minimal latency overhead. For instance, FlatQuant achieves an average performance of 71.9\% on the 32B model, with a performance drop of only 2.9\%, significantly surpassing both MXFP4 (11.8\%$\downarrow$) and QuaRot (5.9\%$\downarrow$).
Note that substantial accuracy loss still exists under W4A4KV4 quantization on 1.5B and 7B models even for FlatQuant. We analyze its reason in Appendix~\ref{sec:apdx_1.5b_7b}.

% \textbf{FlatQuant is the leading algorithm for weight-activation quantization.} For all 8 settings with varying model sizes and quantization configurations, FlatQuant outperforms the other algorithms in 7 cases. While FlatQuant does not exhibit clear advantages on W8A8KV8 quantization, it significantly outperforms the other algorithms on W4A4KV4. For instance, FlatQuant achieves an average performance of 71.9\% on the 32B model, with a performance drop of only 2.9\%, significantly surpassing both MXFP4 (11.8\%$\downarrow$) and QuaRot (5.9\%$\downarrow$).
% Note that substantial accuracy loss still exists under W4A4KV4 quantization on 1.5B and 7B models even for FlatQuant. We analyze its reason in Appendix~\ref{sec:apdx_1.5b_7b}.

In the rest of this study, we adopt AWQ for weight-only quantization, QuaRot for KV cache quantization, and FlatQuant for weight-activation quantization if not otherwise specified.

\subsection{Impacts of Task Difficulty}
\label{sec:rq3}

\textbf{Quantization is prone to fail on more difficult tasks.}
Based on Table~\ref{tab:r1-qwen}, we compare the performance drop caused by quantization on the three math reasoning benchmarks with varying difficulty levels, and find that more difficult tasks incur a more severe performance drop.
Among them, AIME-120 is the most challenging task, with samples selected from American Math Competition questions. 
MATH-500  contains a mix of easy and hard problems, followed by GSM8K with easy primary school-level questions. 
The performance degradations are in descending order among the three benchmarks. 
For instance, for the 32B model with W4A4KV4 quantization, the performance drops are 3.9\% on AIME-120, 1.2\% on MATH-500, and 0.0\% on GSM8K.

In addition to Table~\ref{tab:r1-qwen}, we also examine the accuracy at the 5 different difficulty levels explicitly defined by MATH-500~\citep{lightman2023let} in Appendix~\ref{apdx:task_difficulty}, where the observations remain consistent.

\subsection{The Impacts of the Origin of Reasoning LLMs}
\label{sec:rq4}
In this section, we compare the quantization performance on reasoning models trained from different methods, with the suggested algorithms in \S~\ref{sec:rq2}.
% of different LLM origins on quantization. 
% \hou{consider differences during both the pre-training and post-training stages. For the difference in the pre-training stage,}
Specifically, we study 
\begin{enumerate}[leftmargin=4ex]
\item the reasoning models obtained with either distillation~(i.e., DeepSeek-R1-Distill-Qwen-32B) and reinforcement learning (i.e., QwQ-32B); and
\item different LLM families (i.e., LLaMA and Qwen) distilled from DeepSeek-R1.
\end{enumerate}

\vspace{-2ex}

% \paragraph{RL-based LLMs are more error-prone to quantization.} 
% Comparing DeepSeek-R1-Distill-Qwen-32B and QwQ-32B in Figure~\ref{fig:llm_origin} (a)-(c), we find that QwQ-32B generally suffers greater performance drops across reasoning tasks and quantization configurations. 
% For example, the accuracy of QwQ drops by 14.98\% for W3G128 and 16.66\% for KV3 on AIME-120, higher than Qwen's 9.98\% and 8.32\%.
% The exception is KV cache quantization on MATH-500, where no significant damage is observed. 
% We speculate because distilled models learn generalizable reasoning abilities from a stronger reasoner model, resulting in more modular and stable representations robust to compression. 
% In contrast, RL-based models learn highly specialized policies, making them vulnerable to representation perturbations.

% 6.4\% 5.4\%; 4.0\% 7.1\%; 3.6\% 2.5\%

\begin{figure*}[t]
\centering
\subfigure[Weight-only Quant.]{\includegraphics[width=.32\textwidth]{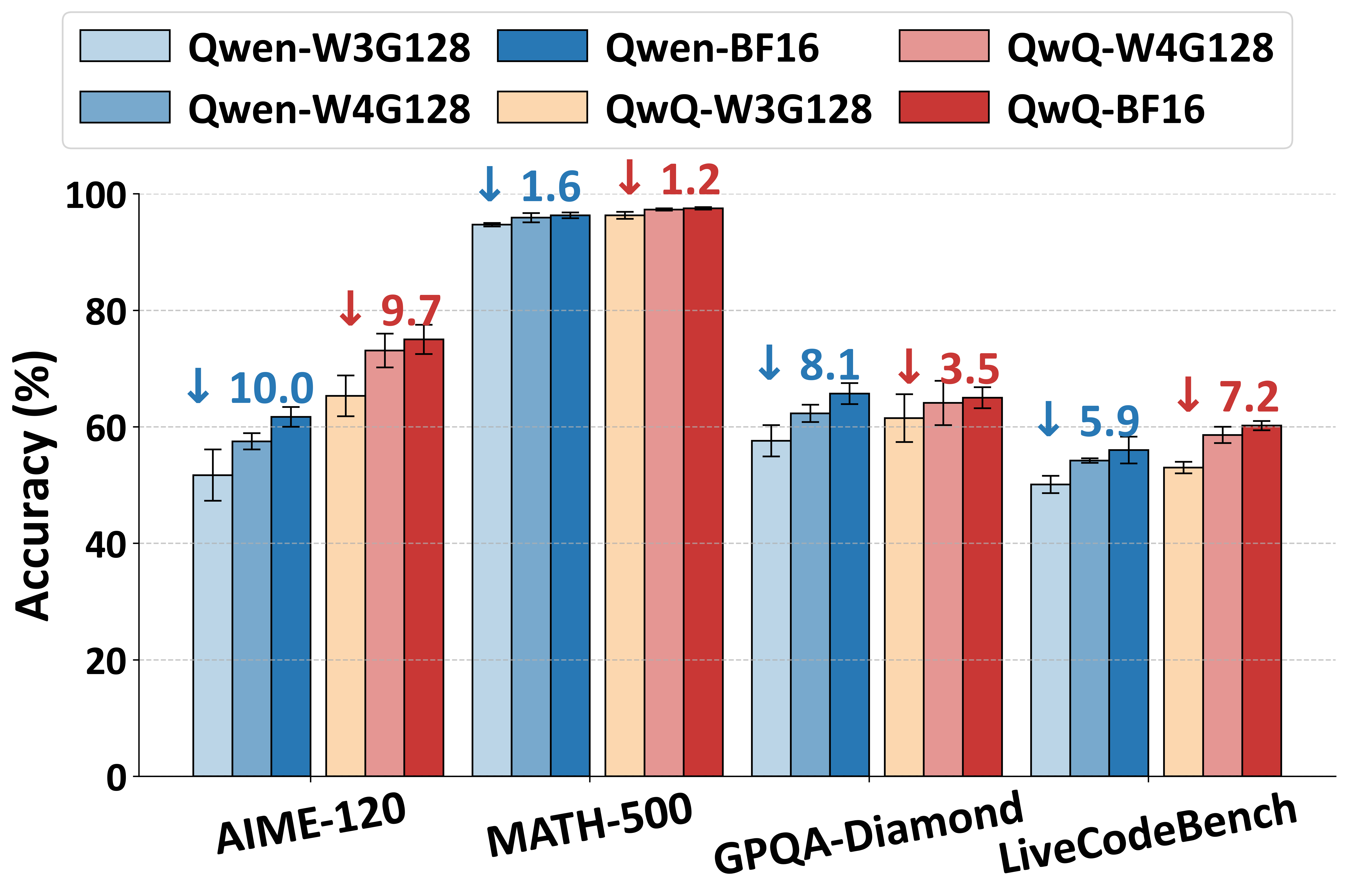}}
% \hspace{-8pt}
\subfigure[KV cache Quant.]{\includegraphics[width=.32\textwidth]{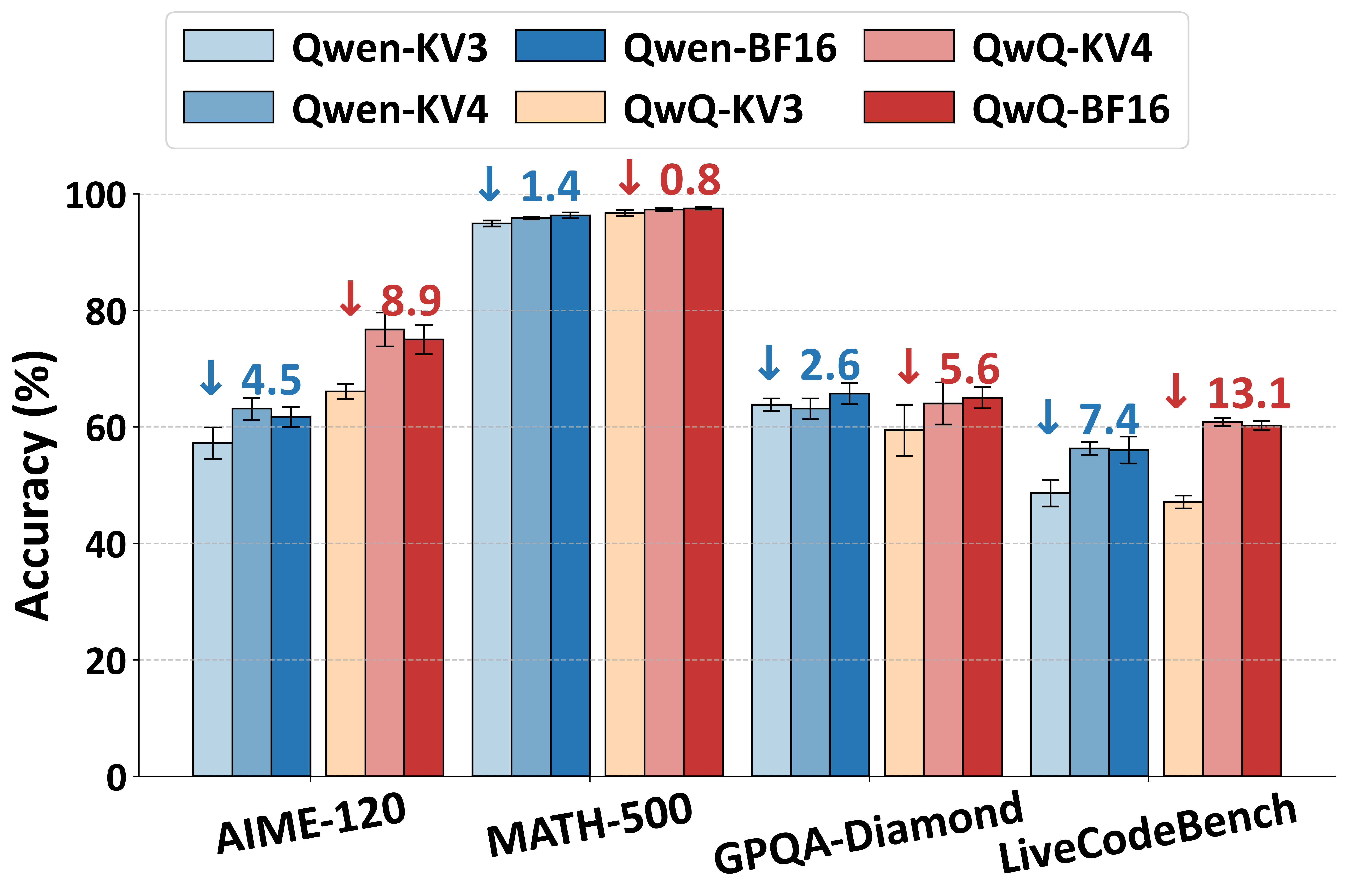}}
% \hspace{-8pt}
\subfigure[Weight-Act. Quant.]{\includegraphics[width=.32\textwidth]{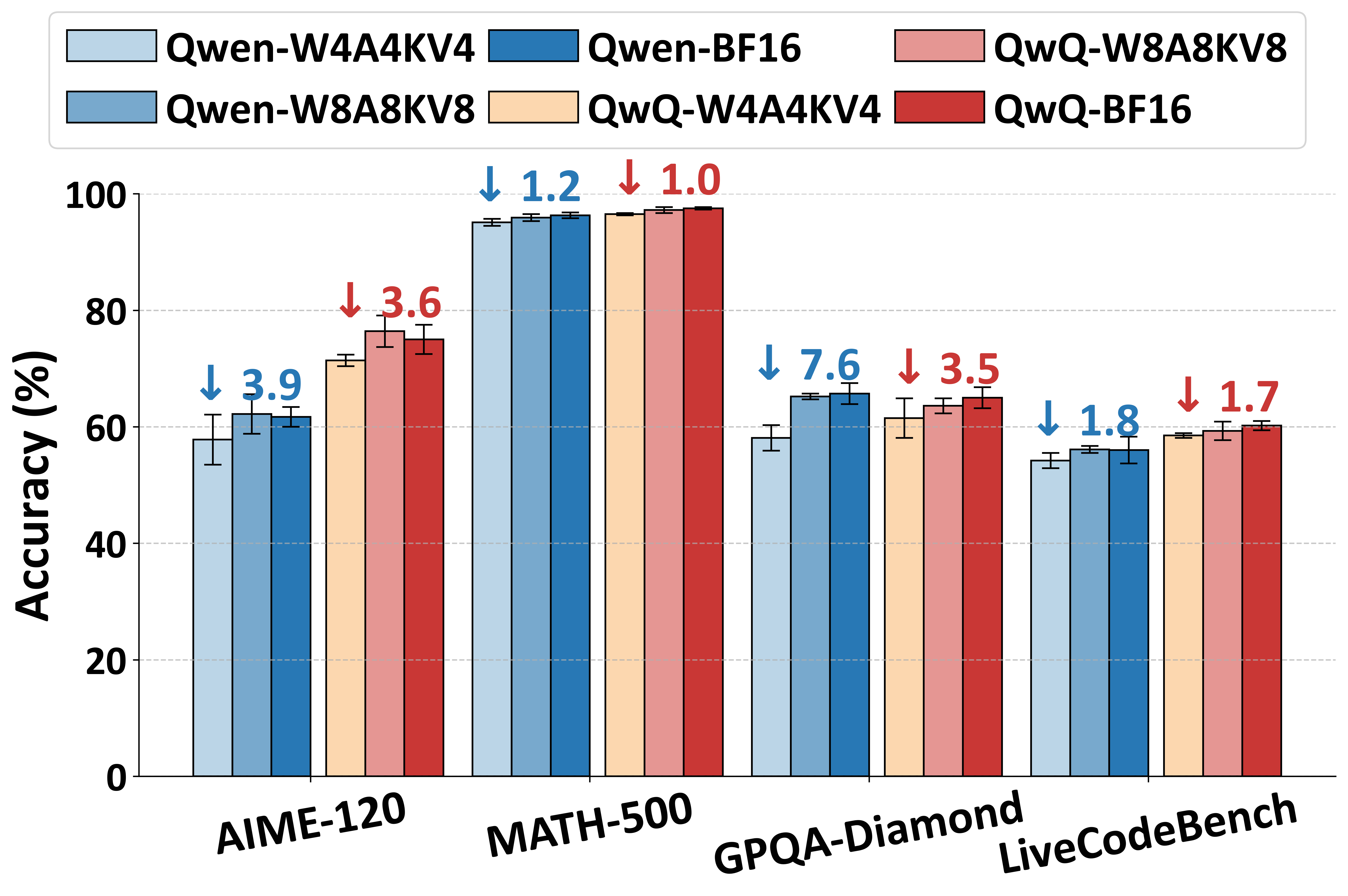}}
\subfigure[Weight-only Quant.]{\includegraphics[width=.32\textwidth]{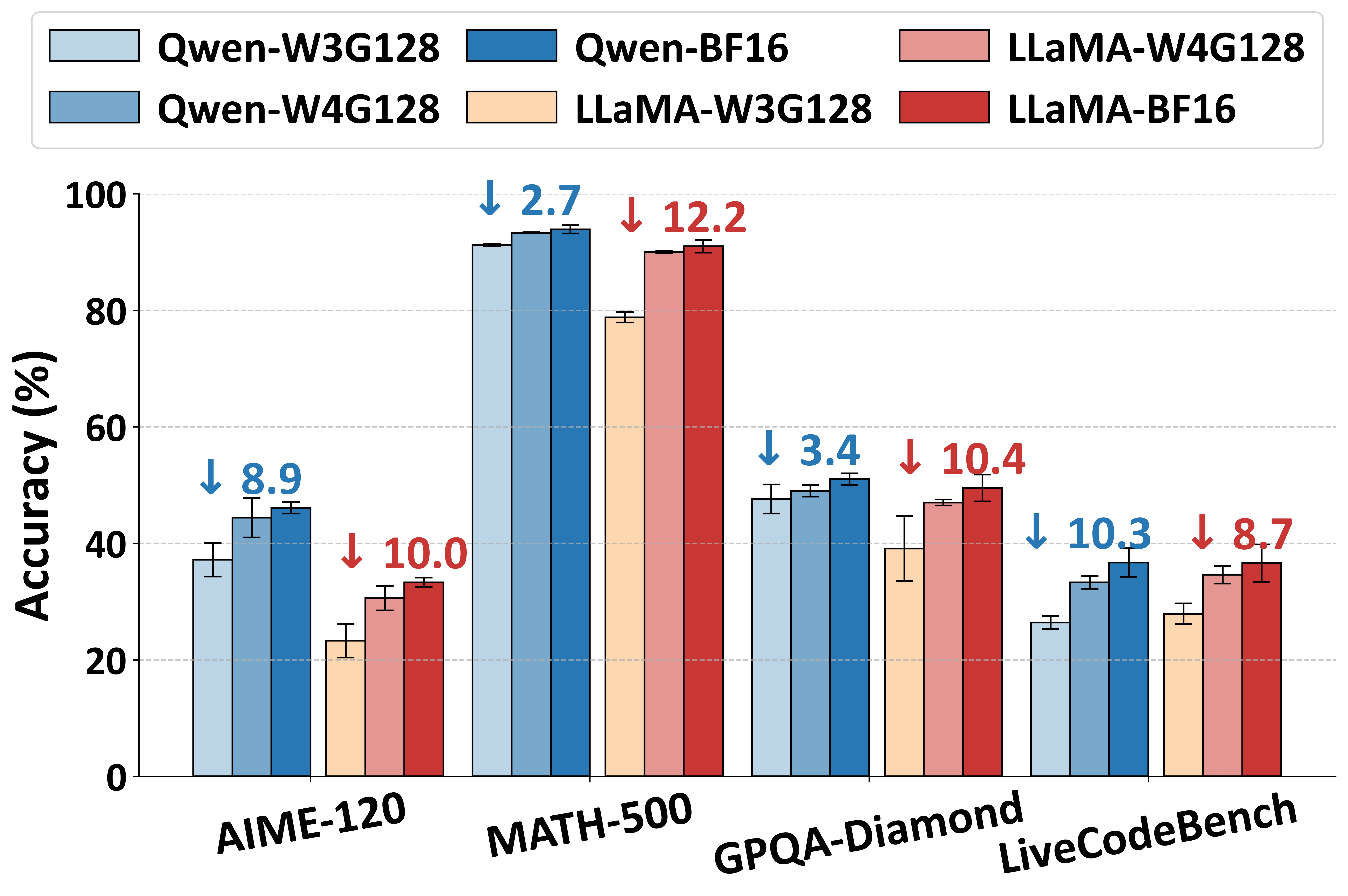}}
% \hspace{-8pt}
\subfigure[KV cache Quant.]{\includegraphics[width=.32\textwidth]{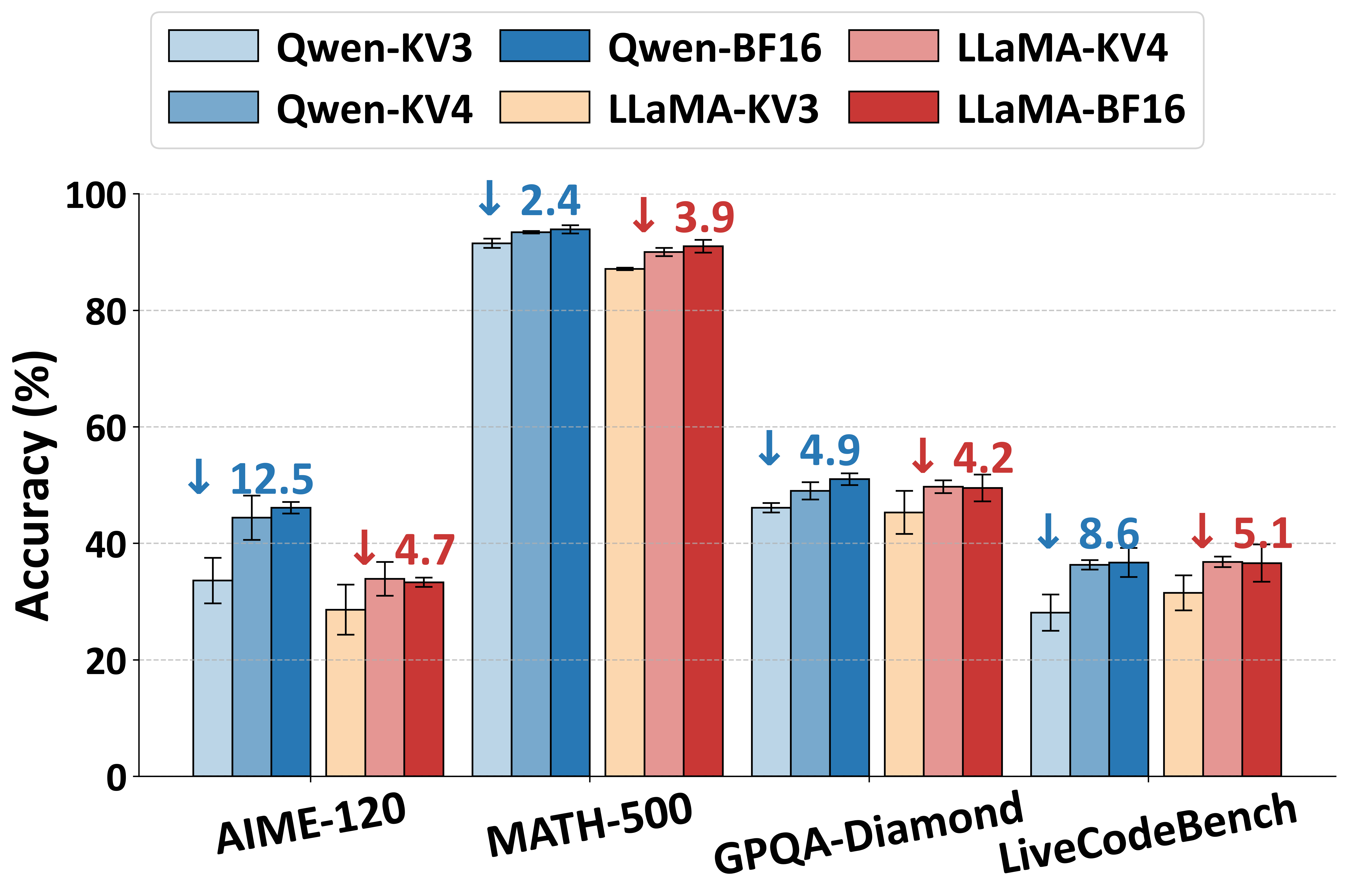}}
% \hspace{-8pt}
\subfigure[Weight-Act. Quant.]{\includegraphics[width=.32\textwidth]{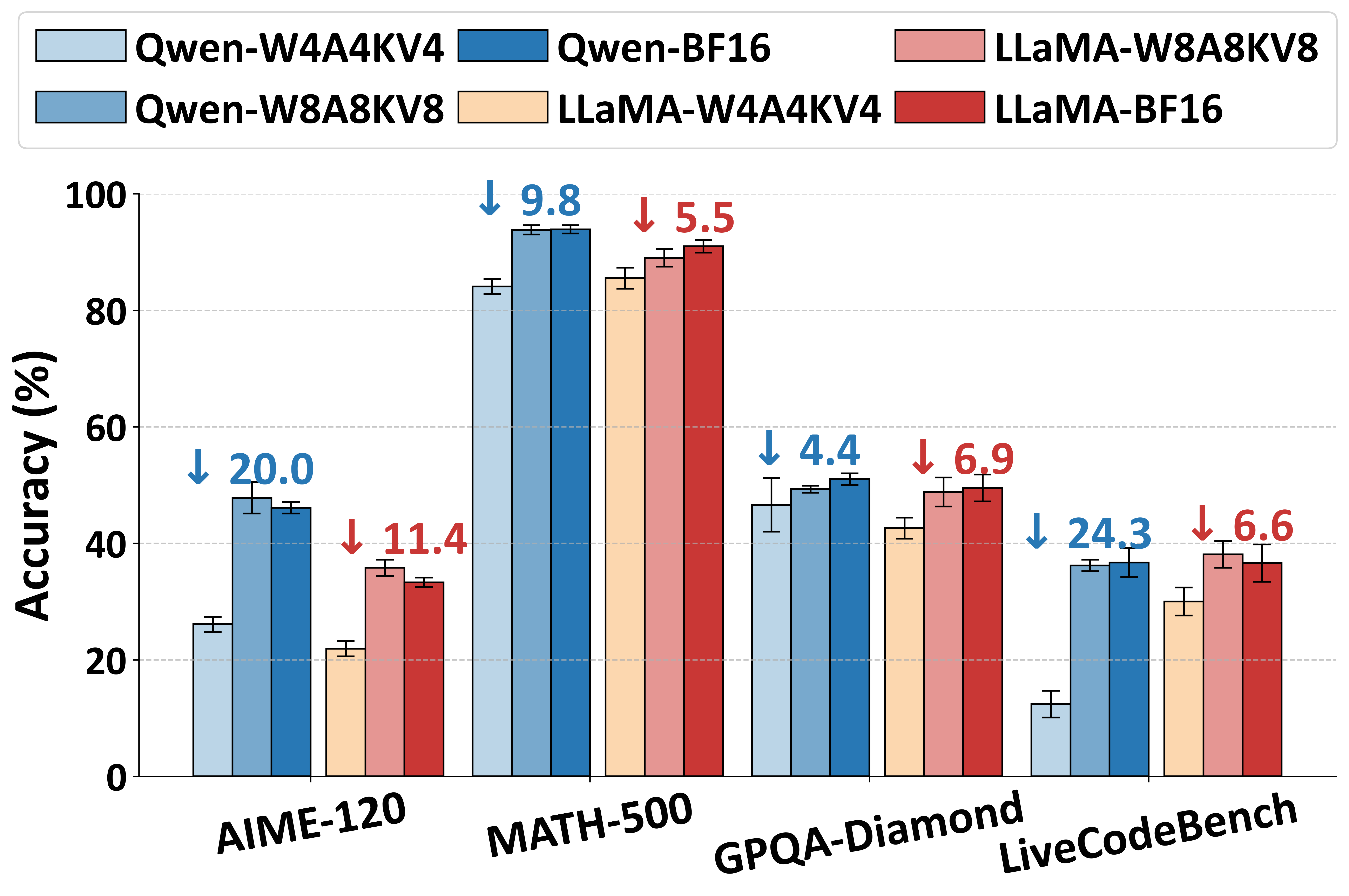}}
\caption{(a)-(c) shows the comparisons between DeepSeek-R1-Distill-Qwen-32B (SFT-based) and QwQ-32B (RL-based) on different quantization configurations. 
(d)-(f) shows comparisons between DeepSeek-R1-Distill-Qwen-7B and DeepSeek-R1-Distill-LLaMA-8B on different quantization configurations.}
\label{fig:llm_origin}
\end{figure*}

\paragraph{Qwen-32B is more robust to KV cache quantization, while QwQ-32B is more resilient to weight-only and weight-activation quantization.}
As seen in Figure~\ref{fig:llm_origin} (a)-(c), when comparing Qwen-32B and QwQ-32B, we observe that the models exhibit varying tolerance to different quantization configurations. For instance, Qwen-32B experiences a larger accuracy drop of 6.4\% for W3G128 and 3.8\% for KV3 compared to QwQ-32B’s 5.4\% and 2.5\%, respectively. However, Qwen-32B gains an edge in KV cache quantization, with 3.1\% less quantization loss.
We hypothesize that although both models originate from the same base model, differences in their training recipes lead to distinct training dynamics, shaping their intermediate representations differently and ultimately influencing quantization accuracy.
% Further investigation is needed to better understand how training methods affect quantization performance.

\paragraph{Qwen-7B is more robust to weight-only quantization, while LLaMA-8B is more resilient to KV cache and weight-activation quantization.} 
To study the impact of LLM origins, we compared DeepSeek-R1-Distill-Qwen-7B and DeepSeek-R1-Distill-LLaMA-8B, which are of similar size and distilled from DeepSeek-R1 with the same data. 
As shown in Figure~\ref{fig:llm_origin} (d)-(f), Qwen-7B is more robust to weight-only quantization (e.g., 8.9\% vs. 10.0\% for W3G128 on AIME-120).
However, it is more sensitive for KV cache and activation quantization (e.g., 12.5\% vs. 4.7\% drop for KV3 on AIME-120, and 20.0\% vs. 11.4\% drop for W4A4KV4 on AIME-120). 
Potential reasons include differences in pretraining data, fine-tuning configurations, and training dynamics.

More detailed results of QwQ-32B and DeepSeek-R1-Distill-LLaMA for different quantization configurations and algorithms can be found in Appendix~\ref{sec:apdx_more_llms}.

\section{Discussions}
\label{sec:discussion}
In this section, we provide further analysis for quantized reasoning models, including their output length, choice of calibration data, scaling effects w.r.t both model size and latency, as well as test-time scaling. The analysis is based on the best-performing quantization algorithms according to Section~\ref{sec:rq2}.
Due to limited space, we provide more analysis on the choice of calibration data in Appendix~\ref{sec:apdx_discussion_calib},  quantization of DeepSeek-R1-Distill-Qwen-1.5B and 7B model in Appendix~\ref{sec:apdx_1.5b_7b}, and qualitative examples in Appendix~\ref{sec:apdx_qualitative}.

\subsection{Do Quantized Reasoning Models Think Longer?}
\label{sec:output_length}

\begin{figure*}[t]
\centering
\includegraphics[width=0.85\linewidth]{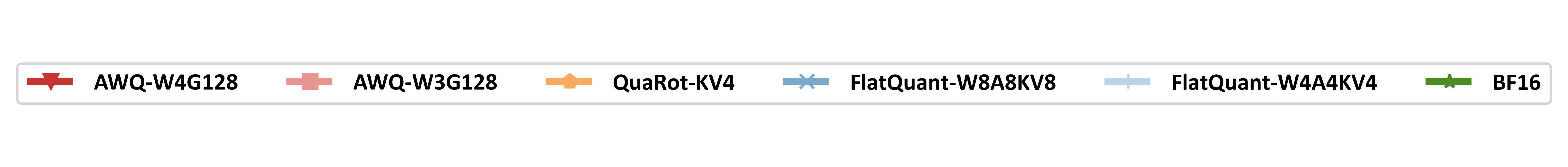}
\vspace{-2ex}
\subfigure[DS-R1-Distill-Qwen-1.5B.]{\includegraphics[width=.32\textwidth]{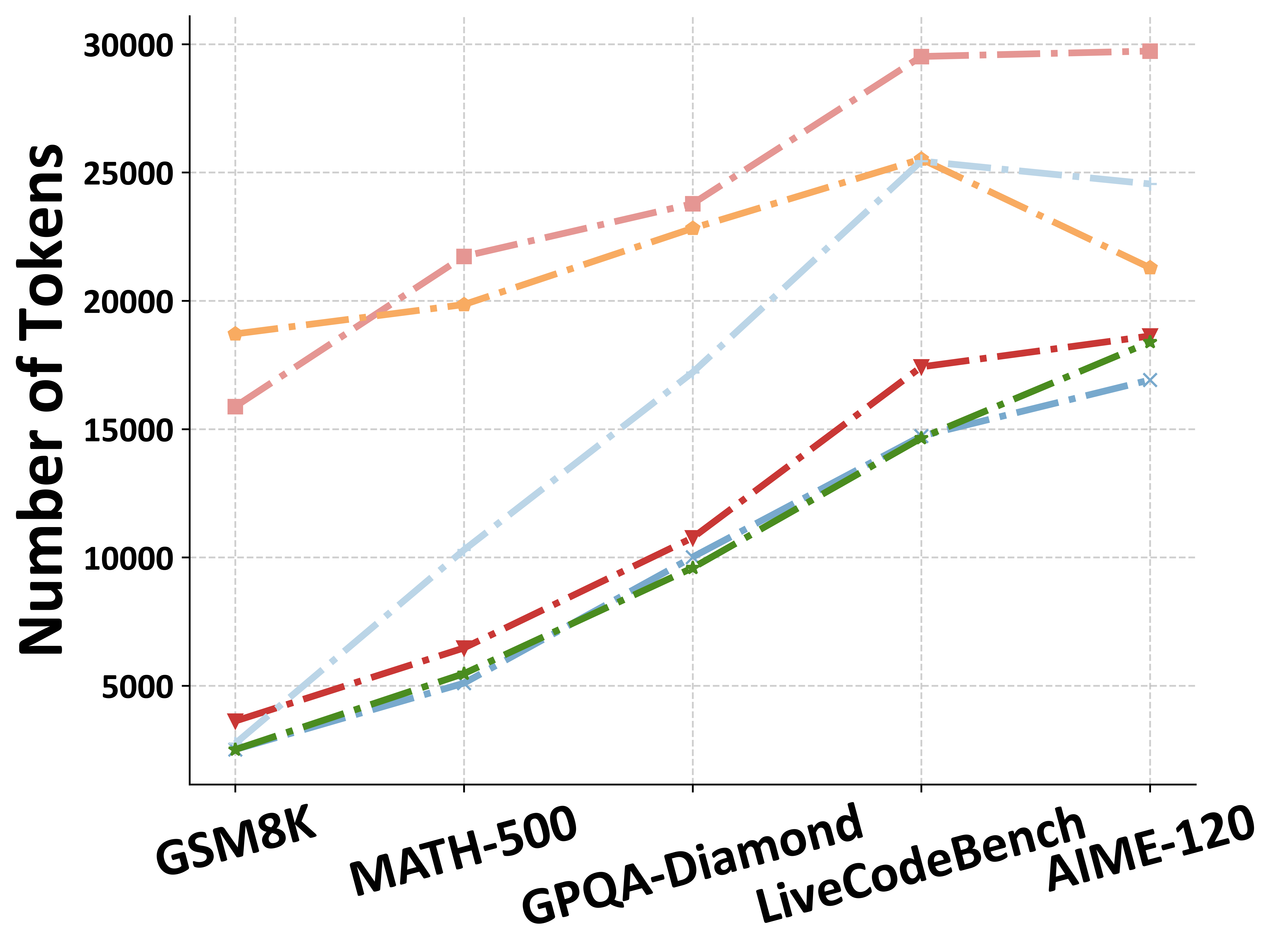}}
% \hspace{-8pt}
\subfigure[DS-R1-Distill-Qwen-14B.]{\includegraphics[width=.32\textwidth]{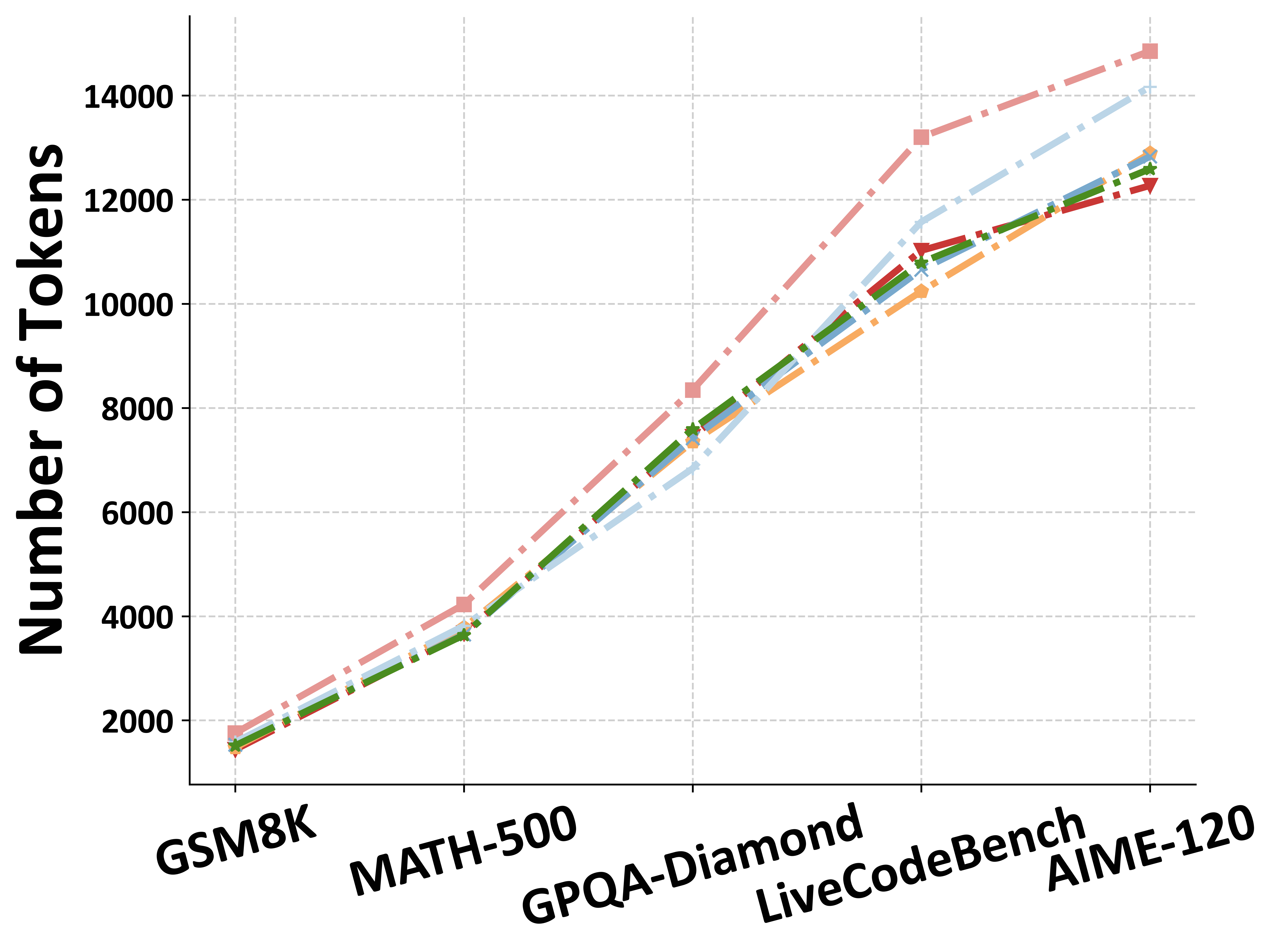}}
% \hspace{-8pt}
\subfigure[QwQ-32B.]{\includegraphics[width=.32\textwidth]{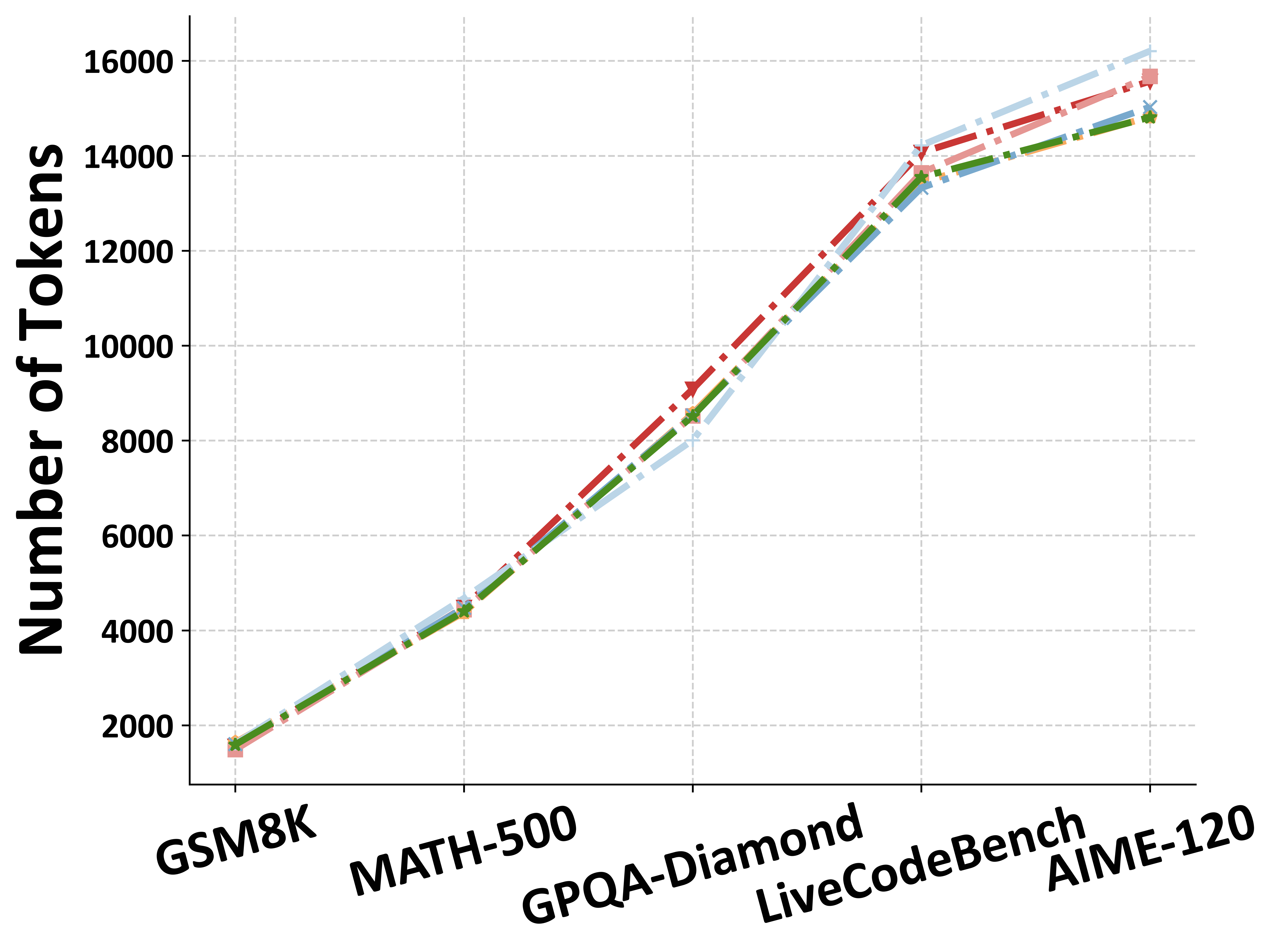}}
% \hspace{-8pt}
% \subfigure[DeepSeek-R1-Distill-Qwen-14B]{\includegraphics[width=.24\textwidth]{figs/num_tokens/DeepSeek-R1-Distill-Qwen-14B_notitle.png}}
% \vspace{-1ex}
\caption{The lengths of generated tokens across different datasets and models.}
\label{fig:num_t_all}
\end{figure*}

% \textbf{Quantized LLMs with minor performance drop do not exhibit longer output across varying task difficulty.} 
Since quantization inherently reduces model capacity due to precision loss, one concern with quantized LLMs is whether quantized models need to think longer to compensate for the reduced representation power, 
which could prolong the end-to-end latency. 
We analyze the output length statistics of quantized reasoning models in Figure~\ref{fig:num_t_all}. \textbf{It is found that for various quantization configurations (e.g., W4G128, W8A8KV8, and KV4), the generated output lengths closely align with those of the BF16 models. }
However, under more aggressive low-bit quantization (e.g., W3G128 and W4A4KV4), LLMs tend to produce longer outputs. 
This effect is particularly pronounced in smaller models like DeepSeek-R1-Distill-Qwen-1.5B, where lower-bit quantization results in more pronounced performance degradation, as previously discussed in Table~\ref{tab:r1-qwen}.
% We analyze the statistics of output lengths of quantized reasoning models in Figure~\ref{fig:num_t_all}. 
% For well-performing quantization configurations (e.g., W4G128, W8A8KV8, and KV4), the output lengths mostly overlap with those of BF16 models. However, under lower-bit quantizations (e.g., W3G128 and W4A4), LLMs tend to generate longer outputs, especially for smaller models like DeepSeek-R1-Distill-Qwen-1.5B, where quantization causes more significant performance degradation according to Table~\ref{tab:r1-qwen}.

While more difficult tasks tend to generate longer outputs, they do not inherently trigger longer reasoning steps in quantized LLMs. Similar observations are provided in Appendix~\ref{sec:apdx_length}, where we analyze the output lengths across five difficulty levels on MATH-500. 
% However, when quantization incurs a large performance drop, models may generate repetitive outputs that significantly increase the reasoning path, as seen in DeepSeek-R1-Distill-Qwen-1.5B with AWQ-W3G128 quantization. 

% \begin{figure}
%     \centering
%     \includegraphics[width=1.0\linewidth]{figs/num_tokens/QwQ-32B.png}
%     \caption{Number of tokens used by QwQ-32B and different quantized models across different hard level of tasks. }
%     \label{fig:num_t_qwq32}
% \end{figure}

 % Be consistent: W3G128 or W3g128. 

\begin{table*}[!t]
\begin{center}
\resizebox{1.00\linewidth}{!}
{
    \begin{tabular}{l|l|l|rrr|r|r|r|r}
    \hline\hline
    \textbf{Methods} & \textbf{\makecell{W-A-KV\\\# Bits}} & \textbf{\makecell{Calibration\\Domains}} & \textbf{\makecell{AIME-\\120}}  & \textbf{\makecell{MATH-\\500}} & \textbf{GSM8K} & \textbf{\makecell{GPQA-\\Diamond}} & \textbf{\makecell{LiveCode-\\Bench}} & \textbf{Avg.} & \textbf{$\Delta$} \\ 
    \hline
    %%%%%%%%%%%%%%%%%%%%%%%%%%%%%%%%%%%%%%%%%%%%%%%%%%%%%%%%%%%%%%%%%%%%%%%%%%%%%%%%%%%%%%
    \textbf{BF16} & - & - & \textbf{21.7} & \textbf{84.4} & \textbf{84.6} & \textbf{36.9} & \textbf{16.0} & \textbf{48.7} & - \\
    %%%%%%%%%%%%%%%%%% AWQ-W3G128 %%%%%%%%%%%%%%%%%%
    \cline{1-10}
    \multirow{2}{*}{\textbf{AWQ}} &   & WikiText2 & 5.8 & 54.4 & 68.4 & 31.8 & 3.7 & 32.8 & \cellcolor{orange!10} \\ 
    & \multirow{-2}{*}{3-16-16} & Numina-Math-1.5 & 5.8 & 60.2 & 71.3 & 28.8 & 6.7 & 34.6 & \cellcolor{orange!10}\multirow{-2}{*}{{1.8}} \\
    %%%%%%%%%%%%%%%%%% GPTQ-W3G128 %%%%%%%%%%%%%%%%%%
    \cline{1-10}
    \multirow{2}{*}{\textbf{GPTQ}} & & WikiText2 & 3.3 & 49.4 & 57.5 & 27.3 & 3.7 & 28.3 & \cellcolor{red!10} \\ 
    & \multirow{-2}{*}{3-16-16} & Numina-Math-1.5 & 10.0 & 71.6 & 75.7 & 23.7 & 9.3 & 38.1 & \cellcolor{red!10}\multirow{-2}{*}{{9.8}} \\
    %%%%%%%%%%%%%%%%%% KVQuant-KV4 %%%%%%%%%%%%%%%%%%
    \cline{1-10}
    \multirow{2}{*}{\textbf{KVQuant$^{*}$}} &  & WikiText2 & 20.8 & 84.6 & 84.0 & 33.8 & 13.4 & 47.3 & \cellcolor{green!10} \\ 
    & \multirow{-2}{*}{16-16-4} & Numina-Math-1.5 & 20.0 & 83.8 & 84.5 & 33.8 & 16.4 & 47.7 & \cellcolor{green!10}\multirow{-2}{*}{{0.4}} \\
    %%%%%%%%%%%%%%%%%% SmoothQuant-W8A8KV8 %%%%%%%%%%%%%%%%%%
    \cline{1-10}
    \multirow{2}{*}{\textbf{SmoothQuant}} &  & WikiText2 & 17.5 & 79.8 & 83.2 & 34.3 & 15.7 & 46.1 & \cellcolor{green!10} \\ 
    & \multirow{-2}{*}{8-8-8} & Numina-Math-1.5 & 20.8 & 79.8 & 82.7 & 35.9 & 15.7 & 47.0 & \cellcolor{green!10}\multirow{-2}{*}{{0.9}} \\
    %%%%%%%%%%%%%%%%%% FlatQuant-W4A4KV4 %%%%%%%%%%%%%%%%%%
    \cline{1-10}
    \multirow{2}{*}{\textbf{FlatQuant}} &  & WikiText2 & 10.0 & 64.8 & 78.6 & 31.8 & 6.7 & 38.4 & \cellcolor{green!10} \\ 
    & \multirow{-2}{*}{4-4-4} & Numina-Math-1.5 & 8.3 & 65.6 & 78.0 & 33.3 & 7.5 & 38.6 & \cellcolor{green!10}\multirow{-2}{*}{{0.2}} \\
%%%%%%%%%%%%%%%%%%%%%%%%%%%%%%%%%%%%%%%%%%%%%%%%%%%%%%%%%%%%%%%%%%%%%%%%%%%%%%%%%%%%%%
    \hline\hline
    \end{tabular}
    }
    \end{center}
    \caption{
        The impact of calibration data domain on different quantization methods. The evaluation is conducted on the quantized DeepSeek-R1-Distill-Qwen-1.5B model across various reasoning benchmarks.
        The \textcolor[HTML]{009900}{green}, \textcolor{orange}{orange} and \textcolor{red}{red} cells stand for the robust (\textcolor[HTML]{009900}{$\leq$1\%}), the fair (\textcolor{orange}{1\%-3\%}) and the sensitive (\textcolor{red}{$\geq$3\%}) quantization methods respectively.
    }
\label{tab:calib-ablat}
\end{table*}

\subsection{Choice of Calibration Data}
\label{sec:discussion_calib}
Calibration data has always been an important part of post-training quantization.
While previous studies~\citep{williams2023impact,lin2023awq} suggest that the overall performance of quantization methods on pre-trained and instruction-tuned models is robust to the source domain of calibration data, we find this is not true for reasoning models.
Instead, the choice of calibration data domain may have a huge impact on the quantization accuracy.
In addition, this effect is highly dependent on the quantization method used. Due to limited space, we leave more details in Appendix~\ref{sec:apdx_discussion_calib}, including the analysis on the distributions of calibration data from different domains.

\paragraph{The domain of calibration data affects GPTQ, but not the rest of the quantization algorithms.}

Prior works~\citep{frantar2022optq,lin2023awq} on post-training quantization usually sample calibration data from pre-training datasets like WikiText2~\citep{merity2016wiki} or C4~\citep{raffel2020c4}. However, we find that using reasoning data for calibration is crucial for quantization methods that rely heavily on the calibration data for quantization error compensation (e.g. GPTQ~\citep{frantar2022optq}). In Figure~\ref{fig:calib-domain-tsne}, we visualize the activation distributions from different source domains. It can be seen that domain gaps exist between reasoning (i.e. Numina-Math-1.5 and LiveCodeBench) and pre-training (i.e. WikiText2) data, which can cause training-inference inconsistency issues that seriously deteriorate the quantized model if it is calibrated on the pre-training data. As shown in Table~\ref{tab:calib-ablat}, switching the GPTQ calibration set from WikiText2 to Numina-Math-1.5 leads to an average accuracy gain of 9.8\%. Also, note that the source domain of the calibration data has a relatively small impact on other quantization methods such as KVQuant$^*$~\citep{hooper2024kvquant}, and FlatQuant~\citep{sun2024flatquant}. These methods only rely on calibration data for quantization parameter computation or outlier channel pattern identification. Given the high consistency of channel-wise magnitude distribution across different domains as shown in Figure~\ref{fig:calib-domain-flatness}, these methods are more robust to domain shift than GPTQ.

\subsection{Scaling Effects of Quantized Reasoning Models}
\label{sec:scaling_size}
% In this section, we study the scaling effect of accuracy w.r.t. to the quantized model size and latency.
In this section, we study the performance change when we scale the size and latency of the quantized model. 
% In this section, we study these caling effects w.r.t. the model size, latency, and reasoning lengths after quantization. 
Specifically, we aim to identify cost-effective quantization configurations that optimally balance accuracy with model size or latency. 
% We also explore whether manually increasing the reasoning steps of quantized LLMs can compensate for the performance drop 
 % caused by quantization and yield further improvements.
% We seek to know that with quantization, what is the most economic configuration that balances accuracy and model size or latency? Meanwhile, how about manually scaling up the reasoning steps of quantized LLMs for further improvement? 

\begin{figure}
    \centering
    \begin{minipage}{0.49\textwidth}
        % \vspace{26pt}
        \centering
        \subfigure[Model Size (GB).]{\includegraphics[width=.48\textwidth]{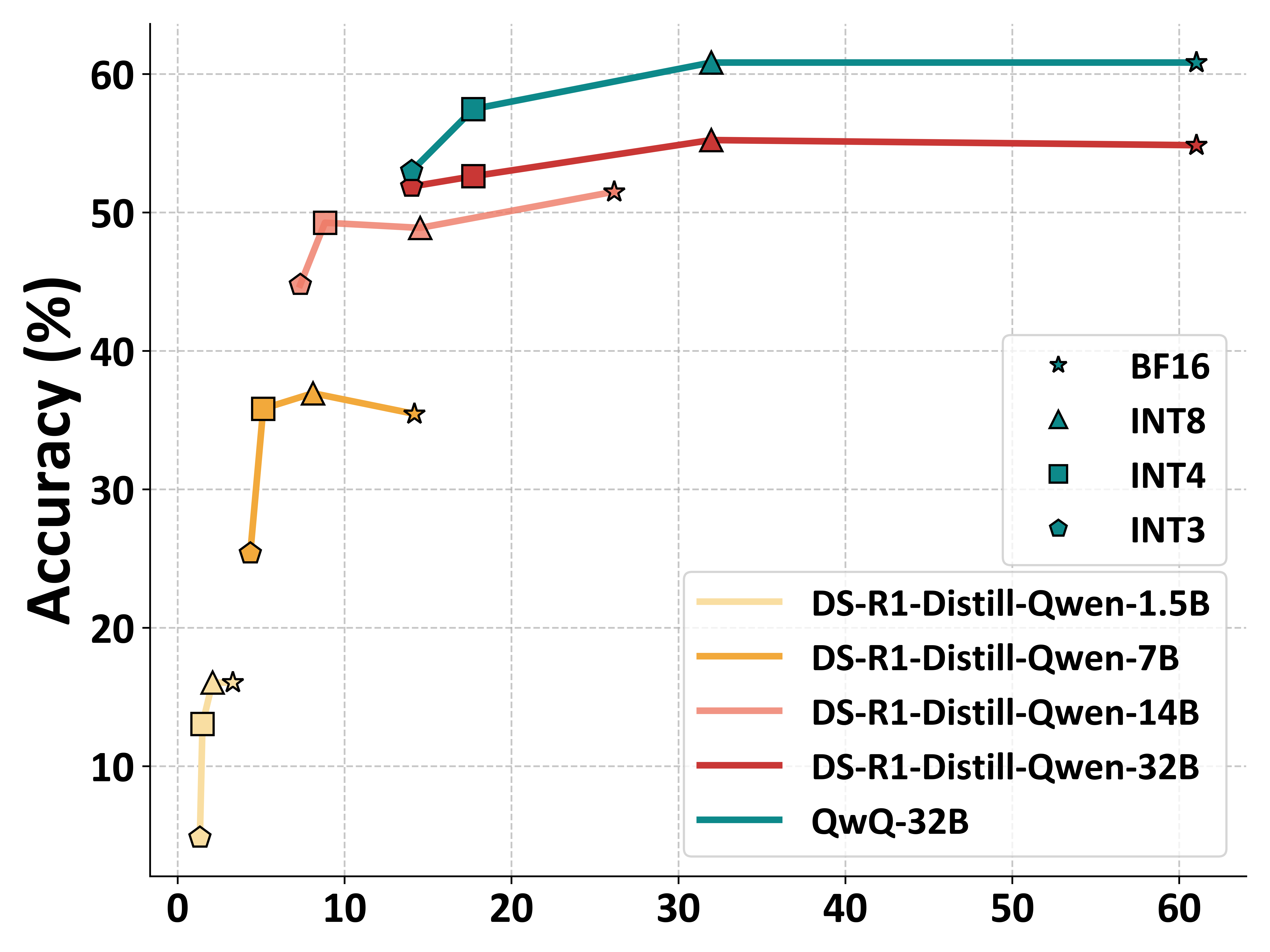}}
        \subfigure[Latency (s).]{\includegraphics[width=.48\textwidth]{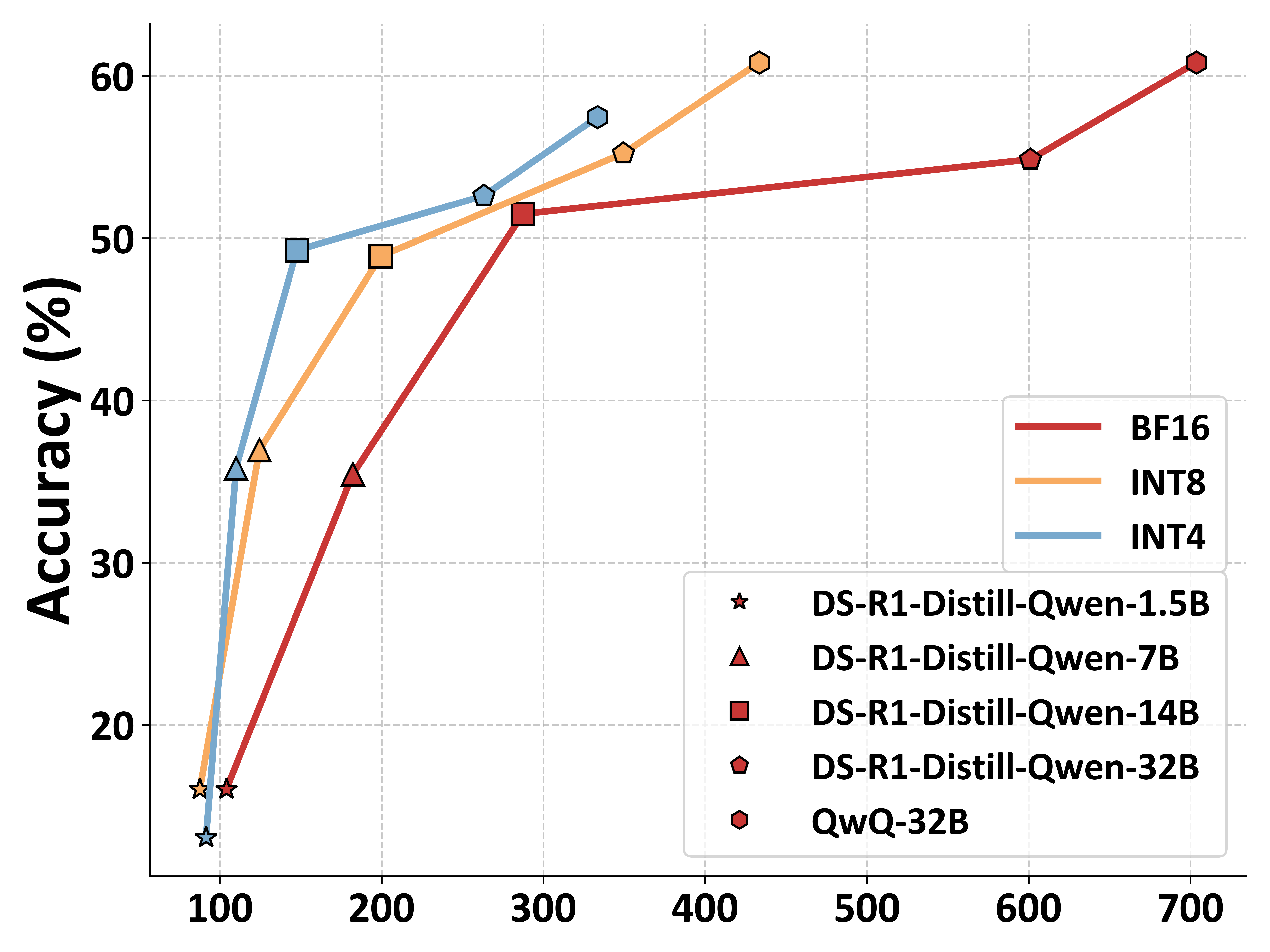}}
        \vspace{-6pt}
        \caption{Model size and average latency \textit{v.s.} accuracy (\%) 
        % under different models and weight quantization bits 
        on LiveCodeBench. }
        \label{fig:infra_scaling}
    \end{minipage}
    \hspace{1.5pt}
    \begin{minipage}{0.49
    \textwidth}
        \centering
        % \includegraphics[width=0.95\linewidth]{figs/scaling/32b_legend_new.png}
        % \vspace{-10pt}
        % \subfigure[DS-R1-Distill-Qwen-1.5B.]{
        \subfigure[Qwen-1.5B.]{
            \includegraphics[width=0.48\linewidth]{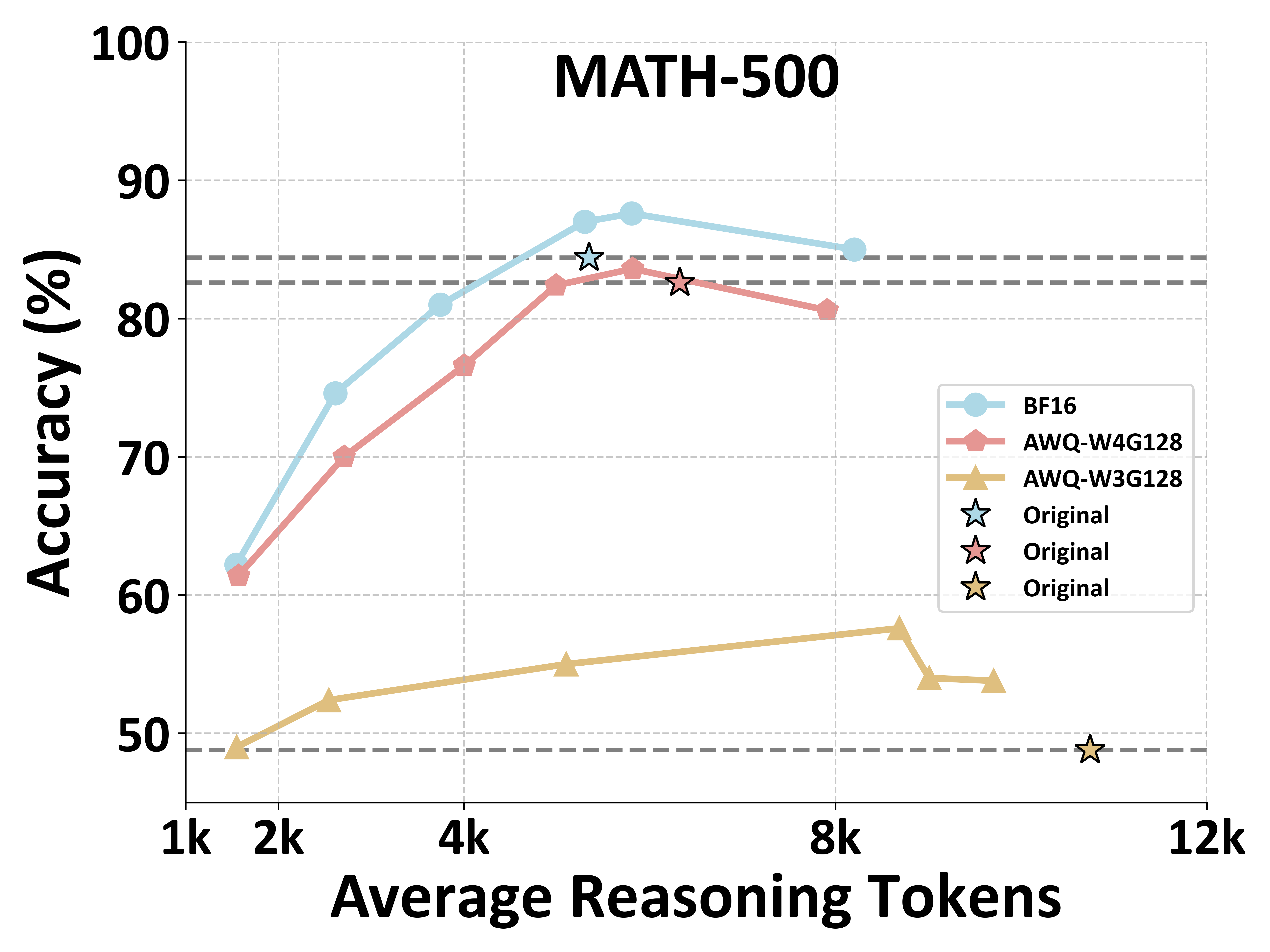}
        }
        \hspace{-8pt}
        % \subfigure[DS-R1-Distill-Qwen-32B.]{
        \subfigure[Qwen-32B.]{
            \includegraphics[width=0.48\linewidth]{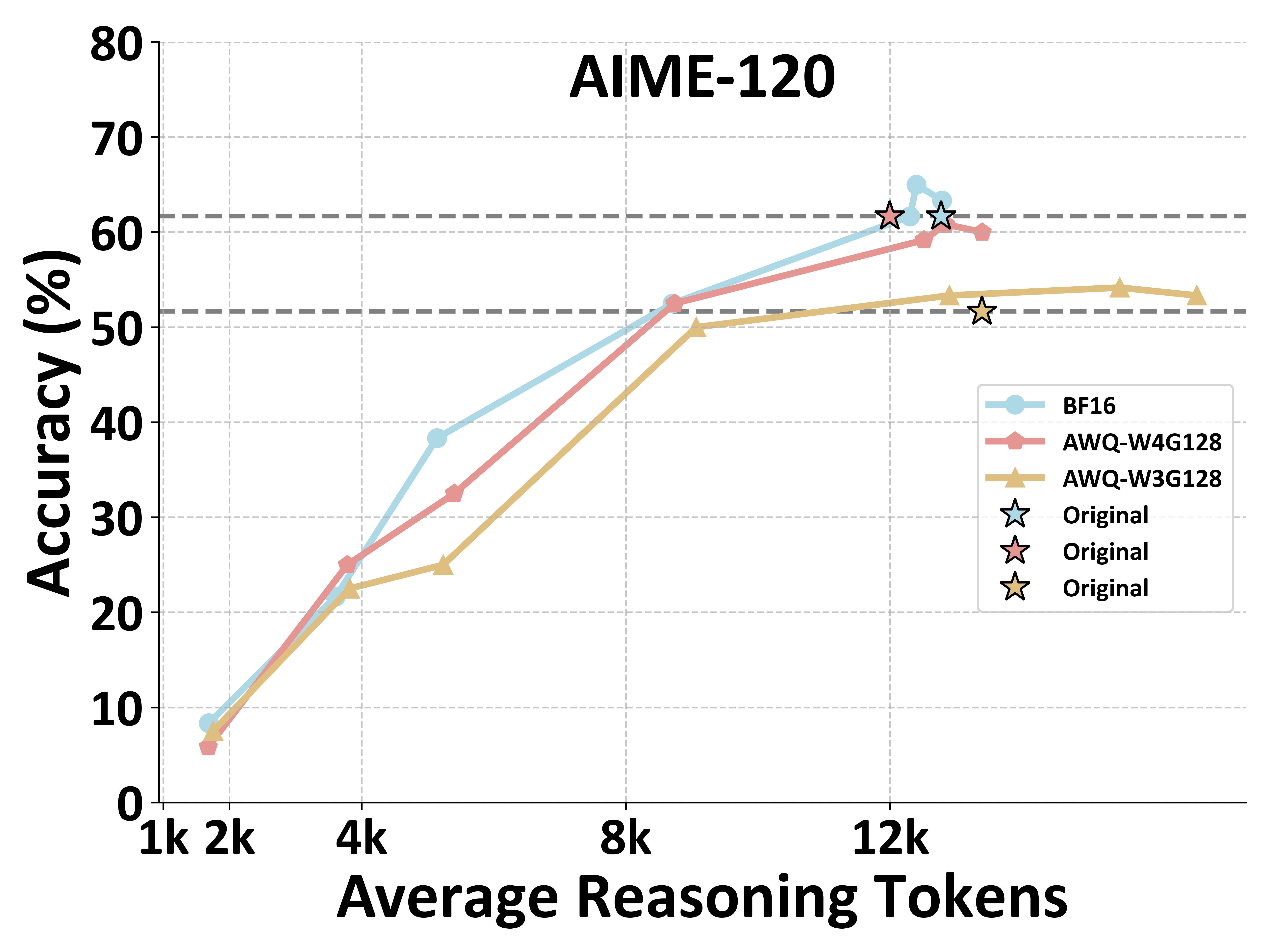}
        }
        \vspace{-6pt}
        \caption{Average reasoning tokens \textit{v.s.} accuracy (\%) on DS-R1-Distill-Qwen models.}
        % under different models and benchmarks. }
        \label{fig:scaling}
    \end{minipage}
\end{figure}

% \input{figs/infra_scaling/infra_scaling}
\paragraph{Model Size Scaling.}
% \textbf{The larger the model, the higher the tolerance for Weight-only, KV cache Quantization and Activation Quantization. }
% \textbf{Low-bit quantized models offer a superior accuracy–memory trade-off compared to both smaller BF16 and uncompressed large models.} 
In Figure~\ref{fig:infra_scaling}(a), we present the trade-off between accuracy and model size based on the series of DeepSeek-R1-Distill Qwen and QwQ-32B. We evaluate different weight-only quantization bit-widths with AWQ (i.e., 8-bit, 4-bit, and 3-bit) against the BF16 model, and report the accuracy on LiveCodeBench. 
It can be found that 
as the model size increases, accuracy improves rapidly. 
Under the same model size, the large LLMs in low precisions are more accurate than small LLMs in BF16. 
% that 
For instance, while both the 3-bit quantized QwQ-32B and Qwen-32B are similar in size (i.e., $\sim$14.5GB) with Qwen-7B in BF16, they significantly outperform Qwen-7B by around 17\%.

\paragraph{Latency Scaling.} 
% \textbf{quantized models demonstrate a favorable accuracy–latency trade-off.} 
% Latency is a critical metric for real-world deployment.
While large models demonstrate better accuracy-size trade-off, they are inevitably slower for inference. 
Here we also study the scaling effect of latency w.r.t. the accuracy. 
The latency is measured as the end-to-end time cost over the LiveCodeBench dataset, i.e., the multiplication between the number of generated tokens with the Time Per Output Token~(TPOT)\footnote{For TPOT, we refer to the official latency report by Qwen: \url{https://qwen.readthedocs.io/en/latest/benchmark/speed_benchmark.html}}. 
According to Figure~\ref{fig:infra_scaling}(b), 
% All speed are evaluted on one NVIDIA A100 80GB and vLLM with the sequence length 6k.
4-bit quantized models provide superior accuracy–latency trade-off than the 8-bit and BF16 counterparts, which mainly benefits from increased generation speed while maintaining the same reasoning length.

\subsection{Test-time Scaling of Quantized Reasoning Models}
\label{sec:scaling_steps}

% \paragraph{Scaling Reasoning Lengths: Longer reasoning steps do improve quantized LLMs, but at lower increasing rate compared with the BF16 model.}

Test-time scaling has become a prevalent technique for enhancing reasoning model performance~\citep{snell2024scaling,muennighoff2025s1,liu2025can}.
To investigate the behavior of quantized reasoning models under test-time scaling, we adopt the method of~\citet{muennighoff2025s1} by controlling reasoning length through dynamic adjustment of average reasoning tokens.
For suppression, early termination forces partial reasoning integration; for promotion, "wait" tokens extend reasoning until token budgets are met.

% Test-time scaling has been widely adopted to improve the performance of reasoning models~\citep{snell2024scaling,muennighoff2025s1,liu2025can}.
% To learn how quantized reasoning models behave with test-time scaling, following prior work~\citep{muennighoff2025s1}, we control the reasoning length by adjusting the number of reasoning tokens. 
% To inhibit reasoning (the first three points of each curve), we add the final answer early when the length exceeds a certain threshold. 
% This forces the model to combine existing reasoning with the final answer. 
% To promote reasoning (the last three points of each curve), we replace the final answer with a certain number of ``wait'' tokens, encouraging more reasoning steps until the target budget.

From Figure~\ref{fig:scaling}, longer reasoning lengths can generally enhance the performance of each model. 
% \bai{\sout{In some cases, they can even outperform the original model without length control. }}
For example, on MATH-500, both AWQ-W4G128 and AWQ-W3G128 exhibit clear improvements with extended reasoning. 
However, the increasing rates of quantized LLMs are generally lower than that of BF16. 
We speculate that it is due to the quantization error accumulated along the sequence, which hinders effective reasoning.
% A similar trend is observed on AIME120.
In addition, excessively long reasoning lengths also result in performance degradation, as also discovered in~\citet{muennighoff2025s1}. 
The output of unnecessary steps could degrade overall accuracy.
These findings highlight the importance of only scaling the reasoning length within an appropriate range.
% which is still 
How to achieve consistent improvement over a longer range is an open challenge.

\section{Conclusion}

In this study, we present the first systematic study on the quantization of reasoning models. 
We examine the performance of different state-of-the-art quantization methods on different reasoning models under various quantization configurations.
We also recommend the lossless quantization settings like W8A8 and W4A16 quantization, as well as the corresponding preferred quantization methods. 
Additionally, we find that task difficulty and model origins have a large impact on the quantized model's performance. In particular, more difficult tasks incur more severe performance degradation caused by quantization. Besides, both the training recipes and model families influence the quantization sensitivity.
Like the BF16 models, quantized models also show consistent improvement when gradually increasing test-time compute budget over a certain range, but may behave worse afterward. 
While our study identifies critical factors influencing the performance of quantized reasoning models, the underlying mechanisms driving these effects remain incompletely understood. Future work will focus on elucidating these causal relationships while advancing systematic methods to optimize the accuracy-efficiency trade-off in reasoning systems.

\section{Acknowledgment}
We sincerely thank Dan Alistarh and Eldar Kurtić for their helpful suggestions in improving our work. We also appreciate the valuable feedback from all anonymous reviewers during the peer-review process.
% \input{latex/sections/limit}
% \input{latex/sections/ack}

% \section*{Author Contributions}

% \section*{Acknowledgments}
% Use unnumbered first level headings for the acknowledgments. All

% \section*{Ethics Statement}
% Authors can add an optional ethics statement to the paper. 
% For papers that touch on ethical issues, this section will be evaluated as part of the review process. The ethics statement should come at the end of the paper. It does not count toward the page limit, but should not be more than 1 page. 

% \bibliography{latex/custom}
% \bibliography{colm2025_conference}
\bibliographystyle{colm2025_conference}

\newpage
\appendix
\section{Quantization Algorithms and Implementation Details}
\label{sec:apdx_algo}

Below, we briefly review the quantization algorithms implemented in this study, including weight-only quantizaiton, KV-Cache quantization, and weight-activation quantization.

\subsection{Weight-only Quantization}
\paragraph{GPTQ}~\citep{frantar2022optq} aims to construct the optimal weights $\hat{\m W}^*$ that minimize the squared error between full-precision and quantized outputs given input $\m X$, i.e., $\arg\min_{\hat{\m W}} \|\hat{\m W}\m X - \m W \m X\|_F$.
% \begin{equation}
%     \arg\min_{\hat{\m W}} \|\hat{\m W}\m X - \m W \m X\|_F,
% \end{equation}
Specifically, it iteratively quantizes weights and updates the remaining ones using approximate second-order information derived from the inverse Hessian matrix. 
This process is optimized via Cholesky decomposition to enhance numerical stability and computational efficiency. In this work, we adopt asymmetric quantization with a group size of 128, i.e., the step size is calculated per 128 consecutive elements per channel in $\m W$. We use activation reorder to deal with outlier channels, with static group activated to reduce latency overhead except for DeepSeek-R1-Distill-LLaMA-70B model. To mitigate the effects of domain shift as detailed in Appendix~\ref{sec:apdx_discussion_calib}, we use reasoning models to self-generate reasoning data on Numina-Math-1.5~\citep{numina_math_datasets} dataset, and randomly sample 128 text sequences of length 2048 to construct the calibration set.

\paragraph{AWQ}~\citep{lin2023awq} aims to identify and protect salient weights to improve the quantized LLMs. Since activations $\m X$ of LLMs are rich in outliers, AWQ employs per-channel scaling, e.g., $\m Y = (\m X \cdot \ms c^{-1})(\ms c \cdot \m W^\top )$, where the channel-wise scaling factor $\ms c\in \mathbb{R}^{n}$ can be analytically derived to balance the magnitudes of input activations and weights.
The optimal $\ms c$ can be obtained by $\ms c = \ms c_{\m X}^{\alpha} \cdot \ms c_{\m W}^{-\beta}$, where $c_{\m X}$ and $c_{\m W}$ are the average channel statistics of activations and weights, and $\alpha, \beta\in [0,1]$ are the associated hyper-parameters. Moreover, the scaled weights $\ms c \cdot \m W^{\top}$ can be merged together to eliminate runtime overhead. Similar to GPTQ, we adopt asymmetric quantization with a group size of 128. For the calibration data, we follow~\citet{lin2023awq} and sample 128 text sequences of length 512 from Pile~\citep{gao2020pile} dataset. 

\subsection{KV-Cache Quantization}
\label{sec:apdx_kv_algo}
\paragraph{KVQuant}~\citep{hooper2024kvquant} leverages the outlier channels in K cache, proposing to quantize K Cache with static per-channel quantization. To better preserve the outlier patterns of K cache, it quantizes K cache before applying RoPE~\citep{su2024roformer}. Additionally, to accommodate the non-uniform distribution of the KV cache, it incorporates non-uniform and mixed-precision quantization. In this work, we skip the non-uniform and mixed-precision quantization in KVQuant and implement KVQuant$^*$ as a simplified baseline, employing static per-channel quantization for K cache and dynamic per-token quantization for V cache with a group size of 128. For DeepSeek-R1-Distill-Qwen-1.5B and 7B models, we further incorporate pre-bias K cache quantization to suppress the extreme outlier channels in K cache, with more details in Appendix~\ref{sec:apdx_1.5b_7b}. We randomly sample 128 text sequences of length 512 from Pile~\citep{gao2020pile} dataset as the calibration set.

\paragraph{QuaRot}~\citep{ashkboos2024quarot} can be applied for both KV cache quantization and weight-activation quantization. It enables end-to-end 4-bit inference by using Hadamard transformations to eliminate outliers in activations and KV caches. Based on the orthogonality of Hadamard matrices (i.e., $\m H^\top\m H = \m I$), the output of quantized linear layer is $\hat{\m Y} = \mathcal{Q}(\m X \m H)\cdot \mathcal{Q}(\m H^{\top} \m W^{\top})$, where the transformed weight $\m W\m H$ can be pre-processed offline to reduce additional runtime overhead. It also integrates GPTQ~\citep{frantar2022optq} to reduce quantization error on weights. For KV cache quantization, we use per-token asymmetric quantization with a group size of 128. For weight-activation quantization, we apply per-channel weight symmetric quantization and per-token activation asymmetric quantization for computing efficiency. We leave the queries in full precision to leverage the memory-bound nature of self-attention operation following~\citet{ashkboos2024quarot}.

\subsection{Weight-Activation Quantization}
\paragraph{SmoothQuant}~\citep{xiao2023smoothquant} employs per-channel scaling, i.e., $\m Y = (\m X \cdot \ms c^{-1})(\ms c \cdot \m W^\top )$, which migrates the quantization difficulty from activations to weights, achieving the optimal balance between weight quantization and activation quantization. SmoothQuant enables 8-bit weight-activation quantization without compromising accuracy. To facilitate the quantization of weights, we further integrate GPTQ. We randomly sample 128 text sequences of length 2048 from the self-generated reasoning dataset as the calibration set and leave the other quantization settings the same as QuaRot.

\paragraph{FlatQuant}~\citep{sun2024flatquant} is the recently proposed approach. Different from QuaRot, it adaptively learns separate Kronecker-decomposed affine transformations to mitigate outliers for each linear layer, i.e., $\m P^*=\arg\min_{\m P} \|\m Y - \mathcal{Q}(\m X \m P)\mathcal{Q}(\m P^{-1}\m W^{\top})\|_F^2$. In addition, FlatQuant also introduces learnable clipping and channel-wise scaling to further reduce the quantization error. By default, we use 128 text sequences of length 2048 from WikiText2~\citep{merity2016wiki} as the calibration set following~\citet{sun2024flatquant}. For DeepSeek-R1-Distill-Qwen-1.5B and 7B models, we extend the sequence length to 4096 to mitigate the extreme outlier issues in the K cache as discussed in Appendix~\ref{sec:apdx_1.5b_7b}. The quantization settings are the same as QuaRot.

\paragraph{MXFP4}~\citep{rouhani2023microscaling} is part of the Microscaling (MX) data format family, which aims to improve the representation ability of low-precision formats through fine-grained per-group quantization. MXFP4 uses a 4-bit floating-point representation with an E2M1 (2-bit exponent and 1-bit mantissa) format for each element, and a shared 8-bit exponential scaling factor for every group of 32 elements. Similar to QuaRot, we employ a mixed-precision strategy for queries in the self-attention module.

% \section{Related Work}
% \bai{TODO}
% \subsection{LLM Quantization}

% \subsection{Reasoning LLMs}

\section{Additional Experiments}
\label{sec:apdx-exp}

\subsection{Results on More Quantized LLMs}
\label{sec:apdx_more_llms}
Below, we present the complete results of QwQ, DeepSeek-R1-Distill-LLaMA, and Qwen3-8B on various reasoning benchmarks.

\begin{table*}[!t]
\begin{center}
\resizebox{1.00\linewidth}{!}
{
    \begin{tabular}{c|l|l|l|rrr|r|r|r|r}
    \hline\hline
    \textbf{Model} & \textbf{Quantization} & \textbf{\makecell{W-A-KV\\\# Bits}} & \textbf{Methods} & \textbf{\makecell{AIME-\\120}}  & \textbf{\makecell{MATH-\\500}} & \textbf{GSM8K} & \textbf{\makecell{GPQA-\\Diamond}} & \textbf{\makecell{LiveCode-\\Bench}} & \textbf{Avg.} & \textbf{Drop}$\downarrow$ \\ 
    \hline
    %%%%%%%%%%%%%%%%%%%%%%%%%%%%%%%%%%%%%%%%%%%%%%%%%%%%%%%%%%%%%%%%%%%%%%%%%%%%%%%%%%%%%%
    % QwQ-32B
    \multirow{15}{*}{\rotatebox[origin=c]{90}{\textbf{QwQ-32B}}}
    & \textbf{BF16} & - & - & \textbf{75.0$_{\pm 2.5}$} & \textbf{97.5$_{\pm 0.2}$} & \textbf{95.6$_{\pm 0.2}$} & \textbf{65.0$_{\pm 1.8}$} & \textbf{60.2$_{\pm 0.8}$} & \textbf{78.7$_{\pm 0.8}$} & - \\
    \cline{2-11}
    & \multirow{4}{*}{\textbf{\textit{Weight-only}}} &  & AWQ & 73.3$_{\pm 2.9}$ & 97.3$_{\pm 0.5}$ & 95.7$_{\pm 0.3}$ & 64.3$_{\pm 1.8}$ & 58.7$_{\pm 0.6}$ & 77.9$_{\pm 1.1}$ & \cellcolor{green!10}{ -0.8} \\  
    % &  & \multirow{-2}{*}{4-16-16} & \multirow{-2}{*}{128} & GPTQ & 73.33 & 96.40 & 95.91 & 64.65 & 58.58 & 77.77 & \cellcolor{green!10}{ -0.24} \\  
    &  & \multirow{-2}{*}{4-16-16} & GPTQ & 73.1$_{\pm 2.9}$ & 97.3$_{\pm 0.2}$ & 95.5$_{\pm 0.2}$ & 64.1$_{\pm 3.8}$ & 58.6$_{\pm 1.4}$ & 77.7$_{\pm 0.3}$ & \cellcolor{green!10}{ -0.9} \\ 
    \cline{3-11}
    &  & \multirow{2}{*}{3-16-16} & AWQ & 60.0$_{\pm 1.4}$ & 96.1$_{\pm 0.5}$ & 95.4$_{\pm 0.2}$ & 57.1$_{\pm 1.3}$ & 50.3$_{\pm 1.1}$ & 71.8$_{\pm 0.6}$ & \cellcolor{red!10}{ -6.9}  \\  
    % &  &  & & GPTQ & 54.17 & 94.40 & 95.15 & 59.09 & 48.13 & 70.19 & \cellcolor{red!10}{ -7.83} \\
    &  &  & GPTQ & 65.3$_{\pm 3.5}$ & 96.3$_{\pm 0.6}$ & 95.5$_{\pm 0.0}$ & 61.5$_{\pm 4.1}$ & 53.0$_{\pm 1.0}$ & 74.3$_{\pm 1.0}$ & \cellcolor{red!10}{-4.4} \\
    \cline{2-11}
    & \multirow{4}{*}{\textbf{\textit{KV Cache}}} &  & KVQuant$^*$ & 75.8$_{\pm 0.0}$ & 97.7$_{\pm 0.2}$ & 95.3$_{\pm 0.2}$ & 65.0$_{\pm 0.6}$ & 58.0$_{\pm 1.1}$ & 78.4$_{\pm 0.1}$ & \cellcolor{green!10}{ -0.3} \\  
    &  & \multirow{-2}{*}{16-16-4} & QuaRot & 76.7$_{\pm 2.9}$ & 97.3$_{\pm 0.3}$ & 95.7$_{\pm 0.2}$ & 64.0$_{\pm 3.6}$ & 60.8$_{\pm 0.7}$ & 78.9$_{\pm 0.8}$ & \cellcolor{green!10}{ 0.3} \\  
    \cline{3-11}
    &  &  & KVQuant$^*$ & 60.3$_{\pm 1.0}$ & 95.7$_{\pm 0.1}$ & 95.5$_{\pm 0.6}$ & 58.8$_{\pm 1.2}$ & 54.7$_{\pm 0.6}$ & 73.0$_{\pm 0.6}$ & \cellcolor{red!10}{ -5.7} \\  
    &  & \multirow{-2}{*}{16-16-3} & QuaRot & 66.1$_{\pm 1.3}$ & 96.7$_{\pm 0.5}$ & 95.0$_{\pm 0.5}$ & 59.4$_{\pm 4.4}$ & 47.1$_{\pm 1.1}$ & 72.9$_{\pm 0.3}$ & \cellcolor{red!10}{ -5.8} \\
    \cline{2-11}
    & \multirow{6}{*}{\textbf{\textit{Weight-Act.}}} & \multirow{3}{*}{} & SmoothQuant & 75.8$_{\pm 0.8}$ & 98.0$_{\pm 0.4}$ & 95.7$_{\pm 0.0}$ & 65.5$_{\pm 5.1}$ & 60.0$_{\pm 1.2}$ & 79.0$_{\pm 1.2}$ & \cellcolor{green!10}{ 0.3} \\
    &  & 8-8-8 & QuaRot & 73.6$_{\pm 1.0}$ & 97.2$_{\pm 0.6}$ & 95.7$_{\pm 0.2}$ & 65.7$_{\pm 2.3}$ & 60.8$_{\pm 1.0}$ & 78.6$_{\pm 0.5}$ & \cellcolor{green!10}{ -0.1} \\
    &  &  & FlatQuant & 76.4$_{\pm 2.7}$ & 97.2$_{\pm 0.5}$ & 95.8$_{\pm 0.4}$ & 63.6$_{\pm 1.3}$ & 59.3$_{\pm 1.6}$ & 78.5$_{\pm 0.7}$ & \cellcolor{green!10}{ -0.2} \\
    \cline{3-11}
    &  &  & MXFP4 & 34.4$_{\pm 4.1}$ & 92.7$_{\pm 0.4}$ & 94.6$_{\pm 0.2}$ & 55.1$_{\pm 2.7}$ & 34.5$_{\pm 1.7}$ & 62.2$_{\pm 0.5}$ & \cellcolor{red!10}{ -16.4} \\  
    % &  & 4-4-4 & \textit{N/A} & QuaRot & 49.17 & 95.40 & 95.15 & 64.65 & 51.12 & 71.10 & \cellcolor{red!10}{ -6.92} \\
    &  & 4-4-4 & QuaRot & 55.3$_{\pm 4.1}$ & 96.0$_{\pm 0.4}$ & 94.9$_{\pm 0.0}$ & 58.4$_{\pm 1.6}$ & 51.2$_{\pm 1.8}$ & 71.2$_{\pm 1.3}$ & \cellcolor{red!10}{-7.5} \\
    &  &  & FlatQuant & 71.4$_{\pm 1.0}$ & 96.5$_{\pm 0.2}$ & 95.5$_{\pm 0.2}$ & 61.5$_{\pm 3.4}$ & 58.5$_{\pm 0.4}$ & 76.7$_{\pm 0.5}$ & \cellcolor{orange!10}{ -2.0} \\

    \cline{3-11}
    \hline\hline
    %%%%%%%%%%%%%%%%%%%%%%%%%%%%%%%%%%%%%%%%%%%%%%%%%%%%%%%%%%%%%%%%%%%%%%%%%%%%%%%%%%%%%%
    \end{tabular}
    }
    \end{center}
    \caption{
    The overall assessment of quantized QwQ-32B model on various reasoning benchmarks. 
    The \textcolor[HTML]{009900}{green}, \textcolor{orange}{orange} and \textcolor{red}{red} cells stand for the lossless (\textcolor[HTML]{009900}{$\leq$1\%}), the fair (\textcolor{orange}{1\%-3\%}) and the risky (\textcolor{red}{$\geq$3\%}) respectively.
}
\label{tab:qwq}
\end{table*}

\begin{table*}[!t]
\begin{center}
\resizebox{1.00\linewidth}{!}
{
    \begin{tabular}{c|l|l|l|rrr|r|r|r|r}
    \hline\hline
    \textbf{Model} & \textbf{Quantization} & \textbf{\makecell{W-A-KV\\\# Bits}} & \textbf{Methods} & \textbf{\makecell{AIME-\\120}}  & \textbf{\makecell{MATH-\\500}} & \textbf{GSM8K} & \textbf{\makecell{GPQA-\\Diamond}} & \textbf{\makecell{LiveCode-\\Bench}} & \textbf{Avg.} & \textbf{Drop}$\downarrow$ \\ 
    \hline
    %%%%%%%%%%%%%%%%%%%%%%%%%%%%%%%%%%%%%%%%%%%%%%%%%%%%%%%%%%%%%%%%%%%%%%%%%%%%%%%%%%%%%%
    % DeepSeek-R1-Distill-LLaMA-8B
    \multirow{15}{*}{\rotatebox[origin=c]{90}{\textbf{DeepSeek-R1-Distill-LLaMA-8B}}}
    & \textbf{BF16} & - & - & \textbf{33.3$_{\pm 0.8}$} & \textbf{91.0$_{\pm 1.1}$} & \textbf{88.7$_{\pm 0.4}$} & \textbf{49.5$_{\pm 2.3}$} & \textbf{36.6$_{\pm 3.2}$} & \textbf{59.8$_{\pm 0.8}$} & - \\
    \cline{2-11}
    & \multirow{4}{*}{\textbf{\textit{Weight-only}}} &  & AWQ & 30.3$_{\pm 3.4}$ & 86.9$_{\pm 0.1}$ & 87.9$_{\pm 0.7}$ & 41.1$_{\pm 0.3}$ & 33.6$_{\pm 2.0}$ & 56.0$_{\pm 0.7}$ & \cellcolor{red!10}{ -3.9} \\  
    % &  & \multirow{-2}{*}{4-16-16} & \multirow{-2}{*}{128} & GPTQ & 30.00 & 89.00 & 88.32 & 41.92 & 35.07 & 56.86 & \cellcolor{orange!10}{ -2.84} \\  
    &  & \multirow{-2}{*}{4-16-16} & GPTQ & 30.6$_{\pm 2.1}$ & 90.0$_{\pm 0.2}$ & 89.0$_{\pm 0.5}$ & 47.0$_{\pm 0.5}$ & 34.6$_{\pm 1.5}$ & 58.2$_{\pm 0.2}$ & \cellcolor{orange!10}{-1.6} \\ 
    \cline{3-11}
    &  & \multirow{2}{*}{3-16-16} & AWQ  & 14.4$_{\pm 1.0}$ & 77.3$_{\pm 1.4}$ & 83.5$_{\pm 1.1}$ & 37.2$_{\pm 1.1}$ & 22.5$_{\pm 0.2}$ & 47.0$_{\pm 0.2}$ & \cellcolor{red!10}{ -12.8} \\  
    &  &  & GPTQ & 23.3$_{\pm 2.9}$ & 78.8$_{\pm 0.9}$ & 73.1$_{\pm 0.6}$ & 39.1$_{\pm 5.6}$ & 27.9$_{\pm 1.8}$ & 48.4$_{\pm 0.6}$ & \cellcolor{red!10}{ -11.4} \\
    \cline{2-11}
    & \multirow{4}{*}{\textbf{\textit{KV Cache}}} &  & KVQuant$^*$ & 36.7$_{\pm 3.8}$ & 88.5$_{\pm 0.8}$ & 87.8$_{\pm 0.1}$ & 45.5$_{\pm 2.2}$ & 35.5$_{\pm 1.9}$ & 58.8$_{\pm 1.6}$ & \cellcolor{orange!10}{ -1.0} \\  
    &  & \multirow{-2}{*}{16-16-4} & QuaRot & 33.9$_{\pm 2.9}$ & 90.0$_{\pm 0.7}$ & 89.1$_{\pm 0.4}$ & 49.7$_{\pm 1.1}$ & 36.8$_{\pm 0.9}$ & 59.9$_{\pm 0.5}$ & \cellcolor{green!10}{ 0.1} \\  
    \cline{3-11}
    &  &  & KVQuant$^*$ & 19.4$_{\pm 2.7}$ & 82.0$_{\pm 0.2}$ & 87.4$_{\pm 0.4}$ & 41.6$_{\pm 0.8}$ & 27.1$_{\pm 0.6}$ & 51.5$_{\pm 0.6}$ & \cellcolor{red!10}{ -8.3} \\  
    &  & \multirow{-2}{*}{16-16-3} & QuaRot & 28.6$_{\pm 4.3}$ & 87.1$_{\pm 0.2}$ & 87.9$_{\pm 0.9}$ & 45.3$_{\pm 3.7}$ & 31.5$_{\pm 3.0}$ & 56.1$_{\pm 0.8}$ & \cellcolor{red!10}{ -3.7} \\ 
    \cline{2-11}
    & \multirow{6}{*}{\textbf{\textit{Weight-Act.}}} & \multirow{3}{*}{} & SmoothQuant & 36.7$_{\pm 2.2}$ & 89.3$_{\pm 0.5}$ & 88.8$_{\pm 0.5}$ & 47.8$_{\pm 4.4}$ & 36.4$_{\pm 1.4}$ & 59.8$_{\pm 0.3}$ & \cellcolor{green!10}{ 0.0} \\   
    &  & 8-8-8 & QuaRot & 38.9$_{\pm 2.9}$ & 89.9$_{\pm 0.9}$ & 88.7$_{\pm 0.6}$ & 48.3$_{\pm 2.8}$ & 37.7$_{\pm 2.3}$ & 60.7$_{\pm 0.9}$ & \cellcolor{green!10}{0.9} \\  
    &  &  & FlatQuant & 35.8$_{\pm 1.4}$ & 89.0$_{\pm 1.5}$ & 88.7$_{\pm 0.3}$ & 48.8$_{\pm 2.5}$ & 38.1$_{\pm 2.3}$ & 60.1$_{\pm 1.1}$ & \cellcolor{green!10}{ 0.3} \\  
    \cline{3-11}
    &  & \multirow{3}{*}{4-4-4} & MXFP4 & 8.9$_{\pm 1.0}$ & 71.1$_{\pm 1.4}$ & 81.4$_{\pm 0.4}$ & 28.5$_{\pm 1.2}$ & 16.0$_{\pm 0.4}$ & 41.2$_{\pm 0.3}$ & \cellcolor{red!10}{ -18.7} \\  
    % &  &  & \textit{N/A} & QuaRot & 10.83 & 68.20 & 68.16 & 34.34 & 18.28 & 39.96 & \cellcolor{red!10}{ -19.74} \\  
    &  &  & QuaRot & 16.4$_{\pm 1.0}$ & 77.3$_{\pm 1.2}$ & 71.4$_{\pm 1.4}$ & 32.8$_{\pm 3.2}$ & 23.3$_{\pm 0.6}$ & 44.2$_{\pm 1.1}$ & \cellcolor{red!10}{ -15.6} \\  
    &  &  & FlatQuant & 21.9$_{\pm 1.3}$ & 85.5$_{\pm 1.8}$ & 85.5$_{\pm 0.6}$ & 42.6$_{\pm 1.8}$ & 30.0$_{\pm 2.4}$ & 53.1$_{\pm 1.3}$ & \cellcolor{red!10}{ -6.7} \\
    \cline{3-11}
    \hline\hline
    %%%%%%%%%%%%%%%%%%%%%%%%%%%%%%%%%%%%%%%%%%%%%%%%%%%%%%%%%%%%%%%%%%%%%%%%%%%%%%%%%%%%%%
    % DeepSeek-R1-Distill-LLaMA-70B
    \multirow{15}{*}{\rotatebox[origin=c]{90}{\textbf{DeepSeek-R1-Distill-LLaMA-70B}}}
    & \textbf{BF16} & - & - & \textbf{58.3$_{\pm 3.8}$} & \textbf{95.6$_{\pm 0.2}$} & \textbf{94.1$_{\pm 0.4}$} & \textbf{67.7$_{\pm 1.8}$} & \textbf{55.9$_{\pm 0.6}$} & \textbf{74.3$_{\pm 0.9}$} & - \\
    \cline{2-11}
    & \multirow{4}{*}{\textbf{\textit{Weight-only}}} &  &  AWQ & 58.3$_{\pm 5.5}$ & 95.3$_{\pm 0.5}$ & 93.7$_{\pm 0.3}$ & 66.3$_{\pm 1.5}$ & 54.1$_{\pm 0.4}$ & 73.6$_{\pm 1.2}$ & \cellcolor{green!10}{ -0.8} \\  
    % &  & \multirow{-2}{*}{4-16-16} & \multirow{-2}{*}{128} & GPTQ & 46.67 & 92.60 & 92.95 & 64.14 & 49.63 & 69.20 & \cellcolor{red!10}{ -4.88} \\   
    &  & \multirow{-2}{*}{4-16-16} & GPTQ & 59.4$_{\pm 1.3}$ & 94.4$_{\pm 0.7}$ & 94.2$_{\pm 0.3}$ & 63.3$_{\pm 2.5}$ & 54.6$_{\pm 1.3}$ & 73.2$_{\pm 0.6}$ & \cellcolor{orange!10}{ -1.1} \\  
    \cline{3-11}
    &  & \multirow{2}{*}{3-16-16} & AWQ  & 42.8$_{\pm 2.7}$ & 93.3$_{\pm 1.2}$ & 93.5$_{\pm 0.3}$ & 62.5$_{\pm 3.3}$ & 48.4$_{\pm 0.9}$ & 68.1$_{\pm 1.1}$ & \cellcolor{red!10}{ -6.2} \\  
    % &  &  & & GPTQ & 11.67 & 70.40 & 87.34 & 29.80 & 21.64 & 44.17 & \cellcolor{red!10}{ -29.91} \\
    &  &  & GPTQ & 51.4$_{\pm 1.3}$ & 93.6$_{\pm 0.4}$ & 94.4$_{\pm 0.2}$ & 64.0$_{\pm 2.5}$ & 48.8$_{\pm 2.4}$ & 70.4$_{\pm 0.8}$ & \cellcolor{red!10}{ -3.9} \\
    \cline{2-11}
    & \multirow{4}{*}{\textbf{\textit{KV Cache}}} &  &  KVQuant$^*$ & 55.6$_{\pm 4.6}$ & 95.1$_{\pm 0.8}$ & 94.1$_{\pm 0.2}$ & 69.5$_{\pm 2.4}$ & 52.5$_{\pm 0.4}$ & 73.4$_{\pm 1.1}$ & \cellcolor{green!10}{ -1.0} \\  
    &  & \multirow{-2}{*}{16-16-4} & QuaRot & 55.6$_{\pm 1.7}$ & 95.1$_{\pm 0.4}$ & 94.2$_{\pm 0.0}$ & 67.5$_{\pm 2.0}$ & 54.1$_{\pm 1.5}$ & 73.3$_{\pm 0.4}$ & \cellcolor{orange!10}{ -1.0} \\  
    \cline{3-11}
    &  &  &  KVQuant$^*$ & 51.4$_{\pm 4.6}$ & 93.3$_{\pm 0.8}$ & 93.8$_{\pm 0.1}$ & 63.0$_{\pm 1.1}$ & 50.1$_{\pm 0.6}$ & 70.3$_{\pm 1.4}$ & \cellcolor{red!10}{ -4.0} \\  
    &  & \multirow{-2}{*}{16-16-3} & QuaRot & 50.6$_{\pm 1.7}$ & 94.5$_{\pm 0.3}$ & 94.2$_{\pm 0.4}$ & 65.0$_{\pm 2.6}$ & 52.2$_{\pm 1.3}$ & 71.3$_{\pm 0.6}$ & \cellcolor{red!10}{ -3.0} \\
    \cline{2-11}
    & \multirow{6}{*}{\textbf{\textit{Weight-Act.}}} &  & SmoothQuant & 58.1$_{\pm 2.7}$ & 95.3$_{\pm 0.5}$ & 94.3$_{\pm 0.4}$ & 69.2$_{\pm 2.7}$ & 54.0$_{\pm 1.4}$ & 74.2$_{\pm 0.7}$ & \cellcolor{green!10}{ -0.1} \\   
    &  & 8-8-8 & QuaRot & 59.7$_{\pm 3.4}$ & 95.3$_{\pm 0.3}$ & 94.1$_{\pm 0.3}$ & 65.0$_{\pm 0.8}$ & 53.7$_{\pm 1.5}$ & 73.6$_{\pm 0.4}$ & \cellcolor{green!10}{ -0.7} \\  
    &  &  & FlatQuant & 62.5$_{\pm 2.5}$ & 95.7$_{\pm 0.8}$ & 94.2$_{\pm 0.2}$ & 66.8$_{\pm 1.8}$ & 54.9$_{\pm 2.3}$ & 74.8$_{\pm 0.9}$ & \cellcolor{green!10}{ 0.5} \\  
    \cline{3-11}
    &  &  & MXFP4 & 6.1$_{\pm 0.5}$ & 68.7$_{\pm 0.6}$ & 88.8$_{\pm 0.3}$ & 37.9$_{\pm 0.5}$ & 12.9$_{\pm 0.2}$ & 42.9$_{\pm 0.1}$ & \cellcolor{red!10}{ -31.4} \\  
    % &  & 4-4-4 & \textit{N/A} & QuaRot & 3.33 & 65.60 & 84.08 & 30.81 & 15.67 & 39.90 & \cellcolor{red!10}{ -34.18} \\  
    &  & 4-4-4 & QuaRot & 9.4$_{\pm 2.7}$ & 70.0$_{\pm 0.4}$ & 86.5$_{\pm 0.4}$ & 33.3$_{\pm 1.8}$ & 22.1$_{\pm 0.8}$ & 44.3$_{\pm 0.9}$ & \cellcolor{red!10}{ -30.0} \\  
    &  &  & FlatQuant & 54.2$_{\pm 0.8}$ & 94.7$_{\pm 0.6}$ & 94.1$_{\pm 0.5}$ & 66.8$_{\pm 4.1}$ & 49.5$_{\pm 1.1}$ & 71.9$_{\pm 0.9}$ & \cellcolor{orange!10}{ -2.5} \\
    \cline{3-11}
    \hline\hline    %%%%%%%%%%%%%%%%%%%%%%%%%%%%%%%%%%%%%%%%%%%%%%%%%%%%%%%%%%%%%%%%%%%%%%%%%%%%%%%%%%%%%%
    \end{tabular}
    }
    \end{center}
    \caption{
    The overall assessment of quantized DeepSeek-R1-Distill-LLaMA models on various reasoning benchmarks. 
    The \textcolor[HTML]{009900}{green}, \textcolor{orange}{orange} and \textcolor{red}{red} cells stand for the lossless (\textcolor[HTML]{009900}{$\leq$1\%}), the fair (\textcolor{orange}{1\%-3\%}) and the risky (\textcolor{red}{$\geq$3\%}) respectively.
}
\label{tab:r1-llama}
\end{table*}

\begin{table*}[!t]
\begin{center}
\resizebox{1.00\linewidth}{!}
{
    \begin{tabular}{c|l|l|l|rrr|r|r|r|r}
    \hline\hline
    \textbf{Model} & \textbf{Quantization} & \textbf{\makecell{W-A-KV\\\# Bits}} & \textbf{Methods} & \textbf{\makecell{AIME-\\120}}  & \textbf{\makecell{MATH-\\500}} & \textbf{GSM8K} & \textbf{\makecell{GPQA-\\Diamond}} & \textbf{\makecell{LiveCode-\\Bench}} & \textbf{Avg.} & \textbf{Drop}$\downarrow$ \\ 
    \hline
    %%%%%%%%%%%%%%%%%%%%%%%%%%%%%%%%%%%%%%%%%%%%%%%%%%%%%%%%%%%%%%%%%%%%%%%%%%%%%%%%%%%%%%
    % Qwen3-8B
    \multirow{14}{*}{\rotatebox[origin=c]{90}{\textbf{Qwen3-8B}}}
    & \textbf{BF16} & - & - & \textbf{68.6$_{\pm 1.3}$} & \textbf{97.1$_{\pm 0.6}$} & \textbf{95.3$_{\pm 0.4}$} & \textbf{60.6$_{\pm 0.9}$} & \textbf{57.2$_{\pm 0.4}$} & \textbf{75.8$_{\pm 0.5}$} & - \\
    \cline{2-11}
    & \multirow{4}{*}{\textbf{\textit{Weight-only}}} &  & AWQ & 66.1$_{\pm 1.3}$ & 97.0$_{\pm 1.0}$ & 95.0$_{\pm 0.2}$ & 59.6$_{\pm 0.5}$ & 54.7$_{\pm 2.1}$ & 74.5$_{\pm 0.2}$ & \cellcolor{orange!10}{ -1.3} \\  
    &  & \multirow{-2}{*}{4-16-16} & GPTQ & 66.9$_{\pm 2.1}$ & 96.5$_{\pm 0.6}$ & 95.2$_{\pm 0.1}$ & 59.4$_{\pm 2.0}$ & 53.2$_{\pm 0.4}$ & 74.3$_{\pm 0.6}$ & \cellcolor{orange!10}{ -1.5} \\ 
    \cline{3-11}
    &  & \multirow{2}{*}{3-16-16} & AWQ & 44.7$_{\pm 2.4}$ & 92.9$_{\pm 1.0}$ & 94.1$_{\pm 0.3}$ & 46.8$_{\pm 1.5}$ & 35.3$_{\pm 1.7}$ & 62.8$_{\pm 0.1}$ & \cellcolor{red!10}{ -13.0}  \\ 
    &  &  & GPTQ & 43.3$_{\pm 2.5}$ & 92.8$_{\pm 0.7}$ & 94.1$_{\pm 0.2}$ & 44.6$_{\pm 1.5}$ & 31.5$_{\pm 1.6}$ & 61.3$_{\pm 0.4}$ & \cellcolor{red!10}{-14.5} \\
    \cline{2-11} 
    & \multirow{4}{*}{\textbf{\textit{KV Cache}}} &  & KVQuant$^*$ & 66.7$_{\pm 2.2}$ & 97.0$_{\pm 0.2}$ & 95.4$_{\pm 0.2}$ & 60.9$_{\pm 0.6}$ & 56.6$_{\pm 0.4}$ & 75.3$_{\pm 0.5}$ & \cellcolor{green!10}{ -0.5} \\
    &  & \multirow{-2}{*}{16-16-4} & QuaRot & 70.0$_{\pm 6.0}$ & 97.3$_{\pm 0.1}$ & 95.2$_{\pm 0.2}$ & 59.6$_{\pm 1.0}$ & 56.7$_{\pm 1.1}$ & 75.8$_{\pm 1.2}$ & \cellcolor{green!10}{ 0.0} \\  
    \cline{3-11}
    &  &  & KVQuant$^*$ & 54.4$_{\pm 3.5}$ & 95.3$_{\pm 0.2}$ & 94.7$_{\pm 0.3}$ & 54.0$_{\pm 1.8}$ & 43.3$_{\pm 0.4}$ & 68.4$_{\pm 0.6}$ & \cellcolor{red!10}{ -7.4} \\  
    &  & \multirow{-2}{*}{16-16-3} & QuaRot & 57.8$_{\pm 2.7}$ & 95.7$_{\pm 1.1}$ & 94.1$_{\pm 0.2}$ & 50.7$_{\pm 2.4}$ & 43.5$_{\pm 2.7}$ & 68.3$_{\pm 0.6}$ & \cellcolor{red!10}{ -7.4} \\
    \cline{2-11}
    & \multirow{5}{*}{\textbf{\textit{Weight-Act.}}} & \multirow{3}{*}{} & SmoothQuant & 71.1$_{\pm 4.1}$ & 96.6$_{\pm 0.4}$ & 95.3$_{\pm 0.2}$ & 59.8$_{\pm 1.1}$ & 56.3$_{\pm 1.4}$ & 75.8$_{\pm 0.4}$ & \cellcolor{green!10}{ 0.1} \\
    &  & 8-8-8 & QuaRot & 71.1$_{\pm 2.6}$ & 96.7$_{\pm 0.1}$ & 95.5$_{\pm 0.1}$ & 59.6$_{\pm 1.3}$ & 56.8$_{\pm 0.6}$ & 76.0$_{\pm 0.3}$ & \cellcolor{green!10}{ 0.2} \\
    &  &  & FlatQuant & 73.6$_{\pm 2.6}$ & 96.9$_{\pm 0.8}$ & 95.3$_{\pm 0.3}$ & 59.3$_{\pm 2.8}$ & 57.2$_{\pm 1.6}$ & 76.5$_{\pm 1.4}$ & \cellcolor{green!10}{ 0.7} \\
    \cline{3-11}
    &  &  & QuaRot & 50.0$_{\pm 1.4}$ & 94.9$_{\pm 0.4}$ & 94.1$_{\pm 0.1}$ & 49.5$_{\pm 2.0}$ & 40.6$_{\pm 1.5}$ & 65.8$_{\pm 0.8}$ & \cellcolor{red!10}{ -10.0} \\
    &  & \multirow{-2}{*}{4-4-4} & FlatQuant   & 61.1$_{\pm 2.1}$ & 95.5$_{\pm 0.3}$ & 94.8$_{\pm 0.2}$ & 53.7$_{\pm 0.8}$ & 47.1$_{\pm 0.8}$ & 70.5$_{\pm 0.5}$ & \cellcolor{red!10}{ -5.3} \\

    \cline{3-11}
    \hline\hline
    %%%%%%%%%%%%%%%%%%%%%%%%%%%%%%%%%%%%%%%%%%%%%%%%%%%%%%%%%%%%%%%%%%%%%%%%%%%%%%%%%%%%%%
    \end{tabular}
    }
    \end{center}
    \caption{
    The overall assessment of quantized Qwen3-8B model on various reasoning benchmarks. 
    The \textcolor[HTML]{009900}{green}, \textcolor{orange}{orange} and \textcolor{red}{red} cells stand for the lossless (\textcolor[HTML]{009900}{$\leq$1\%}), the fair (\textcolor{orange}{1\%-3\%}) and the risky (\textcolor{red}{$\geq$3\%}) respectively.
}
\label{tab:qwen3}
\end{table*}

\paragraph{Results on QwQ.}
From Table~\ref{tab:qwq}, it can be seen that conclusions from \S~\ref{sec:rq} remain consistent. For example, 4-bit weight-only quantization reaches lossless results while 3-bit induces non-negligible accuracy loss, e.g., over 7\% degradation on LiveCodeBench for both AWQ and GPTQ. For KV cache quantization, both KVQuant$^*$ and QuaRot achieve lossless 4-bit quantization, but all suffer from severe quantization loss when the KV cache is further quantized into 3 bits. The degradation is most pronounced on difficult tasks with long response lengths such as AIME-120 and LiveCodeBench. For weight-activation quantization, 8-bit quantization remains lossless. For the more challenging 4-bit weight-activation quantization, only FlatQuant maintains the accuracy.

\paragraph{Results on DeepSeek-R1-Distill-LLaMA.}
As shown in Table~\ref{tab:r1-llama}, the findings align with the conclusions presented in \S~\ref{sec:rq}. LLaMA models can have lossless quantization results for 4-bit KV cache quantization and 8-bit weight-activation quantization. The 4-bit weight quantization is nearly lossless. For lower bits, all of the evaluated quantization methods become risky, e.g. W3G128 incurs accuracy loss over 11\% for the 8B model.

\paragraph{Results on Qwen3-8B.}
We further evaluate the performance on the recently released Qwen3-8B. As shown in Table~\ref{tab:qwen3}, the overall findings remain consistent with our previous analysis in \S~\ref{sec:rq}. For weight-only quantization, 4-bit methods yield minimal degradation (within 1.5\%). However, reducing weights to 3 bits results in substantial performance drops, exceeding 13\% on average. For KV cache quantization, both KVQuant$^*$ and QuaRot achieve lossless results in 4-bit settings. Yet, when further quantized to 3 bits, they suffer over 7\% accuracy loss. Finally, 8-bit weight-activation quantization remains essentially lossless across all evaluated tasks.

\subsection{Performance v.s. Difficulty Levels}
\label{apdx:task_difficulty}
\begin{figure*}[t]
\centering
\includegraphics[width=0.85\linewidth]{figs/num_tokens/num_tokens_legend.png}
\vspace{-2ex}
\subfigure[DS-R1-Distill-Qwen-1.5B.]{\includegraphics[width=.32\textwidth]{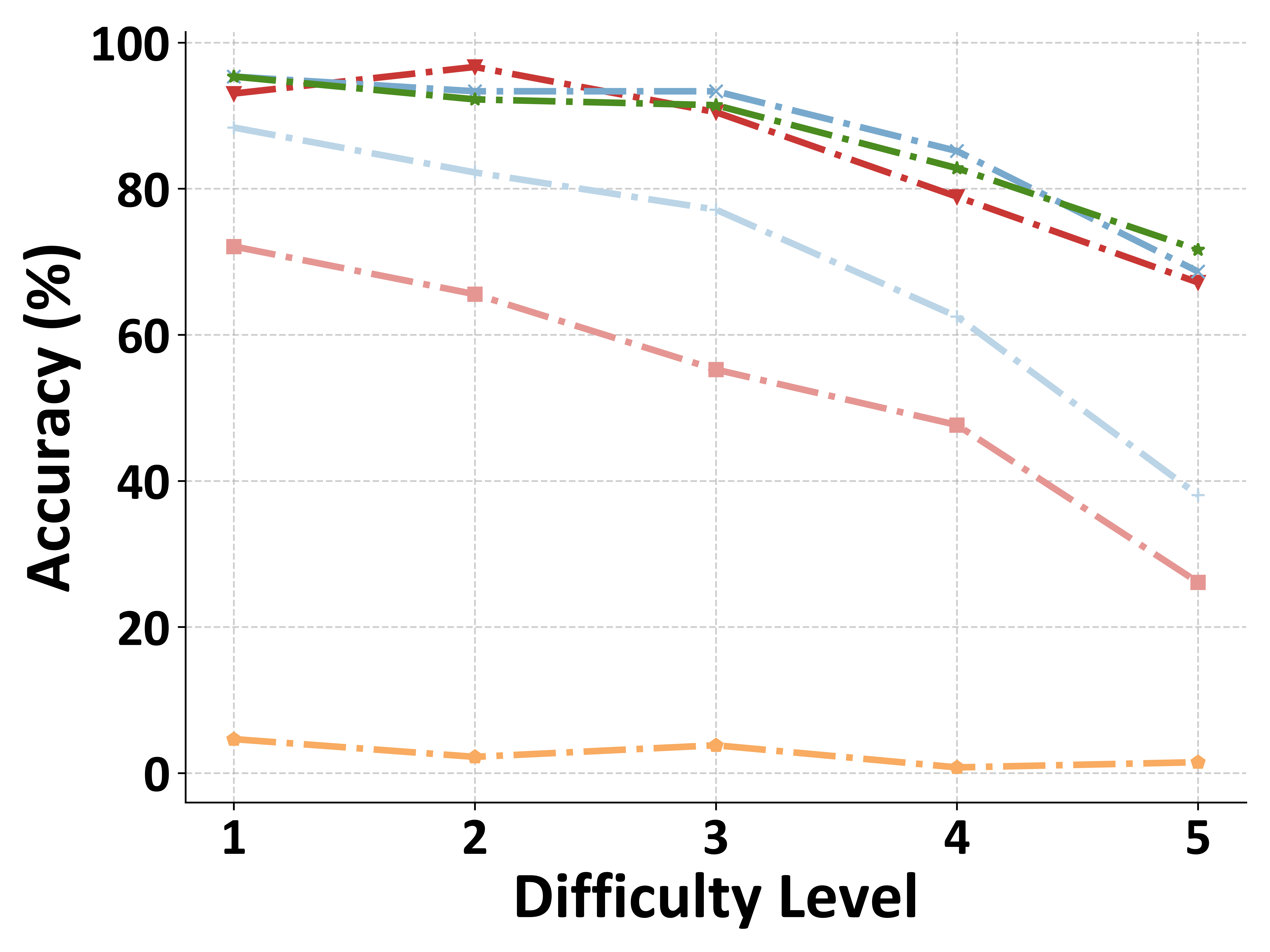}}
% \hspace{-8pt}
\subfigure[DS-R1-Distill-Qwen-14B.]{\includegraphics[width=.32\textwidth]{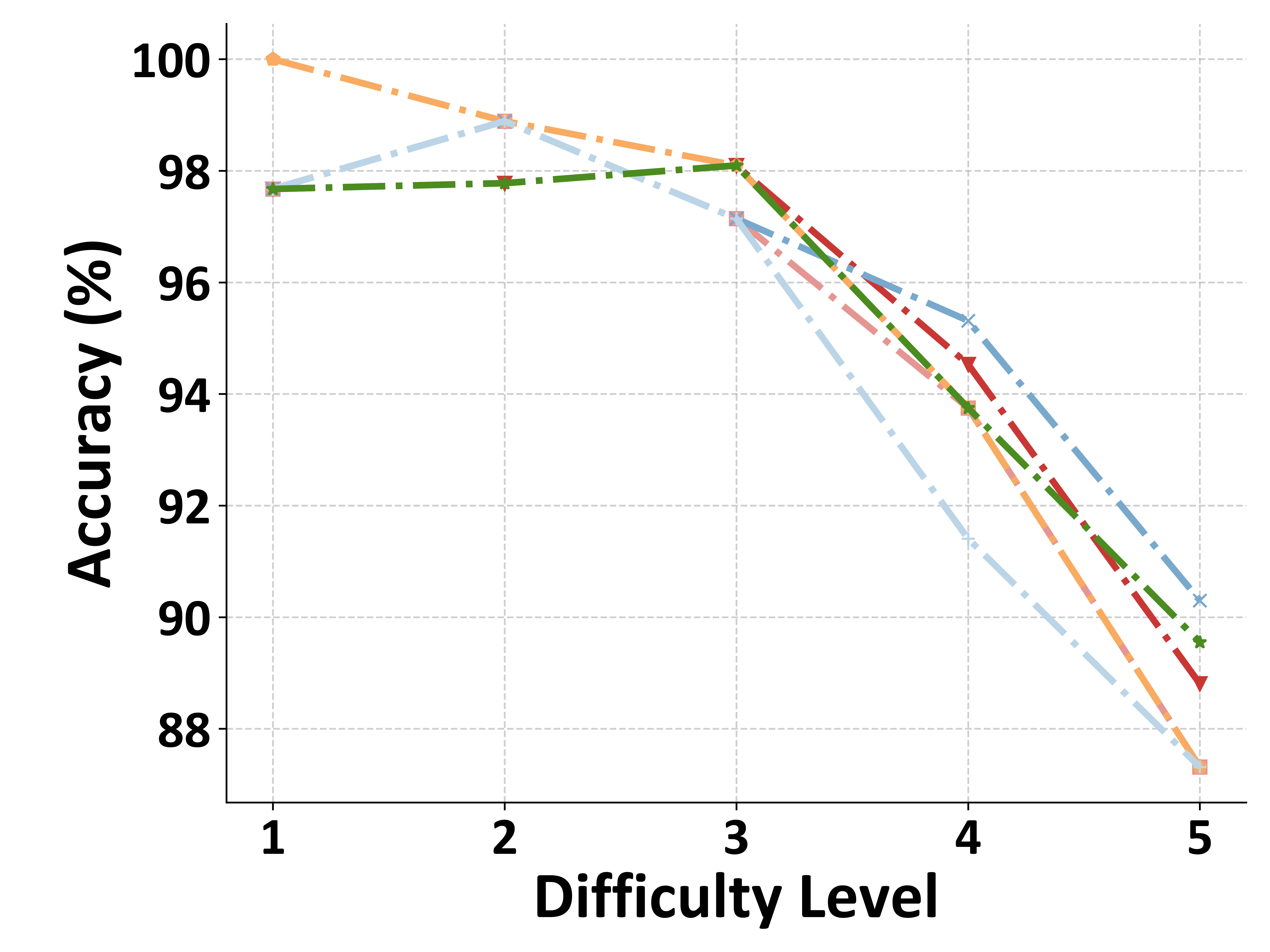}}
% \hspace{-8pt}
\subfigure[QwQ-32B.]{\includegraphics[width=.32\textwidth]{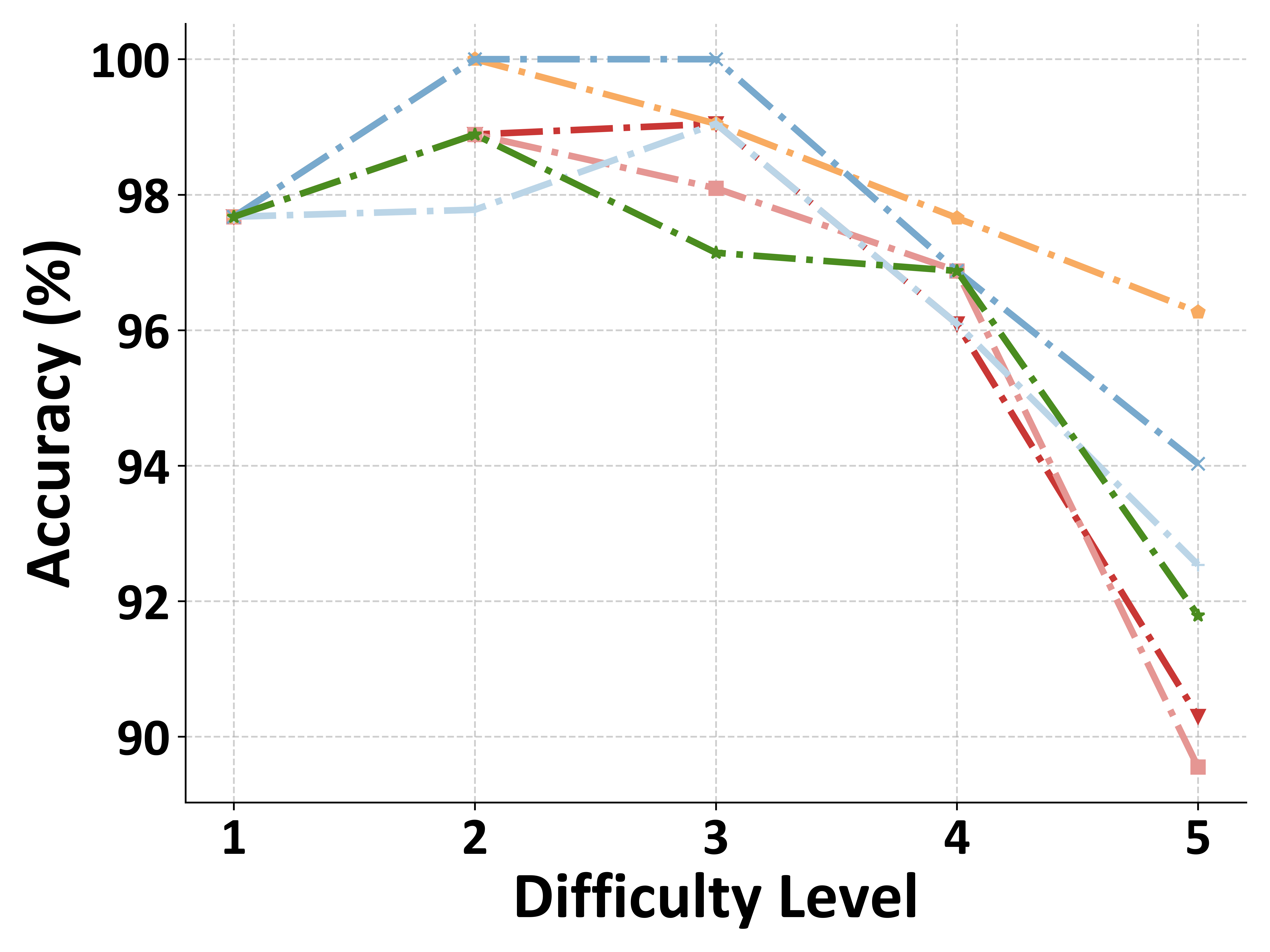}}
% \hspace{-8pt}
% \subfigure[DeepSeek-R1-Distill-Qwen-14B]{\includegraphics[width=.24\textwidth]{figs/num_tokens/DeepSeek-R1-Distill-Qwen-14B_notitle.png}}
% \vspace{-15pt}
\caption{The accuracy of different difficulty levels on MATH-500.}
\label{fig:math500_acc_level}
\end{figure*}
Figure~\ref{fig:math500_acc_level} presents the performance across varying difficulty levels on the MATH-500 benchmark. The results reveal that quantization-induced accuracy loss becomes more pronounced as task difficulty increases. For example, FlatQuant-W4A4KV4 achieves comparable performance to the full-precision 14B model, while it suffers over $2\%$ accuracy loss at level $5$. This indicates that complex tasks exhibit greater vulnerability to precision reduction. 

\subsection{Output Length v.s. Difficulty Levels}
\label{sec:apdx_length}
\begin{figure*}[t]
\centering
\includegraphics[width=0.85\linewidth]{figs/num_tokens/num_tokens_legend.png}
\subfigure[DS-R1-Distill-Qwen-1.5B.]{\includegraphics[width=.32\textwidth]{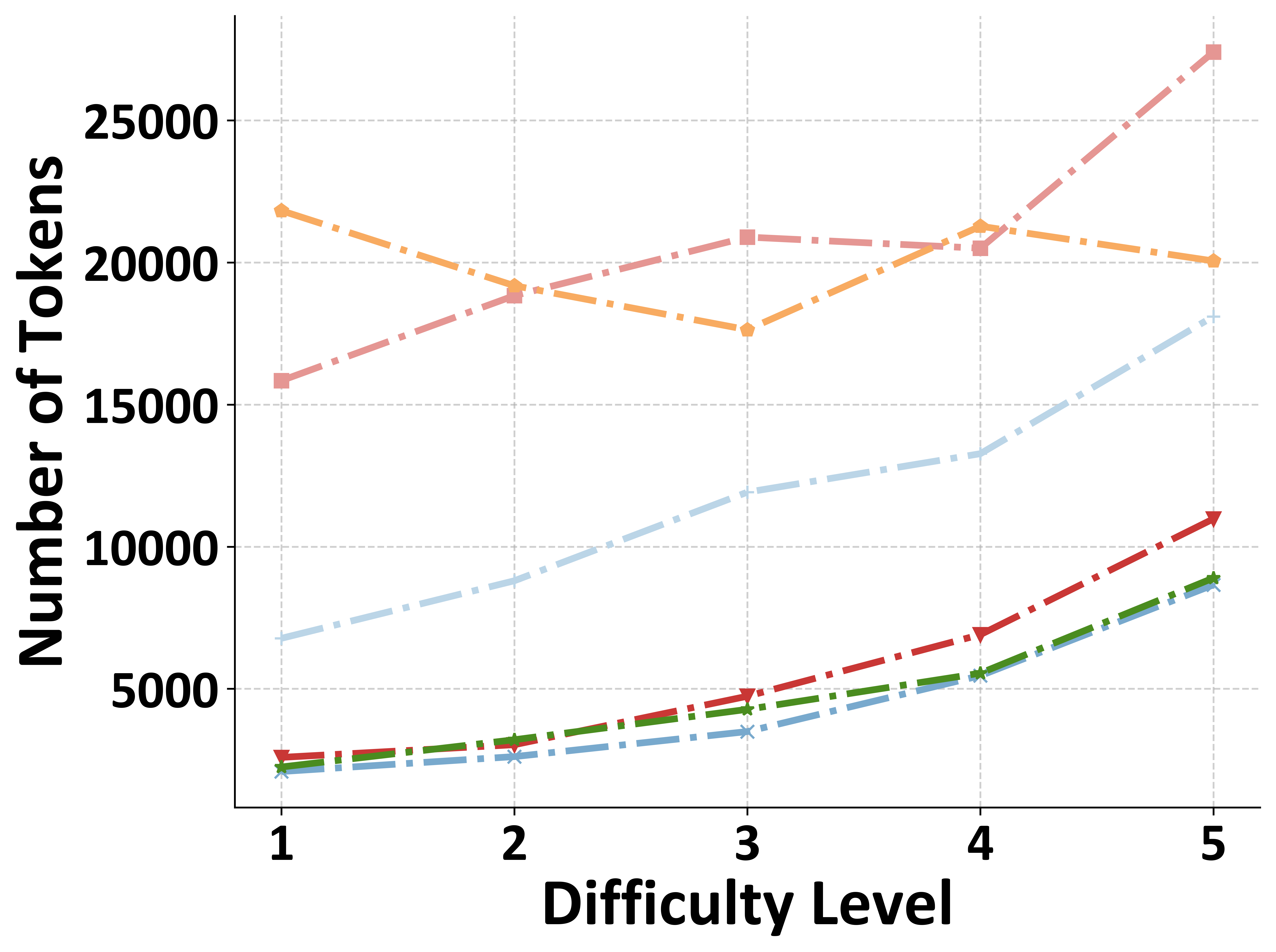}}
\hspace{-8pt}
\subfigure[DS-R1-Distill-Qwen-14B.]{\includegraphics[width=.32\textwidth]{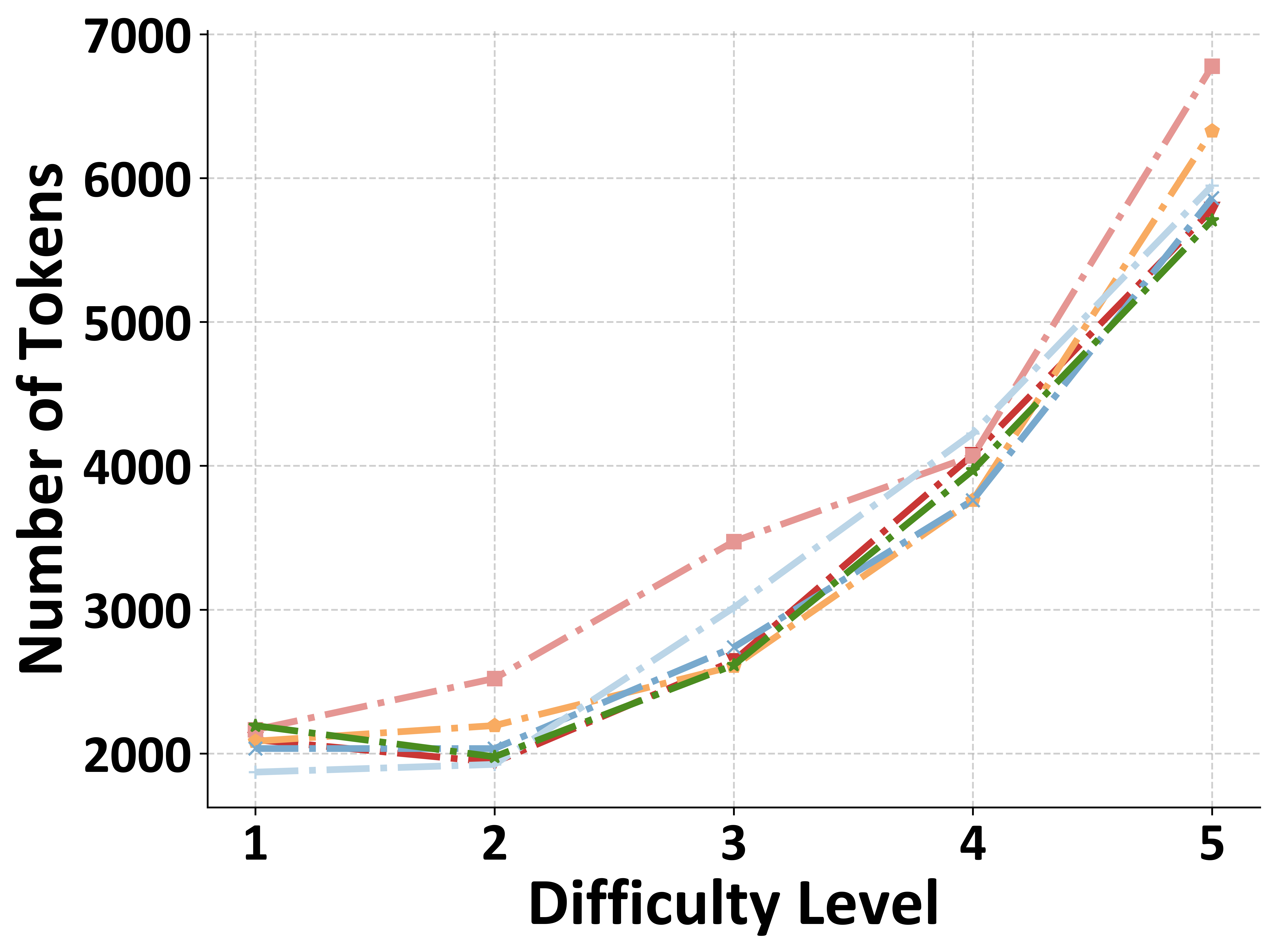}}
\hspace{-8pt}
\subfigure[QwQ-32B.]{\includegraphics[width=.32\textwidth]{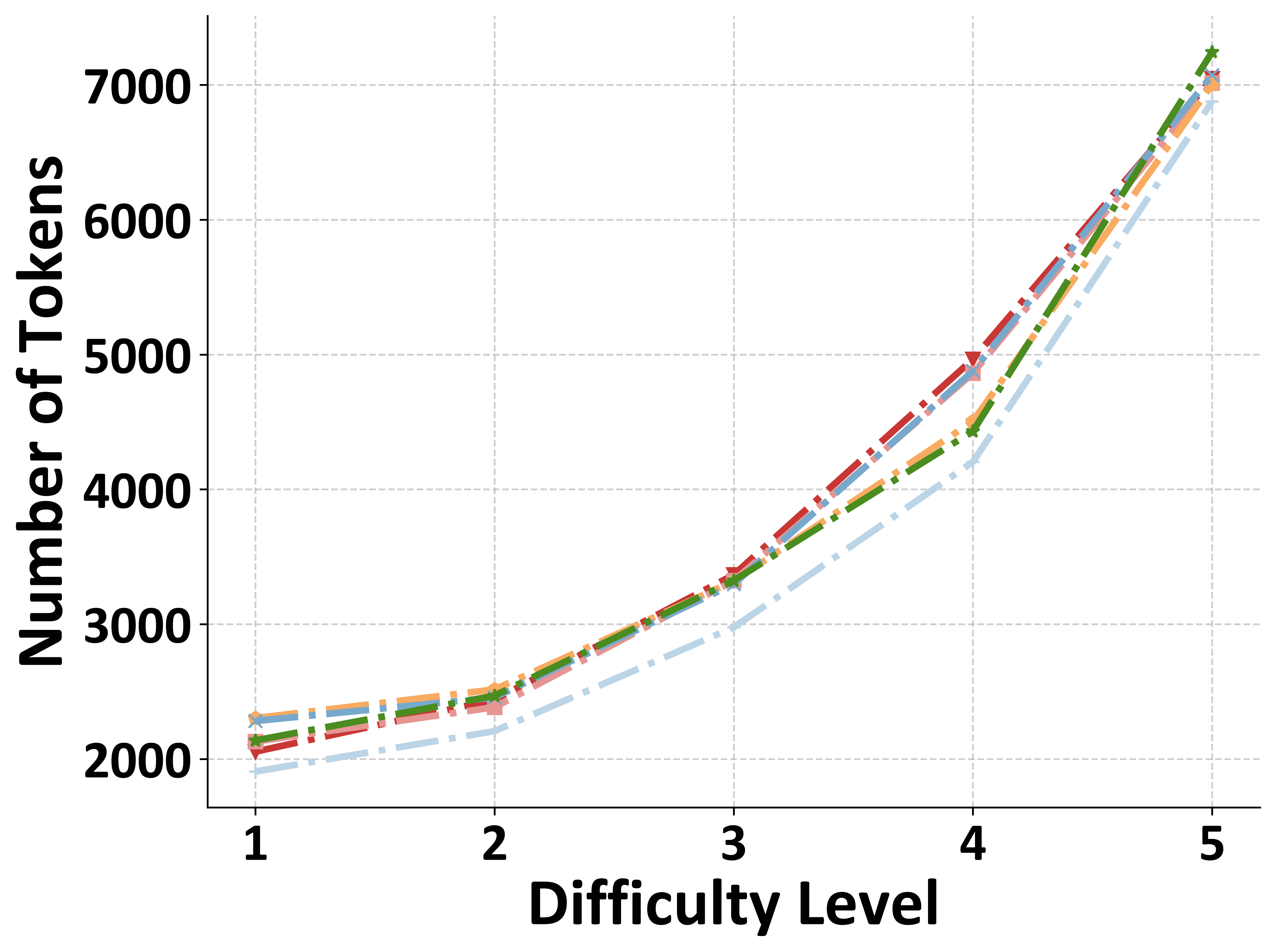}}
\caption{Number of generated tokens of different difficulty levels on MATH-500. }
\label{fig:num_t_math500}
\end{figure*}

Here we further examine the impact of task difficulty on the MATH-500 benchmark, which categorizes problems into five difficulty levels. We have the following observations from Figure~\ref{fig:num_t_math500}: 1) Reasoning models tend to generate more tokens at test time when solving harder problems. 2) Quantized LLMs with minor accuracy degradation (e.g., W4G128, W8A8KV8) do not generate longer responses than the BF16 model. However, lower bit-width on smaller LLMs (in particular, DeepSeek-R1-Distill-Qwen-1.5B with W3G128 and W4A4KV4 quantization) could lead to increased output, which is similar to the conclusion in \S~\ref{sec:output_length}.

\section{Additional Discussions}
\label{sec:apdx_discussion}

\subsection{Choice of Calibration Data}
\label{sec:apdx_discussion_calib}

\begin{figure*}
    \centering
    \begin{minipage}{0.48\textwidth}
        \centering
        \includegraphics[width=0.95\linewidth]{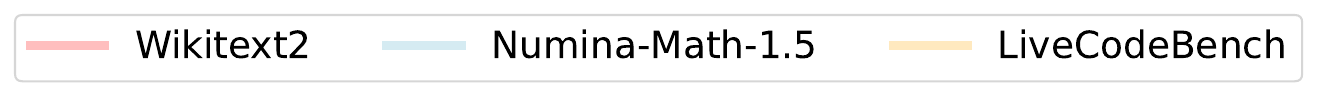}
        \subfigure[$\m X_g$ of the $14^{\textrm{th}}$ Transformer layer in DeepSeek-R1-Distill-Qwen-1.5B.]{\includegraphics[width=0.45\linewidth]{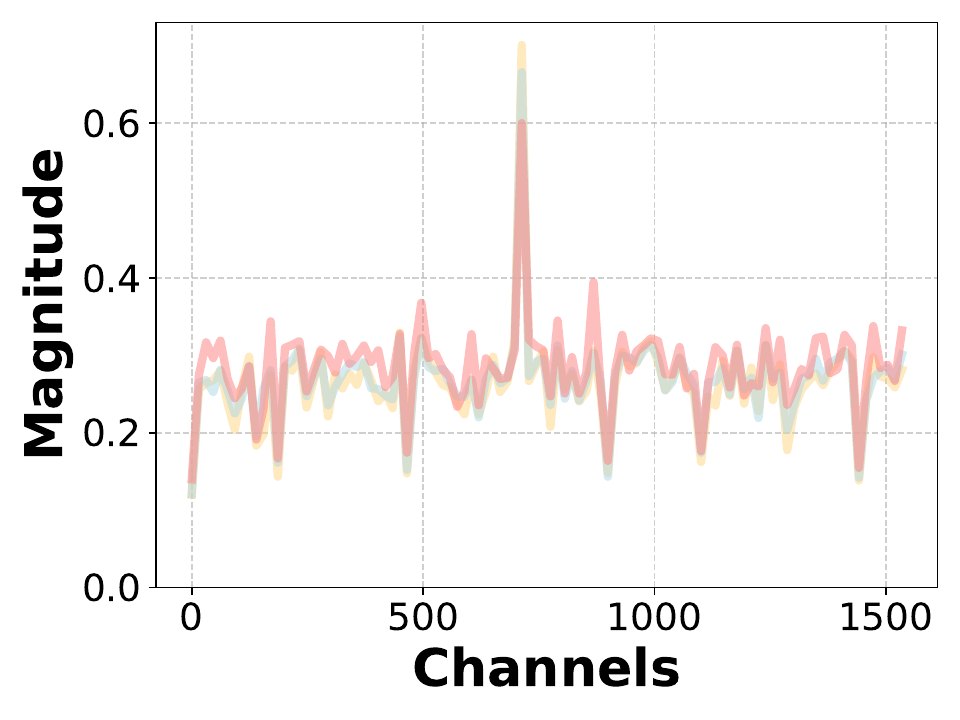}}
        \subfigure[$\m X_q$ of the $21^{\textrm{st}}$ Transformer layer in DeepSeek-R1-Distill-Qwen-1.5B.]{\includegraphics[width=0.45\linewidth]{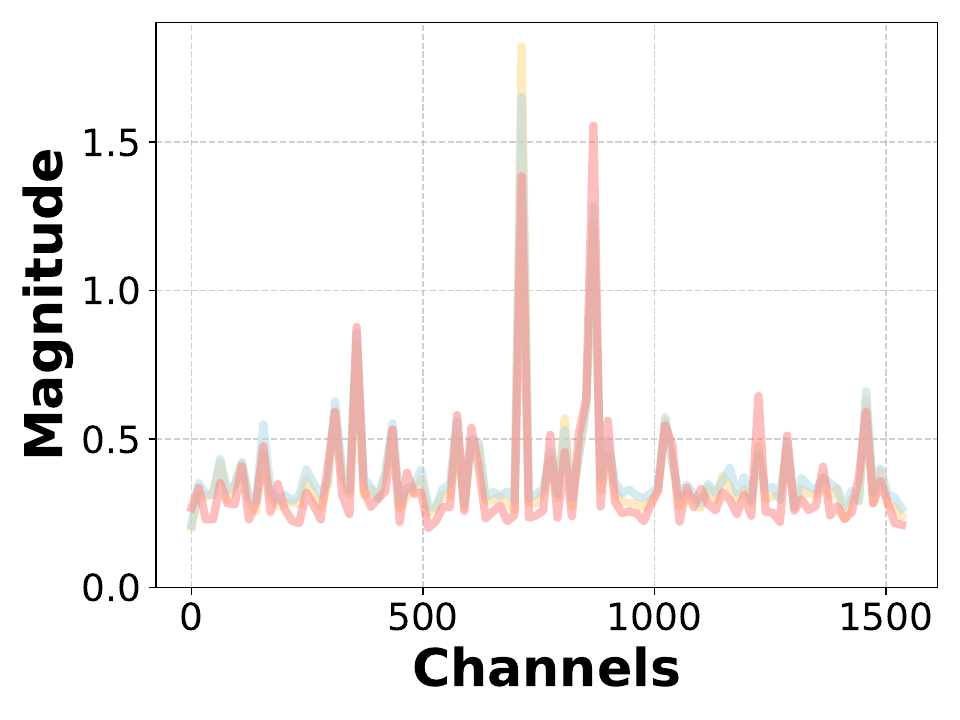}}
        \vspace{-1.5ex}
        \caption{Channel-wise magnitude distributions of activations from different source domains. Reasoning (i.e. Numina-Math-1.5 and LiveCodeBench) and pre-training (i.e. WikiText2) data share similar distributions. The statistics are computed over 32K tokens. $\m X_g$ and $\m X_q$ denote the inputs of the feed-forward gate projection layer and the self-attention query projection layer in a Transformer layer, respectively.}
        \label{fig:calib-domain-flatness}
    \end{minipage}
    \hfill
    \begin{minipage}{0.48\textwidth}
        \centering
        \includegraphics[width=0.95\linewidth]{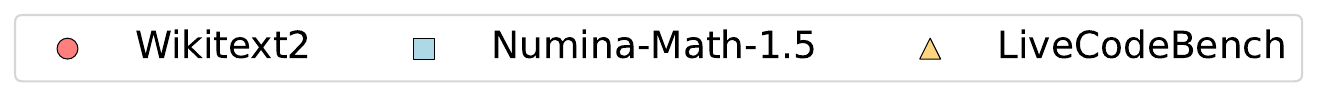}
        \subfigure[$\m X_g$ of the $14^{\textrm{th}}$ Transformer layer in DeepSeek-R1-Distill-Qwen-1.5B.]{
            \includegraphics[width=0.45\linewidth]{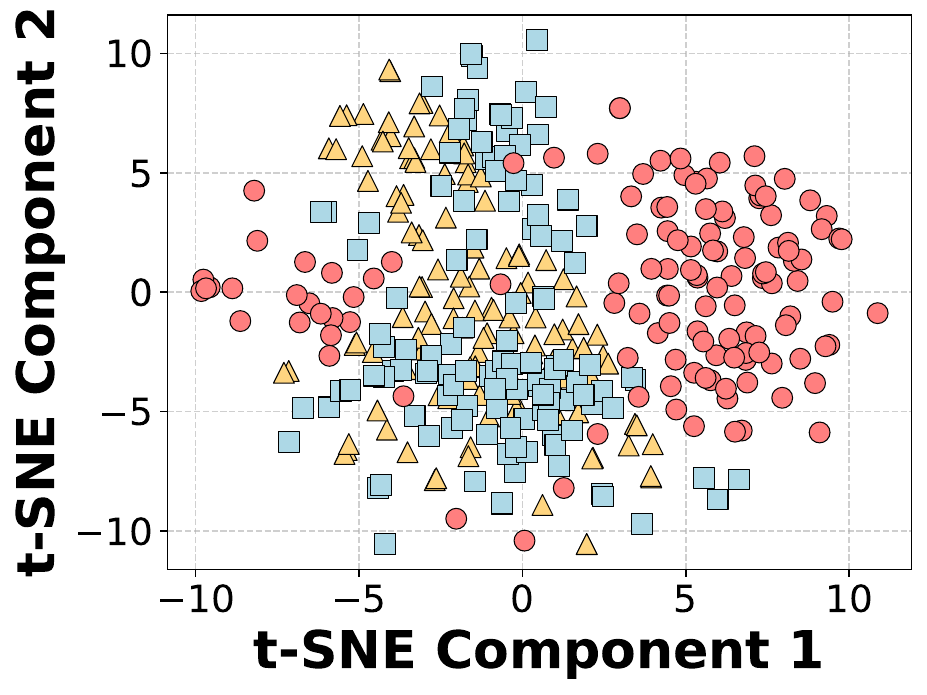}
        }
        \subfigure[$\m X_q$ of the $21^{\textrm{st}}$ Transformer layer in DeepSeek-R1-Distill-Qwen-1.5B.]{
            \includegraphics[width=0.45\linewidth]{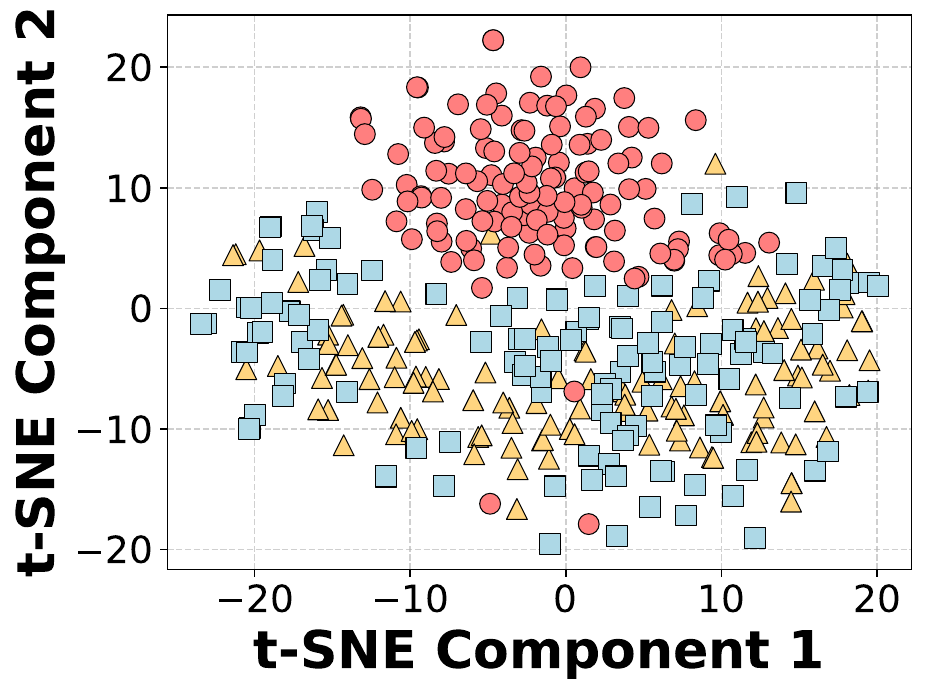}
        }
        \vspace{-1.5ex}
        \caption{t-SNE visualization of activations from different source domains. Domain gaps exist between reasoning (i.e. Numina-Math-1.5 and LiveCodeBench) and pre-training (i.e. WikiText2) data. We randomly sample 128 tokens from each domain. $\m X_g$ and $\m X_q$ denote the inputs of the feed-forward gate projection layer and the self-attention query projection layer in a Transformer layer, respectively.}
        \label{fig:calib-domain-tsne}
    \end{minipage}
    \vspace{-1.5ex}
\end{figure*}

% Calibration data has always been an important part of post-training quantization.
% While previous studies~\citep{williams2023impact,lin2023awq} suggest that the overall performance of quantization methods on pre-trained and instruction-tuned models is robust to the source domain of calibration data, we find this is not true for reasoning models.
% Instead, the choice of calibration data domain may have a huge impact on the quantization accuracy.
% In addition, this effect is highly dependent on the quantization method used. In the following, we first analyze the distributions of calibration data from different domains, and examine its impact on different quantization methods accordingly.

\paragraph{Calibration Data Collection.}
We collect calibration data from different source domains, including reasoning (i.e. Numina-Math-1.5~\citep{numina_math_datasets} and LiveCodeBench~\citep{jain2024livecodebench}) and pre-training (i.e. WikiText2~\citep{merity2016wiki}) data. The reasoning data is self-generated by the reasoning model, given the problems in the original dataset, with generation configurations defined in \S~\ref{sec:exp_setup}.

\paragraph{The Distribution of Calibration Data from Different Domains.}
In Figure~\ref{fig:calib-domain-flatness} and Figure~\ref{fig:calib-domain-tsne}, we visualize the channel-wise magnitude and activation distributions from different source domains, respectively. It can be seen that the channel-wise magnitude distributions across different source domains share the same outlier channels, demonstrating notable similarity. Besides, the activation distributions across different reasoning datasets are also close to each other, regardless of the dataset category. However, as shown in Figure~\ref{fig:calib-domain-tsne}, activations of the reasoning data exhibit a completely distinct distribution from those of the pre-training data, although their channel-wise magnitude distributions resemble each other. This domain gap may raise issues for post-training quantization methods if pre-training data is used for calibration.

\subsection{Analysis on DeepSeek-R1-Distill-Qwen-1.5B and 7B Models}
\label{sec:apdx_1.5b_7b}

As discussed in \S~\ref{sec:rq4}, different LLM origins can have various impacts on the quantization of reasoning models. We find that the K cache of DeepSeek-R1-Distill-Qwen-1.5B and 7B models exhibit extreme outlier channels due to the huge bias terms in the self-attention key projection layer, which significantly increases the difficulty of KV cache quantization on these two models. In this section, we first reveal the phenomenon of extreme outlier channels in DeepSeek-R1-Distill-Qwen-1.5B and 7B models as well as its root, and then propose some practical solutions to mitigate its impact on KV cache quantization.

\subsubsection{Extreme Outlier Channels in K Cache}
\label{sec:apdx_kcache_outlier}

\begin{figure*}
    \centering
    \subfigure[$\m Y_k^{\text{pre\_bias}}$ of the $0^{\textrm{th}}$ Transformer layer in DeepSeek-R1-Distill-Qwen-1.5B.]{\includegraphics[width=0.24\linewidth]{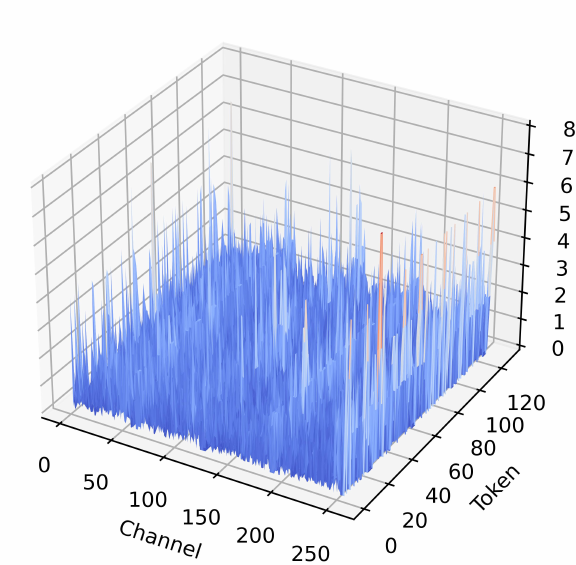}}
    \subfigure[K cache of the $0^{\textrm{th}}$ Transformer layer in DeepSeek-R1-Distill-Qwen-1.5B.]{\includegraphics[width=0.24\linewidth]{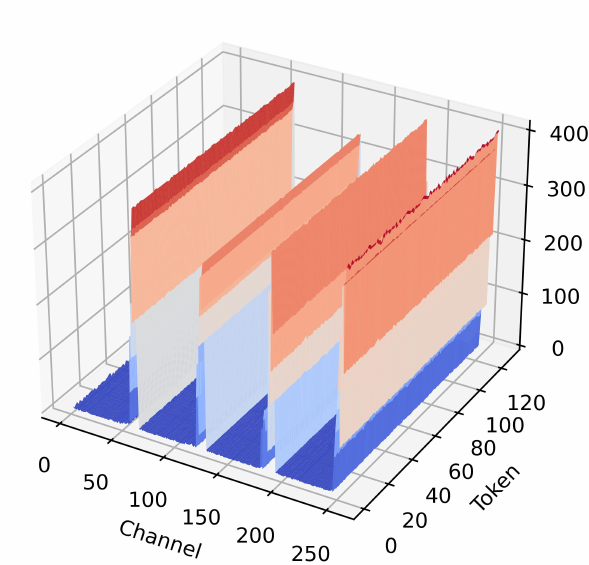}}
    \subfigure[$\m Y_k^{\text{pre\_bias}}$ cache of the $0^{\textrm{th}}$ Transformer layer in DeepSeek-R1-Distill-Qwen-7B.]{\includegraphics[width=0.24\linewidth]{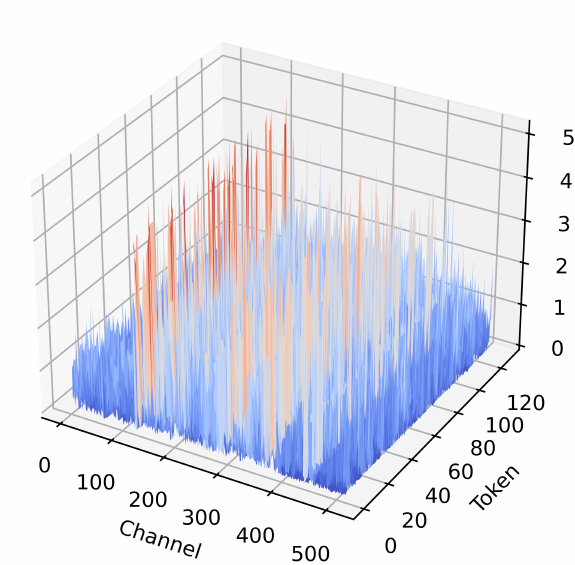}}
    \subfigure[K cache of the $0^{\textrm{th}}$ Transformer layer in DeepSeek-R1-Distill-Qwen-7B.]{\includegraphics[width=0.24\linewidth]{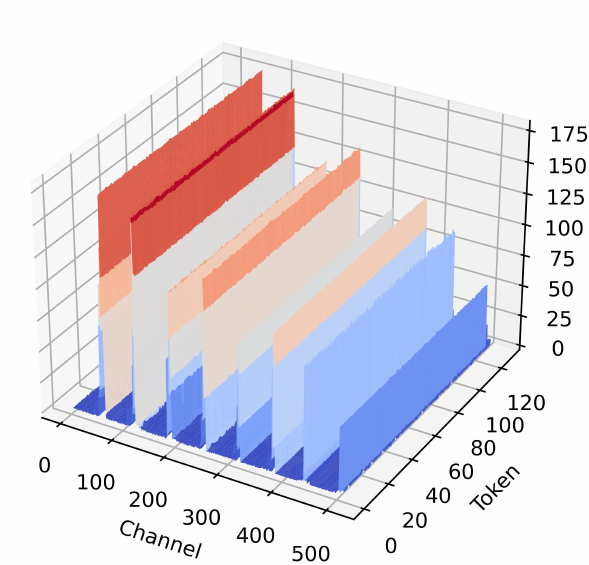}}
    \vspace{-1.5ex}
    \caption{Visualizations of K cache before and after the bias term of key projection layers in DeepSeek-R1-Distill-Qwen-1.5B and 7B models. $\m Y_k^{\text{pre\_bias}}$ denotes the outputs of the self-attention key projection layer before adding the bias term.}
    \label{fig:1.5b_7b_kv}
    \vspace{-1.5ex}
\end{figure*}

Qwen~\citep{bai2023qwen} models add biases in the self-attention query, key, and value projection layers for better length extrapolation ability. However, we find that the bias terms of key projection layers can be extremely large in the pre-trained Qwen-1.5B and 7B models, e.g. the maximum absolute value in key projection bias terms reaches 402 in Qwen-1.5B. The distilled reasoning models inherit the large bias terms from pre-trained models, leading to extreme outlier channels on K cache as shown in Figure~\ref{fig:1.5b_7b_kv}. This phenomenon well explains the large performance gap between KVQuant$^*$ and QuaRot on  DeepSeek-R1-Distill-Qwen-1.5B and 7B models in Table~\ref{tab:r1-qwen}. The presence of extreme outlier channels severely degrades the performance of per-token quantization methods like QuaRot. These outliers force an expansion of the quantization range, leading to catastrophic failures in QuaRot’s accuracy. In contrast, KVQuant$^*$ mitigates the outlier channels by employing per-channel quantization, which effectively constrains the quantization range and achieves higher accuracy.

\subsubsection{Recipes for Extreme Outlier Channels on K Cache}

\begin{table*}[!t]
\begin{center}
\resizebox{1.00\linewidth}{!}
{
    \begin{tabular}{c|l|l|rrr|r|r|r|r}
    \hline\hline
    \textbf{Model} & \textbf{Methods} & \textbf{\makecell{W-A-KV\\\# Bits}} & \textbf{\makecell{AIME-\\120}}  & \textbf{\makecell{MATH-\\500}} & \textbf{GSM8K} & \textbf{\makecell{GPQA-\\Diamond}} & \textbf{\makecell{LiveCode-\\Bench}} & \textbf{Avg.} & \textbf{Drop}$\downarrow$ \\ 
    \hline
    %%%%%%%%%%%%%%%%%%%%%%%%%%%%%%%%%%%%%%%%%%%%%%%%%%%%%%%%%%%%%%%%%%%%%%%%%%%%%%%%%%%%%%
    % BF16
    \multirow{5}{*}{\textbf{1.5B}} & \textbf{BF16} & - & \textbf{21.7} & \textbf{84.4} & \textbf{84.6} & \textbf{36.9} & \textbf{16.0} & \textbf{48.7} & - \\
    %%%%%%%%%%%%%%%%%%%%%%%%%%%%%%%%%%%%%%%%%%%%%%%%%%%%%%%%%%%%%%%%%%%%%%%%%%%%%%%%%%%%%%
    % 1.5B-KVQuant-KV4
    \cline{2-10}
    & \textbf{KVQuant$^*$} & \multirow{2}{*}{16-16-4} & 7.5 & 58.2 & 67.9 & 29.8 & 10.5 & 34.8 & \cellcolor{red!10}{\footnotesize -13.9} \\
    & \textbf{+Pre\_bias} &  & \textbf{20.8} & \textbf{83.6} & \textbf{84.9} & \textbf{31.3} & \textbf{16.4} & \textbf{47.4} & \cellcolor{orange!10}{\footnotesize \textbf{-1.3}} \\
    %%%%%%%%%%%%%%%%%%%%%%%%%%%%%%%%%%%%%%%%%%%%%%%%%%%%%%%%%%%%%%%%%%%%%%%%%%%%%%%%%%%%%%
    % 1.5B-KVQuant-KV3
    \cline{2-10}
    & \textbf{KVQuant$^*$} & \multirow{2}{*}{16-16-3} & 4.2 & 62.4 & 67.2 & 23.7 & 10.5 & 33.6 & \cellcolor{red!10}{\footnotesize -15.1} \\
    & \textbf{+Pre\_bias} &  & \textbf{9.2} & \textbf{66.2} & \textbf{71.3} &\textbf{31.3} & \textbf{10.5} & \textbf{37.7} & \cellcolor{red!10}{\footnotesize \textbf{-11.0}} \\
    %%%%%%%%%%%%%%%%%%%%%%%%%%%%%%%%%%%%%%%%%%%%%%%%%%%%%%%%%%%%%%%%%%%%%%%%%%%%%%%%%%%%%%
    \hline\hline
    %%%%%%%%%%%%%%%%%%%%%%%%%%%%%%%%%%%%%%%%%%%%%%%%%%%%%%%%%%%%%%%%%%%%%%%%%%%%%%%%%%%%%%
    % BF16
    \multirow{5}{*}{\textbf{7B}} & \textbf{BF16} & - & \textbf{45.0} & \textbf{94.6} & \textbf{91.4} & \textbf{50.0} & \textbf{35.5} & \textbf{63.3} & - \\
    %%%%%%%%%%%%%%%%%%%%%%%%%%%%%%%%%%%%%%%%%%%%%%%%%%%%%%%%%%%%%%%%%%%%%%%%%%%%%%%%%%%%%%
    % 7B-KVQuant-KV4
    \cline{2-10}
    & \textbf{KVQuant$^*$} & \multirow{2}{*}{16-16-4} & 0.0 & 7.0 & 5.0 & 25.3 & 0.4 & 7.5 & \cellcolor{red!10}{\footnotesize -55.8} \\
    & \textbf{+Pre\_bias} &  & \textbf{40.8} & \textbf{93.2} & \textbf{91.4} & \textbf{47.5} & \textbf{35.1} & \textbf{61.6} & \cellcolor{orange!10}{\footnotesize \textbf{-1.7}} \\
    %%%%%%%%%%%%%%%%%%%%%%%%%%%%%%%%%%%%%%%%%%%%%%%%%%%%%%%%%%%%%%%%%%%%%%%%%%%%%%%%%%%%%%
    % 7B-KVQuant-KV3
    \cline{2-10}
    & \textbf{KVQuant$^*$} & \multirow{2}{*}{16-16-3} & 20.0 & 74.2 & 81.4 & 44.4 & 13.8 & 46.8 & \cellcolor{red!10}{\footnotesize -16.5} \\
    & \textbf{+Pre\_bias} &  & \textbf{36.7} & \textbf{91.2} &\textbf{ 90.4} & \textbf{46.0} & \textbf{27.2} & \textbf{58.3} & \cellcolor{red!10}{\footnotesize \textbf{-5.0}} \\
    %%%%%%%%%%%%%%%%%%%%%%%%%%%%%%%%%%%%%%%%%%%%%%%%%%%%%%%%%%%%%%%%%%%%%%%%%%%%%%%%%%%%%%
    \hline\hline
    \end{tabular}
}
\end{center}
    \caption{
    The effect of pre-bias quantization on DeepSeek-R1-Distill-Qwen-1.5B and 7B models.
    The \textcolor[HTML]{009900}{green}, \textcolor{orange}{orange} and \textcolor{red}{red} cells stand for the lossless (\textcolor[HTML]{009900}{$\leq$1\%}), the fair (\textcolor{orange}{1\%-3\%}) and the risky (\textcolor{red}{$\geq$3\%}) respectively.
}
\label{tab:pre-bias}
\end{table*}

Based on the observations in \S~\ref{sec:apdx_kcache_outlier}, we propose two simple yet effective approaches to improve the KV cache quantization of DeepSeek-R1-Distill-Qwen-1.5B and 7B models. In Table~\ref{tab:r1-qwen}, we adopt these approaches as default setups and report their results on DeepSeek-R1-Distill-Qwen-1.5B and 7B models accordingly.

\paragraph{Pre-bias Per-channel K Cache Quantization for KVQuant$^*$.}
As detailed in Appendix~\ref{sec:apdx_algo}, KVQuant$^*$ employs pre-RoPE per-channel K cache quantization. Given that the extreme outlier channels are induced by bias terms, we propose to quantize K cache before adding the bias term in the key projection layer to further reduce the quantization range. As shown in Figure~\ref{fig:1.5b_7b_kv}, the pre-bias K cache appears much smoother, suggesting greater compatibility with quantization. In Table~\ref{tab:pre-bias}, it can be seen that pre-bias quantization shows consistent improvements across different models and bit-widths. For instance, it brings an average accuracy improvement of 7.81\% for 3-bit KV cache only quantization.

\begin{figure*}
    \centering
    \includegraphics[width=0.6\linewidth]{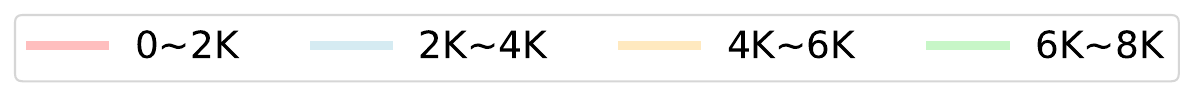}
    \\
    \subfigure[$\m Y_k$ of the $0^{\textrm{th}}$ Transformer layer in DeepSeek-R1-Distill-Qwen-1.5B.]{\includegraphics[width=0.24\linewidth]{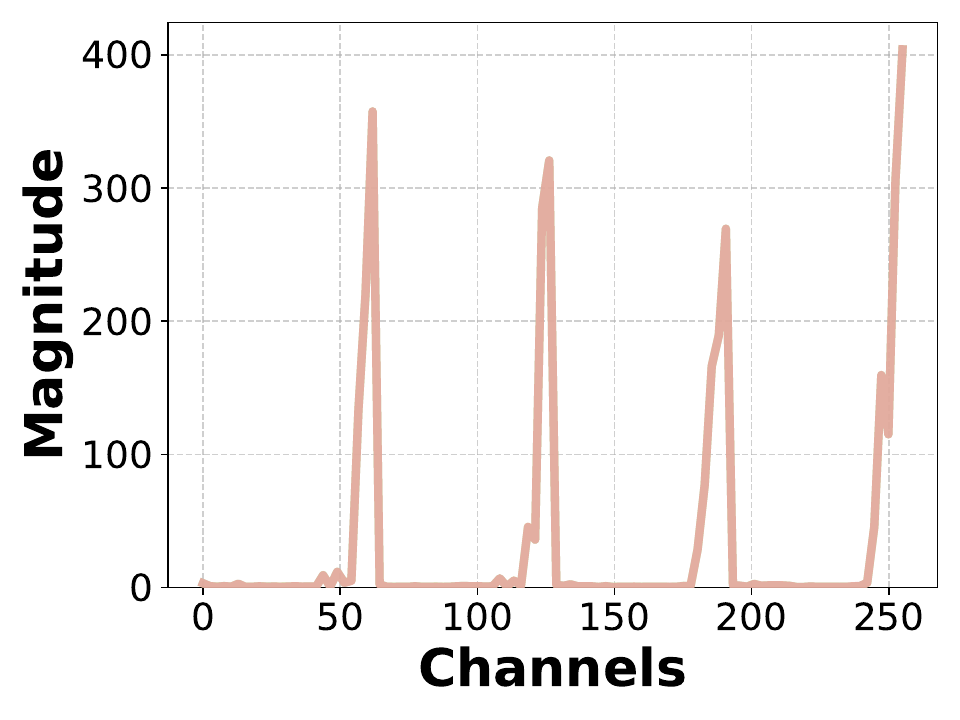}}
    \subfigure[K cache of the $0^{\textrm{th}}$ Transformer layer in DeepSeek-R1-Distill-Qwen-1.5B.]{\includegraphics[width=0.24\linewidth]{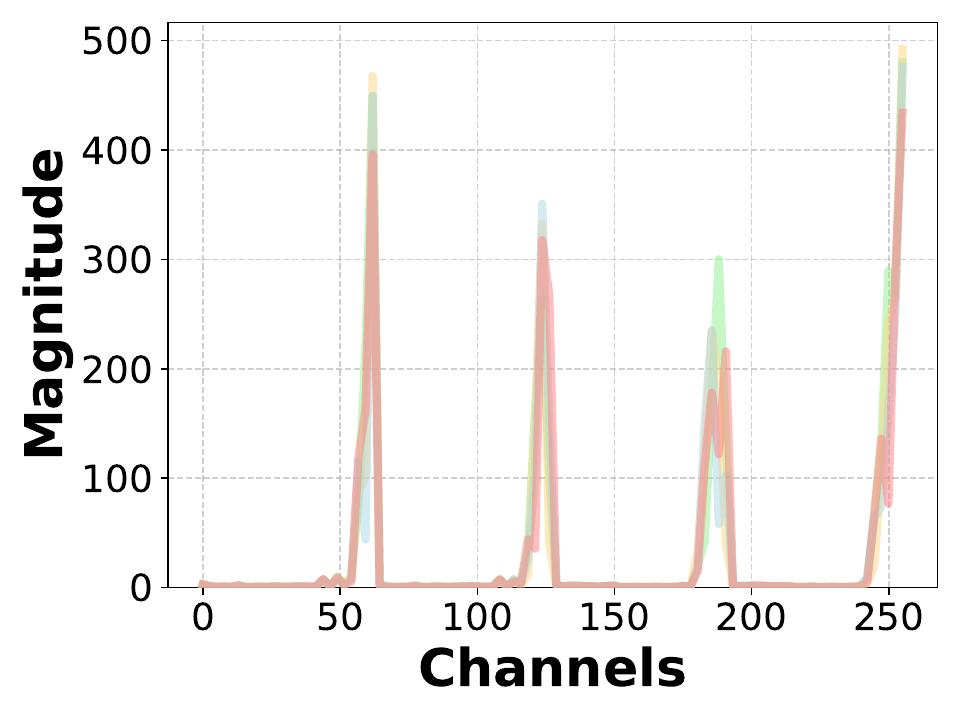}}
    \subfigure[$\m Y_k$ cache of the $0^{\textrm{th}}$ Transformer layer in DeepSeek-R1-Distill-Qwen-7B.]{\includegraphics[width=0.24\linewidth]{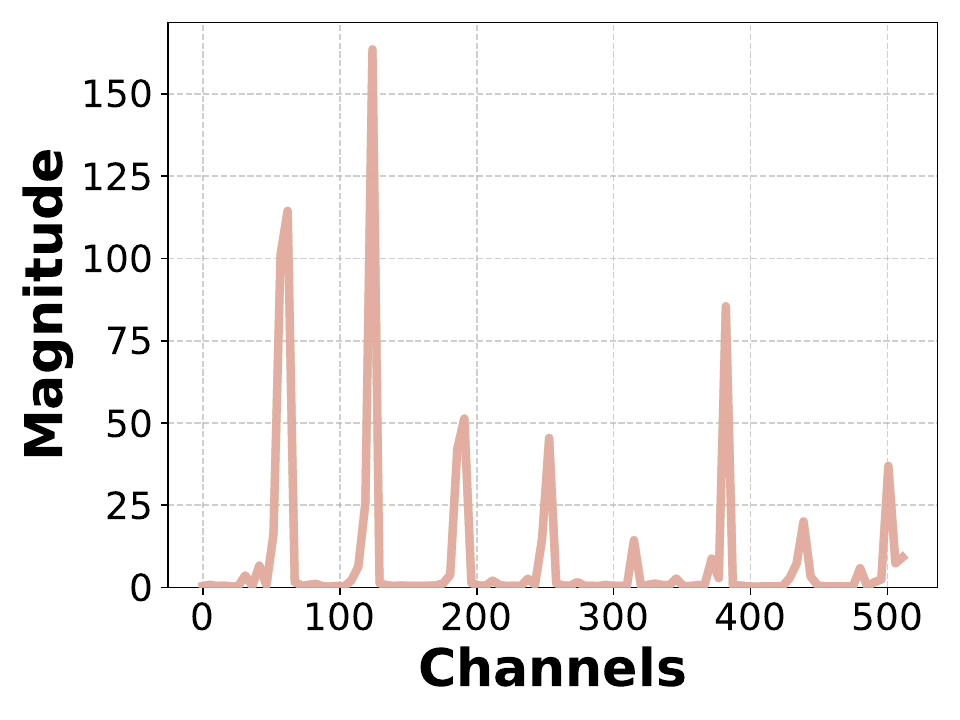}}
    \subfigure[K cache of the $0^{\textrm{th}}$ Transformer layer in DeepSeek-R1-Distill-Qwen-7B.]{\includegraphics[width=0.24\linewidth]{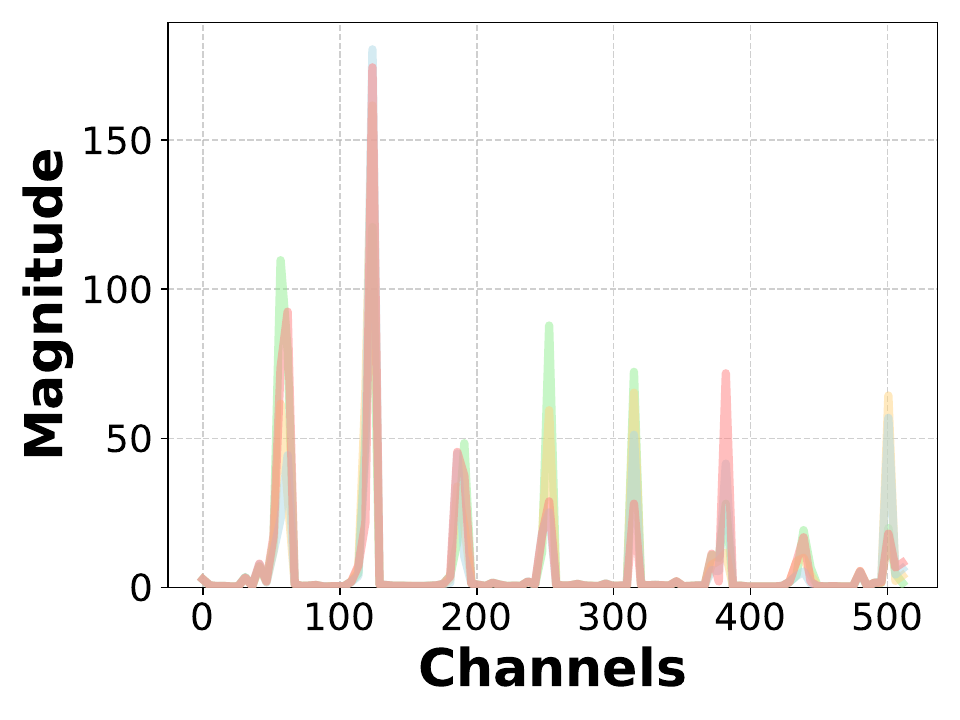}}
    \vspace{-1.5ex}
    \caption{Channel-wise magnitude distributions of K cache before and after RoPE. The statistics are computed over different sequence length ranges. $\m Y_k$ denotes the outputs of the self-attention key projection layer.}
    \label{fig:1.5b_7b_seqlen}
    \vspace{-1.5ex}
\end{figure*}

\begin{table*}[!t]
\begin{center}
\resizebox{1.00\linewidth}{!}
{
    \begin{tabular}{l|l|l|l|rrr|r|r|r|r}
    \hline\hline
    \textbf{Methods} & \textbf{\makecell{Sequence\\Length}} & \textbf{\makecell{W-A-KV\\\# Bits}} & \textbf{\makecell{Group\\Size}} & \textbf{\makecell{AIME-\\120}}  & \textbf{\makecell{MATH-\\500}} & \textbf{GSM8K} & \textbf{\makecell{GPQA-\\Diamond}} & \textbf{\makecell{LiveCode-\\Bench}} & \textbf{Avg.} & \textbf{Drop}$\downarrow$ \\ 
    \hline
    %%%%%%%%%%%%%%%%%%%%%%%%%%%%%%%%%%%%%%%%%%%%%%%%%%%%%%%%%%%%%%%%%%%%%%%%%%%%%%%%%%%%%%
    % BF16
    \textbf{BF16} & - & - & - & \textbf{21.7} & \textbf{84.4} & \textbf{84.6} & \textbf{36.9} & \textbf{16.0} & \textbf{48.7} & - \\
    %%%%%%%%%%%%%%%%%%%%%%%%%%%%%%%%%%%%%%%%%%%%%%%%%%%%%%%%%%%%%%%%%%%%%%%%%%%%%%%%%%%%%%
    % FlatQuant
    \cline{1-11}
    \multirow{3}{*}{\textbf{FlatQuant}} & \hspace{0.5ex}2K & \multirow{3}{*}{4-4-4} & \multirow{3}{*}{\textit{N/A}} & 2.5 & 53.4 & 73.8 & 30.3 & 2.2 & 32.5 & \cellcolor{red!10}{\footnotesize -16.3} \\
     & \hspace{0.5ex}4K & & & 10.0 & 64.8 & 78.6 & 31.8 & 6.7 & 38.4 & \cellcolor{red!10}{\footnotesize -10.3} \\
     & \hspace{0.5ex}8K & & & 15.8 & 73.2 & 78.9 & 37.9 & 9.3 & 43.0 & \cellcolor{red!10}{\footnotesize -5.7} \\
    %%%%%%%%%%%%%%%%%%%%%%%%%%%%%%%%%%%%%%%%%%%%%%%%%%%%%%%%%%%%%%%%%%%%%%%%%%%%%%%%%%%%%%
    \hline\hline
    \end{tabular}
}
\end{center}
    \caption{
    The effect of extending calibration sequence length on DeepSeek-R1-Distill-Qwen-1.5B.
    The \textcolor[HTML]{009900}{green}, \textcolor{orange}{orange} and \textcolor{red}{red} cells stand for the lossless (\textcolor[HTML]{009900}{$\leq$1\%}), the fair (\textcolor{orange}{1\%-3\%}) and the risky (\textcolor{red}{$\geq$3\%}) respectively.
}
\label{tab:seqlen}
\end{table*}

\paragraph{Extending the Sequence Length of Calibration Data for FlatQuant.}
Reasoning models tend to generate long thinking steps before giving the final answer, which makes it important to maintain the quantization accuracy over very long sequences. However, we find that the extreme outliers induced by bias terms pose a great challenge to learning-based outlier smoothing quantization methods such as FlatQuant, severely degrading the accuracy over long sequences. Specifically, the extreme outliers in the output of the key projection layer are further rotated into adjacent channels through RoPE. As shown in Figure~\ref{fig:1.5b_7b_seqlen}, the channel-wise magnitude distributions vary significantly across sequence positions due to differing rotation degrees. As a result, the outlier-smoothing transformations learned on the calibration data with short sequence lengths (e.g. 2K) are unable to generalize to longer sequences. Therefore, we choose to extend the sequence length of calibration data for FlatQuant to mitigate this issue. In Table~\ref{tab:seqlen}, we vary the lengths of the calibration data in \{2K, 4K, 8K\}. The accuracy keeps increasing as the length of calibration data increases. In Table~\ref{tab:r1-qwen}, we report the results of FlatQuant with the calibration data length of 4K to achieve a balance between training efficiency and quantization accuracy.

\section{Qualitative Examples}
\label{sec:apdx_qualitative}

To learn the concrete behaviors of quantized reasoning models, 
We present several qualitative examples in Table~\ref{tab:case1_bf16}-\ref{tab:case2_mxfp4}. 

We divide the reasoning process into four parts — "\textcolor{color_understanding}{understand}", "\textcolor{color_process}{process}", "\textcolor{color_check}{check}", and "\textcolor{color_answer}{final answer}" — and use different colors to represent each part. 
As shown, the quantized models follow a similar reasoning pattern as the BF16 model, even if the problem is incorrectly answered~(Table~\ref{tab:case2_mxfp4}).
% It can be found from that there is no significant change in the thinking pattern before and after quantization. 
Nonetheless, the quantized models may occasionally exhibit calculation errors or adopt slightly different verification strategies, but the overall reasoning structure remains consistent. This suggests that the model's reasoning pattern is relatively robust to quantization, even under aggressive low-bit settings like MXFP4.

\begin{table}[h]
    \centering
    \renewcommand{\arraystretch}{1.2} 
    \setlength{\tabcolsep}{8pt} 
    \begin{tabular}{p{0.95\linewidth}}
        \begin{tcolorbox}[colback=gray!5, colframe=gray!40, sharp corners, width=\linewidth, boxrule=0.5mm]
        \small
        \textbf{Problem: }
        In a class of 50 students, 28 participate in MATHCOUNTS, 21 participate in science club, and 6 students participate in neither. How many students participate in both MATHCOUNTS and science club?
        \end{tcolorbox} \\[-6pt] 
        
        \begin{tcolorbox}[colback=gray!5, colframe=gray!40, sharp corners, width=\linewidth, boxrule=0.5mm]
        
        % \textbf{Output: }\\ 
        \small
        \textbf{Reasoning: }\\ 
        \tiny
            \textcolor{color_understanding}{
            Okay, so I've got this problem here: In a class of 50 students, 28 participate in MATHCOUNTS, 21 participate in science club, and 6 students participate in neither. I need to find out how many students participate in both MATHCOUNTS and the science club. Hmm, let me think about how to approach this.
            }
            \textcolor{color_process}{
            \\\\First, I remember something about sets and maybe using a Venn diagram. Yeah, that sounds right. So, there are two groups here: MATHCOUNTS participants and science club participants. Some students might be in both, and some might be in neither. The total number of students is 50, and 6 are in neither, so that means 50 - 6 = 44 students are in at least one of the two clubs.\\\\Alright, so 44 students are in MATHCOUNTS, science club, or both. Now, I know that the number of students in both can be found using the principle of inclusion-exclusion. The formula is:\\\\Total in either club = Number in MATHCOUNTS + Number in Science Club - Number in Both.\\\\Let me write that down:\\\\Total = M + S - Both.\\\\We know the total number of students in either club is 44, MATHCOUNTS has 28 students, and science club has 21. Plugging those numbers in:\\\\44 = 28 + 21 - Both.\\\\Let me compute 28 + 21 first. That's 49. So:\\\\44 = 49 - Both.\\\\Hmm, so to find Both, I can rearrange the equation:\\\\Both = 49 - 44 = 5.
            }
            \textcolor{color_check}{
            \\\\Wait, so does that mean 5 students are in both MATHCOUNTS and the science club? Let me check if that makes sense.\\\\If 28 are in MATHCOUNTS and 5 of them are also in the science club, then 28 - 5 = 23 are only in MATHCOUNTS. Similarly, 21 are in the science club, and 5 of those are also in MATHCOUNTS, so 21 - 5 = 16 are only in the science club. Adding those together: 23 (only MATHCOUNTS) + 16 (only science club) + 5 (both) = 44, which matches the total number of students in at least one club. And then the 6 students not in either club make the total 50. That all checks out.\\\\So, yeah, I think 5 is the right answer. I don't think I made any mistakes here. I used the inclusion-exclusion principle correctly, and the numbers add up when I break it down. So I'm confident that 5 students participate in both MATHCOUNTS and the science club.\\\\
            }
            \textcolor{color_answer}{
            **Final Answer**\\The number of students participating in both MATHCOUNTS and science club is \boxed{5}.\\
            }

        \end{tcolorbox} \\
        
        % \begin{tcolorbox}[colback=gray!5, colframe=gray!40, sharp corners, width=\linewidth, boxrule=0.5mm]
        % \tiny
        % \textbf{Output: }\\ 
        %     In a class of 50 students, 28 participate in MATHCOUNTS, 21 participate in science club, and 6 students participate in neither. We need to find the number of students who participate in both MATHCOUNTS and science club.\\\\First, we calculate the number of students participating in at least one of the clubs:\\\[\\50 - 6 = 44\\\]\\\\Using the principle of inclusion-exclusion, we set up the equation:\\\[\\\text{Total in either club} = \text{Number in MATHCOUNTS} + \text{Number in Science Club} - \text{Number in Both}\\\]\\\\Substituting the known values:\\\[\\44 = 28 + 21 - \text{Both}\\\]\\\\Solving for the number of students in both clubs:\\\[\\44 = 49 - \text{Both}\\\]\\\[\\\text{Both} = 49 - 44 = 5\\\]\\\\Thus, the number of students participating in both MATHCOUNTS and science club is \boxed{5}.
        % \end{tcolorbox} \\

    \end{tabular}
    \vspace{-12pt}
    \caption{An example from MATH-500 correctly answered by the BF16 DeepSeek-R1-Distill-Qwen-32B model.}
    \label{tab:case1_bf16}
\end{table}

\begin{table}[h]
    \centering
    \renewcommand{\arraystretch}{1.2} 
    \setlength{\tabcolsep}{8pt} 
    \begin{tabular}{p{0.95\linewidth}}
        \begin{tcolorbox}[colback=gray!5, colframe=gray!40, sharp corners, width=\linewidth, boxrule=0.5mm]
        \small
        \textbf{Problem: }
        In a class of 50 students, 28 participate in MATHCOUNTS, 21 participate in science club, and 6 students participate in neither. How many students participate in both MATHCOUNTS and science club?
        \end{tcolorbox} \\[-6pt] 
        
        \begin{tcolorbox}[colback=gray!5, colframe=gray!40, sharp corners, width=\linewidth, boxrule=0.5mm]
        
        % \textbf{Output: }\\ 
        \textbf{Reasoning: }\\ 
        \tiny
            \textcolor{color_understanding}{
            Okay, so I've got this problem here: In a class of 50 students, 28 participate in MATHCOUNTS, 21 participate in science club, and 6 students participate in neither. The question is asking how many students participate in both MATHCOUNTS and science club. Hmm, okay, let me think about how to approach this.
            }
            \textcolor{color_process}{
            \\\\First, I remember that when dealing with overlapping groups like this, it's a classic inclusion-exclusion problem. The formula that comes to mind is:\\\\Total = Number in MATHCOUNTS + Number in Science Club - Number in Both + Number in Neither\\\\Wait, let me write that down to make it clearer. So, the total number of students is equal to those in MATHCOUNTS plus those in science club minus those in both, plus those in neither. Hmm, let me make sure I have that right.\\\\Actually, I think the formula is:\\\\Total = (MATHCOUNTS + Science Club - Both) + Neither\\\\Because the ones in neither are outside both groups. So, the total is the union of MATHCOUNTS and Science Club plus the ones in neither. The union is calculated as MATHCOUNTS plus Science Club minus Both.\\\\Let me write that out:\\\\Total = (MATHCOUNTS + Science Club - Both) + Neither\\\\So, plugging in the numbers:\\\\50 = (28 + 21 - Both) + 6\\\\Okay, let me compute that step by step. First, 28 plus 21 is 49. So, 49 minus Both plus 6 equals 50. So, 49 + 6 is 55, right? So, 55 minus Both equals 50.\\\\Hmm, so 55 - Both = 50. Then, subtracting 50 from both sides, we get 55 - 50 = Both, so Both = 5. So, does that mean 5 students are in both MATHCOUNTS and science club?
            }
            \textcolor{color_check}{
            \\\\Wait, let me double-check because sometimes these inclusion-exclusion problems can be tricky. So, if there are 50 students total, and 6 are in neither, that means 50 - 6 = 44 students are in at least one of the two clubs. So, 44 students are in MATHCOUNTS, science club, or both.\\\\Now, MATHCOUNTS has 28 students, and science club has 21. If we add those together, 28 + 21 = 49. But wait, that's more than 44, which is the number of students in at least one club. So, that means the overlap must be 49 - 44 = 5. So, yeah, that confirms it. 5 students are in both.\\\\Let me make sure I'm not making a mistake here. So, if 5 students are in both, then the number only in MATHCOUNTS would be 28 - 5 = 23, and only in science club would be 21 - 5 = 16. Then, adding those up: 23 (only MATHCOUNTS) + 16 (only science) + 5 (both) + 6 (neither) = 23 + 16 is 39, plus 5 is 44, plus 6 is 50. That adds up correctly.\\\\So, yeah, it seems like 5 is the right answer. I think that makes sense because when you have more students in both clubs than the total number in at least one, the difference is the overlap. So, 28 + 21 = 49, but only 44 are in at least one, so 5 must be in both.\\\\Wait, another way to think about it is using a Venn diagram. Imagine two overlapping circles: one for MATHCOUNTS with 28 and one for science club with 21. The total area covered by the circles is 44 students. The sum of the individual circles is 49, so the overlapping part must account for the difference, which is 5. So, yeah, that's consistent.\\\\I guess another way to write the formula is:\\\\Number in Both = MATHCOUNTS + Science Club - Total in at least one\\\\Which would be:\\\\Number in Both = 28 + 21 - 44 = 5\\\\So, that also gives 5. So, all the methods point to 5. I think I'm confident that 5 is the correct answer.\\\\Just to recap:\\\\- Total students: 50\\- Neither: 6\\- Therefore, in at least one club: 50 - 6 = 44\\- MATHCOUNTS: 28\\- Science Club: 21\\- Total in both clubs: 28 + 21 = 49\\- But since only 44 are in at least one, the overlap is 49 - 44 = 5\\\\Yeah, that seems solid. So, I think the answer is 5 students participate in both MATHCOUNTS and science club.\\\\
            }
            \textcolor{color_answer}{
            **Final Answer**\\The number of students who participate in both MATHCOUNTS and science club is \boxed{5}.
            }

        \end{tcolorbox} \\
        
        % \begin{tcolorbox}[colback=gray!5, colframe=gray!40, sharp corners, width=\linewidth, boxrule=0.5mm]
        % \tiny
        % \textbf{Output: }\\ 
        %     To determine how many students participate in both MATHCOUNTS and science club, we use the principle of inclusion-exclusion. \\\\Given:\\- Total number of students: 50\\- Number of students in MATHCOUNTS: 28\\- Number of students in science club: 21\\- Number of students participating in neither: 6\\\\First, we calculate the number of students participating in at least one of the clubs:\\\[ 50 - 6 = 44 \]\\\\Using the inclusion-exclusion principle, we know:\\\[ \text{Total in at least one club} = \text{MATHCOUNTS} + \text{Science Club} - \text{Both} \]\\\\Plugging in the numbers:\\\[ 44 = 28 + 21 - \text{Both} \]\\\\Solving for the number of students in both clubs:\\\[ 44 = 49 - \text{Both} \]\\\[ \text{Both} = 49 - 44 \]\\\[ \text{Both} = 5 \]\\\\Thus, the number of students who participate in both MATHCOUNTS and science club is \(\boxed{5}\).
        % \end{tcolorbox} \\

    \end{tabular}
    \vspace{-8pt}
    \caption{An example from MATH-500 correctly answered by the DeepSeek-R1-Distill-Qwen-32B model quantized with GPTQ-W4A16G128}
    \label{tab:case1_w4a16}
\end{table}

\begin{table}[h]
    \centering
    \renewcommand{\arraystretch}{1.2} 
    \setlength{\tabcolsep}{8pt} 
    \begin{tabular}{p{0.95\linewidth}}
        \begin{tcolorbox}[colback=gray!5, colframe=gray!40, sharp corners, width=\linewidth, boxrule=0.5mm]
        \small
        \textbf{Problem: }
        Let $A = (1,-11,2),$ $B = (3,-4,1),$ and $C = (-2,1,-1).$  Compute $\angle ABC,$ in degrees.
        \end{tcolorbox} \\[-6pt] 
        
        \begin{tcolorbox}[colback=gray!5, colframe=gray!40, sharp corners, width=\linewidth, boxrule=0.5mm]
        
        % \textbf{Output: }\\ 
        \textbf{Reasoning: }\\\\
        \tiny
            \textcolor{color_understanding}{
            Okay, so I need to find the angle at point B in triangle ABC, which is angle ABC. The points are given as A = (1, -11, 2), B = (3, -4, 1), and C = (-2, 1, -1). Hmm, I remember that to find the angle between two vectors, I can use the dot product formula. Since angle ABC is the angle at point B, I think I need to find the vectors BA and BC, and then find the angle between them. \\\\Let me first visualize this. Point B is the vertex where the angle is, so BA is the vector from B to A, and BC is the vector from B to C. Once I have these two vectors, I can use the dot product formula to find the angle between them. \\\\
            }
            \textcolor{color_process}{
            So, how do I find vectors BA and BC? Well, vectors can be found by subtracting the coordinates of the initial point from the terminal point. So for vector BA, it's A minus B, and for vector BC, it's C minus B. Let me write that down.\\\\Vector BA = A - B = (1 - 3, -11 - (-4), 2 - 1) = (-2, -7, 1)\\Vector BC = C - B = (-2 - 3, 1 - (-4), -1 - 1) = (-5, 5, -2)\\\\Okay, so now I have BA = (-2, -7, 1) and BC = (-5, 5, -2). Next step is to find the dot product of BA and BC, and then divide it by the product of their magnitudes. That should give me the cosine of the angle between them, right?\\\\Let me compute the dot product first. The dot product of BA and BC is calculated by multiplying corresponding components and adding them up. So:\\\\$BA \cdot BC = (-2)(-5) + (-7)(5) + (1)(-2) = 10 - 35 - 2 = -27$\\\\Hmm, that's a negative number. That means the angle is obtuse, since the cosine is negative. Okay, moving on.\\\\Now, I need the magnitudes of BA and BC. The magnitude of a vector is found by taking the square root of the sum of the squares of its components.\\\\First, magnitude of BA:\\\\$|BA| = sqrt[ (-2)^2 + (-7)^2 + (1)^2 ] = sqrt[4 + 49 + 1] = sqrt[54] = 3*sqrt(6)$\\\\Wait, sqrt(54) is equal to sqrt(9*6) which is 3*sqrt(6). That seems right.\\\\Now, magnitude of BC:\\\\$|BC| = sqrt[ (-5)^2 + (5)^2 + (-2)^2 ] = sqrt[25 + 25 + 4] = sqrt[54] = 3*sqrt(6)$\\\\Oh, interesting, both vectors BA and BC have the same magnitude, 3*sqrt(6). So that simplifies things a bit.\\\\Now, using the dot product formula:\\\\$cos(theta) = (BA \cdot BC) / (|BA| |BC|)$\\\\Plugging in the numbers:\\\\$cos(theta) = (-27) / (3*sqrt(6) * 3*sqrt(6)) = (-27) / (9*6) = (-27)/54 = -0.5$\\\\So, cos(theta) = -0.5. Now, to find theta, I need to take the arccosine of -0.5.\\\\I remember that cos(120\textdegree) = -0.5, because cos(60\textdegree) is 0.5 and cosine is negative in the second quadrant. So, arccos(-0.5) is 120 degrees.\\\\Therefore, angle ABC is 120 degrees.\\\\
            % \textellipsis
            }
            \textcolor{color_check}{
            Wait, just to double-check everything. Let me recap:\\\\1. Vectors BA and BC are correct. BA is (-2, -7, 1), BC is (-5, 5, -2). Yes, that's right because we subtract B from A and B from C.\\\\2. Dot product: (-2)(-5) = 10, (-7)(5) = -35, (1)(-2) = -2. So 10 - 35 -2 = -27. That's correct.\\\\3. Magnitudes: sqrt(4 + 49 +1) = sqrt(54), and sqrt(25 +25 +4) = sqrt(54). So both are 3*sqrt(6). Correct.\\\\4. So, cos(theta) = (-27)/(54) = -0.5. Yes, that's right.\\\\5. Arccos(-0.5) is 120 degrees. That's correct because in the unit circle, 120 degrees is in the second quadrant with cosine -0.5.\\\\Therefore, I think I did everything correctly, and the angle is indeed 120 degrees.\\\\
            % \textellipsis
            }
            \textcolor{color_answer}{
            **Final Answer**\\The measure of angle ABC is \boxed{120} degrees.
            }

        \end{tcolorbox} \\

    \end{tabular}
    \vspace{-8pt}
    \caption{An example from MATH-500 correctly answered by the BF16 DeepSeek-R1-Distill-Qwen-32B model.}
    \label{tab:case2_BF16}
\end{table}

\begin{table}[h]
    \centering
    \renewcommand{\arraystretch}{1.2} 
    \setlength{\tabcolsep}{8pt} 
    \begin{tabular}{p{0.95\linewidth}}
        \begin{tcolorbox}[colback=gray!5, colframe=gray!40, sharp corners, width=\linewidth, boxrule=0.5mm]
        \small
        \textbf{Problem: }
        Let $A = (1,-11,2),$ $B = (3,-4,1),$ and $C = (-2,1,-1).$  Compute $\angle ABC,$ in degrees.
        \end{tcolorbox} \\[-6pt] 
        
        \begin{tcolorbox}[colback=gray!5, colframe=gray!40, sharp corners, width=\linewidth, boxrule=0.5mm]
        
        % \textbf{Output: }\\ 
        \textbf{Reasoning: }\\\\ 
        \tiny
            \textcolor{color_understanding}{Okay, so I need to find the angle at point B in triangle ABC. The points given are A = (1, -11, 2), B = (3, -4, 1), and C = (-2, 1, -1). Wait, hold on, the coordinates for point C are written as (-, 1, -1). That doesn't look right. Is that a typo or a mistake? Maybe it should be something like (-1, 1, -1) or (-2, 1, -1) with a specific number? Hmm, maybe it's a typo and supposed to be (-1, 1, -1). I'll assume that for now, so point C is (-1, 1, -1). I think that makes sense.\\So, points are A = (1, -11, 2), B = (3, -4, 1), and C = (-1, 1, -1). I need to compute angle ABC, which is the angle at point B between points A, B, and C. So, angle ABC is the angle at vertex B, formed by sides BA and BC. So, to find this angle, I need to find the vectors BA and BC, then compute the angle between these two vectors using the dot product formula.\\\\
            }
            \textcolor{color_process}{First, I need to compute vectors BA and BC.\\Vector BA is from B to A, so it's A - B. Similarly, vector BC is from B to C, so it's C - B.\\Let me compute vector BA first.\\A is (1, -11, 2), B is (3, -4, 1). So vector BA is A - B, which is (1 - 3, -11 - (-4), 2 - 1). Let me calculate each component:\\1 - 3 is -2.\\-11 - (-4) is -11 + 4, which is -7.\\2 - 1 is 1.\\So vector BA is (-2, -7, 1). Wait, hold on, that would be (-2, -7, 1). Wait, is that right? 1 - 3 is -2, so first component is -2. Second component is -11 - (-4) is -11 + 4 is -7. Third component is 2 - 1 is 1. So vector BA is (-2, -7, 1). Wait, maybe the first component is -2, so it's (-2, -7, 1) or (-2, -7, 1). Hmm, I think it's (-2, -7, 1). Maybe the person meant to write (-2, -7, 1), but the first component is -2. So, in components, it's (-2, -7, 1). So, BA = (-2, -7, 1). Wait, maybe it's (-2, -7, 1), but the first component is -2.\\
            \textellipsis
            \\\\Therefore, arccos(-0.588) is 180 - 54 = 126 degrees.\\\\Wait, 54 degrees, so 180 - 54 is 126 degrees. So, the angle is approximately 126 degrees.\\\\
            }
            \textcolor{color_check}{Wait, let me confirm.\\\\If arccos(-0.588) is approximately 126 degrees because cos(126 degrees) is $cos(180 - 54) = -cos(54) \approx -0.5878$, which is about -0.588. So, that's exactly what we have. So, $cos(theta) \approx -0.588$, so theta is approximately 126 degrees.\\\\So, the angle ABC is approximately 126 degrees.\\\\Wait, let me double-check my calculations.\\\\First, vectors BA and BC:\\\\Vector $BA: A - B = (1 - 3, -11 - (-4), 2 - 1) = (-2, -7, 1)\\\\Vector BC: C - B = (-1 - 3, 1 - (-4), -1 - 1) = (-4, 5, -2)$\\\\
            \textellipsis
            \\\\Vector BC is from B to C, which is $C - B$, which is $(-1 - 3, 1 - (-4), -1 - 1) = (-4, 5, -2)$. Correct.\\\\Dot product: $(-2)*(-4) = 8, (-7)*(5) = -35, (1)*(-2) = -2$. Total: $8 -35 -2 = -29$. Correct.\\\\Magnitudes: $|BA| = sqrt[(-2)^2 + (-7)^2 + 1^2] = sqrt[4 + 49 + 1] = sqrt[54] = 3*sqrt(6)$. Correct.\\\\$|BC| = sqrt[(-4)^2 + 5^2 + (-2)^2] = sqrt[16 + 25 + 4] = sqrt[45] = 3*sqrt(5)$. Correct.\\\\So, $cos(theta) = (-29)/(3*sqrt(6)*3*sqrt(5)) = (-29)/(9*sqrt(30))$.\\\\Yes, that's correct. So, the angle is arccos(-29/(9*sqrt(30))).\\\\So, in degrees, that is approximately 126 degrees.\\\\
            }
            \textcolor{color_answer}{Therefore, the angle ABC is approximately 126 degrees.\\\\I think that's the answer.\\\\**Final Answer**\\The measure of angle ABC is \boxed{126} degrees.
            }

        \end{tcolorbox} \\

    \end{tabular}
    \vspace{-8pt}
    \caption{An example from MATH-500 that was incorrectly answered by the MXFP4-quantized DeepSeek-R1-Distill-Qwen-32B model.}
    \label{tab:case2_mxfp4}
\end{table}

\end{document}